\documentclass[11pt,a4paper]{article}
\usepackage[margin=2.5cm]{geometry}
\usepackage[T1]{fontenc}
\usepackage[utf8]{inputenc}
\usepackage{lmodern}
\usepackage{microtype}
\usepackage{amsmath,amssymb}
\usepackage{booktabs}
\usepackage{array}
\usepackage{longtable}
\usepackage{graphicx}
\usepackage{xcolor}
\usepackage{pgfplots}
\pgfplotsset{compat=1.18}
\usetikzlibrary{positioning,arrows.meta,fit,calc}
\usepackage[numbers,sort&compress]{natbib}
\usepackage[hidelinks]{hyperref}
\usepackage{enumitem}
\usepackage{caption}
\graphicspath{{figures/}}
\newcommand{\hashtrunc}[2]{\texttt{#1\ldots#2}}
\newcommand{\um}{\textmu m}
\newcommand{\tox}{TX100}

\title{When Can a Learned Design Surrogate Be Trusted in a Self-Driving Microfluidics Laboratory?\\
An Offline Closed-Loop Framework Built and Tested on Six Public Tables}
\author{Zhiyuan Zhang\thanks{Independent research. Correspondence: see the public
  repository \url{https://github.com/LibertychaserUS/microfluidic-native-physics}.
  Manuscript text and original figures are released under CC~BY~4.0; code under
  Apache-2.0.}}
\date{v0.5.5 --- 26 October 2026}

\begin{document}
\maketitle

\begin{abstract}
Self-driving laboratories couple a learned design surrogate to an instrument,
and on every new target domain the surrogate must answer four questions: when
it can be trusted, how many target measurements it needs, which to make first,
and when to stop adapting, switch models or escalate. We state a framework that
makes each a measurable quantity (a competence ratio, a budget $k^{*}(\varepsilon)$, a time-to-target and a
priced decision) and instantiate it without a wet experiment on six public
droplet-microfluidic and liposome tables used as offline laboratories: 107
leave-one-group tiers whose rows are hidden labels. The evidence, a sealed and pre-registered programme (two sealed
negative validations and a pre-registered test on three further tables) and
seven frozen closed-loop protocols each run twice, yields twenty-three
findings. Shift cost is
stratified by shift type; no label-free statistic certifies individual
candidates; random-split rankings of model families predicted neither the
sealed nor the loop rankings. Warm-start residual boosting returns a surrogate
to $1.25\times$ its development reference within 20 target rows in 94\% of
chemistry tiers (27\% under a strict reading of the protocol's noise-floor
rule), 90\% of laboratory and 40\% of device tiers, and a budget planned from
the shift type on other tables is not measurably worse on a new one. In the
loop, one gain survives drift and shuffled-label controls over all tiers,
carried by the sparsely sampled ones: proposing at once and warm-starting on
every measurement reaches a 10\% target in 2.80 measurements against 3.31 for
the reference, and in 5.72 against 11.35 for random proposals on 301 hard
units at a 3\% tolerance. No label-free acquisition rule survives the
shuffled-label control, no interval-aware proposal or alternative model family
does better, and no threshold rule on the tested diagnostics beats always
continuing once escalation has a price. No wet-lab validation is claimed.
\end{abstract}

\section{Introduction}
\label{sec:intro}
A self-driving laboratory in droplet microfluidics couples a learned design
surrogate to an instrument: the surrogate proposes a condition (channel
geometry, flow rates, fluid pair, surfactant or lipid chemistry), the
instrument measures the droplet or particle it produces, and the loop updates
and proposes again. Surrogates trained on public experimental
tables~\cite{lashkaripour2021,lashkaripour2024} or on simulated droplet
formation~\cite{samari2025} are now used for this kind of navigation, and
their reported accuracy typically comes from random row-level
train/test splits. The loop, however, is deployed on a new device, a new
fluid pair, a new surfactant or a new laboratory that the training table never
contained, and on every such domain it has to make four decisions: whether
the surrogate can be trusted at all (Q1), how many target measurements it
needs before it is usable (Q2), which conditions to measure first (Q3), and,
once the first measurements are in, whether to keep adapting the surrogate,
stop adapting it, abandon it for a model fitted to the target rows, or
escalate to a full recalibration (Q4).

Two numbers from our independent re-implementation of the DAFD~3.0
(Design Automation based on Fluid Dynamics) baseline~\cite{lashkaripour2024}
motivate this paper. On 15 random 80/20
splits of the 868-row comprehensive table, the published XGBoost configuration
reaches a diameter mean absolute percentage error (MAPE,
$100 \cdot \mathrm{mean}\,|\hat y - y|/y$) of $5.268\% \pm 0.421\%$; on the 64-row external
literature set released with the same paper, the error is
$10.971\% \pm 0.552\%$, roughly a doubling. Nominal 90\% conformalised
quantile-regression intervals calibrated on random splits in the pilot
analysis cover 91.03\% of
internal test rows but only 55.10\% of external rows.

Rather than proposing another inverse-design algorithm or a general
physics-informed architecture, this work states a decision framework for
learned surrogates in such a loop (Section~\ref{sec:framework}) and builds
its evidence base without a wet experiment, by treating public experimental
tables as offline laboratories. The framework defines, for each of the four
questions, a quantity that can be measured offline on a table and is
designed to be measured prospectively on an instrument with the same code; the evidence is specific to
droplet-microfluidic and liposome surrogates on six public tables and makes
no claim about hard versus soft constraints in general. Five components are
reported together:

\begin{enumerate}[leftmargin=*,itemsep=2pt]
\item \textbf{A decision framework.} Objects (source, target, oracle,
  policy), quantities (development reference, competence ratio, budget
  $k^{*}(\varepsilon)$, time-to-target, decision cost), evidence tiers and one
  decision form for every comparative statement (Section~\ref{sec:framework}).
\item \textbf{A sealed and pre-registered evidence programme on six public
  tables.} An independent re-implementation of the DAFD~3.0 baseline with a
  reproduction-fidelity check; two sealed negative validations (the complete
  Triton~X-100 (\tox{}) surfactant group of the Chagot et
  al.~\cite{chagot2022} dataset and an entire second laboratory of a public
  liposome dataset), each with labels split off before any analysis and
  revealed once; a physics-guided hard-bounded regressor frozen under its own
  protocol; three frozen method matrices over seven shift cells; and a
  pre-registered test of five hypotheses on three further tables with sealed
  tiers, followed by frozen-protocol adaptation experiments on the revealed
  sets. These units keep their original labels R1--R10 and are presented by
  question (Sections~\ref{sec:data}--\ref{sec:results}).
\item \textbf{An offline closed-loop benchmark.} Every leave-one-group tier of
  the six tables is replayed as a target laboratory (107 tiers) under seven
  frozen protocols: simulator checks, a point task over 14 loop policies, budget
  curves, a difficulty sensitivity, a priced decision rule learned by table
  rotation, a comparison of six source-model families, the transfer of planned
  budgets across tables, negative controls for every gain, and a replay on hard
  tail targets, each run twice
  (Sections~\ref{sec:eloop}, \ref{sec:q2}--\ref{sec:robust}).
\item \textbf{Twenty-three framework-level findings.} The units are condensed
  into findings F1--F23, each with its question, evidence tier and units, and
  restated as what a loop would do at each stage; negative results are
  retained in full (Section~\ref{sec:synthesis}).
\item \textbf{Open artefacts.} Protocols hashed before their runs, deviation
  registers, two-run manifests, and a ledger from which every number in this
  paper is re-derived (Section~\ref{sec:repro}).
\end{enumerate}

Section~\ref{sec:outlook} states, as a plan rather than a result, how the same
protocols move from tables to an instrument.

We make two deliberately limited claims, each qualified by the search basis in
Appendix~\ref{app:search}. First, to our knowledge, no peer-reviewed work
applies a sealed-prediction reliability validation to microfluidic ML design
surrogates; this paper reports such validations on five independent public tables
(two with a full sealed-prediction chronology and three under one
pre-registered multi-table protocol). Second, to our knowledge, no peer-reviewed work unifies physics
guidance, architecture-level hard constraints and inverse-design evaluation
for microfluidics in a single framework; the nearest neighbours are a
periodicity-enforcing network for deterministic-lateral-displacement
design~\cite{lee2025} and an optimisation-layer repair
operator~\cite{micromixer2026}, which differ from ours in how the components
are combined rather than in kind. We do not claim to be the first microfluidic
inverse-design or ML system, the first physics-guided microfluidic model, or
the first to publish a negative result.

\begin{figure}[t]
\centering\footnotesize
\newcommand{\figkeyswatch}[1]{\raisebox{-0.35ex}{\setlength{\fboxsep}{0pt}\setlength{\fboxrule}{0.4pt}\fcolorbox{black!65}{#1}{\rule{3.4mm}{0pt}\rule{0pt}{2.6mm}}}}
\begin{tikzpicture}[x=1cm,y=1cm,
  band/.style={draw=black!40,fill=black!3,rounded corners=3pt,line width=0.4pt},
  hdr/.style={anchor=north west,font=\footnotesize,inner sep=0pt,text=black!85},
  box/.style={draw=black!65,rounded corners=1pt,align=center,inner sep=3pt,
              font=\scriptsize,text width=43.75mm,minimum height=9mm,
              execute at begin node={\hyphenpenalty=10000 \exhyphenpenalty=10000
                \ifdefined\xeCJKsetup\xeCJKsetup{CJKglue=\hskip 0pt}\fi}},
  wide/.style={text width=68.75mm},
  data/.style={box,fill=black!7},
  retro/.style={box,fill=white},
  sealed/.style={box,fill=blue!12},
  post/.style={box,fill=orange!18},
  frozen/.style={box,fill=teal!14},
  trans/.style={font=\scriptsize\itshape,text=black!60,inner sep=1pt},
  flow/.style={-{Latex[length=1.8mm,width=1.6mm]},line width=0.7pt,black!55},
  key/.style={draw=black!55,dashed,rounded corners=2pt,inner sep=4pt,font=\scriptsize},
  keyttl/.style={fill=white,font=\scriptsize\bfseries,inner sep=1.5pt,text=black!85}]
\draw[band] (0,0.0) rectangle (15.2,-4.25);
\draw[band] (0,-5.0) rectangle (15.2,-8.1);
\draw[band] (0,-8.85) rectangle (15.2,-10.8);
\draw[band] (0,-11.55) rectangle (15.2,-13.5);
\node[hdr] at (0.3,-0.2) {\textbf{Stage 1} --- retrospective evaluation and mechanism diagnostics (D1--D3)};
\node[data]   (d1) at (2.6,-1.25) {D1: DAFD comprehensive and external droplet tables};
\node[data]   (d2) at (7.6,-1.25) {D2: surfactant-laden droplets, six chemistry groups};
\node[data]   (d3) at (12.6,-1.25) {D3: liposome formulations, two laboratories};
\node[retro]  (r1) at (2.6,-2.4) {R1--R4, R7: retrospective, frozen for R7};
\node[sealed] (r5) at (7.6,-2.4) {R5: sealed \tox{} chemistry, one reveal (gate consumed)};
\node[sealed] (r9) at (12.6,-2.4) {R9: sealed laboratory B, one reveal (gate consumed)};
\node[retro,wide] (r6) at (3.85,-3.55) {R6: frozen V2, retrospective after the reveal};
\node[frozen,wide] (r8) at (11.35,-3.55) {R8: method matrix on D1 and D2, frozen, two runs};
\node[trans] (t1) at (7.7,-4.625) {five hypotheses frozen before any Stage~2 download};
\draw[flow] ([xshift=-1.6mm]t1.west |- 0,-4.25) -- ([xshift=-1.6mm]t1.west |- 0,-5.0);
\node[hdr] at (0.3,-5.2) {\textbf{Stage 2} --- pre-registered test of H-A to H-E on three further tables (R10)};
\node[data]   (a1) at (2.6,-6.25) {A1: non-Newtonian flow-focusing droplets};
\node[data]   (b1) at (7.6,-6.25) {R2-A1: co-flow capillary droplets};
\node[data]   (b2) at (12.6,-6.25) {R2-A2: lipid-nanoparticle (LNP) Atlas, literature compilation};
\node[sealed] (s1) at (2.6,-7.4) {S2 fluid family and S3 device sealed; one reveal};
\node[sealed] (s2) at (7.6,-7.4) {S2 fluid pair and S3 device sealed; one reveal};
\node[sealed] (s3) at (12.6,-7.4) {S2 lipid, S3 paper and S4 preparation sealed; one reveal};
\node[trans] (t2) at (7.7,-8.475) {revealed sets only; protocols frozen before the runs};
\draw[flow] ([xshift=-1.6mm]t2.west |- 0,-8.1) -- ([xshift=-1.6mm]t2.west |- 0,-8.85);
\node[hdr] at (0.3,-9.05) {\textbf{Stage 3} --- constructive adaptation under frozen protocols on the revealed sets (post hoc)};
\node[post] (c1) at (2.6,-10.1) {small-sample recalibration (\tox{}, laboratory B)};
\node[post] (c2) at (7.6,-10.1) {warm-start residual boosting (\tox{}, laboratory B, new tiers)};
\node[post] (c3) at (12.6,-10.1) {bias-hedging retrieval portfolio; non-equilibrium feature gate};
\node[trans] (t3) at (7.7,-11.175) {all six tables as offline laboratories; protocols frozen before the runs};
\draw[flow] ([xshift=-1.6mm]t3.west |- 0,-10.8) -- ([xshift=-1.6mm]t3.west |- 0,-11.55);
\node[hdr] at (0.3,-11.75) {\textbf{Stage 4} --- offline closed-loop benchmark on 107 tiers (E0--E7, E1b; Section~\ref{sec:eloop})};
\node[frozen] (e1) at (2.6,-12.8) {E0 simulator checks; E1 point task, 14 policies; E2 budget curves};
\node[frozen] (e2) at (7.6,-12.8) {E1b difficulty; E3 priced decision; E4 source-model families};
\node[frozen] (e3) at (12.6,-12.8) {E5 budget transfer; E6 negative controls; E7 hard targets};
\node[key] (keybox) at (7.6,-14.2) {\figkeyswatch{white}\, retrospective\qquad \figkeyswatch{blue!12}\, pre-registered, sealed\qquad \figkeyswatch{teal!14}\, frozen protocol, two runs\qquad \figkeyswatch{orange!18}\, post hoc, frozen protocol};
\node[keyttl] at (keybox.north) {Key (evidence tier)};
\end{tikzpicture}
\caption{Research design. Six public tables enter in four stages, drawn as four
stage bands; within the first two bands the three columns are the three tables of
that stage, and the fourth band replays all six tables as offline laboratories. Shading marks the evidence tier of each block, as listed in
the key below the bands (white: retrospective; blue: pre-registered with
sealed labels and a single reveal gate; teal: protocol frozen before the run
and the run repeated with a byte-identical artifact manifest; orange: post hoc
on revealed labels under a protocol frozen before the run; R7, frozen with two
runs, shares the white box of R1--R4). Section~\ref{sec:results} keeps the unit labels
R1--R10 and presents them by question.}
\label{fig:design}
\end{figure}
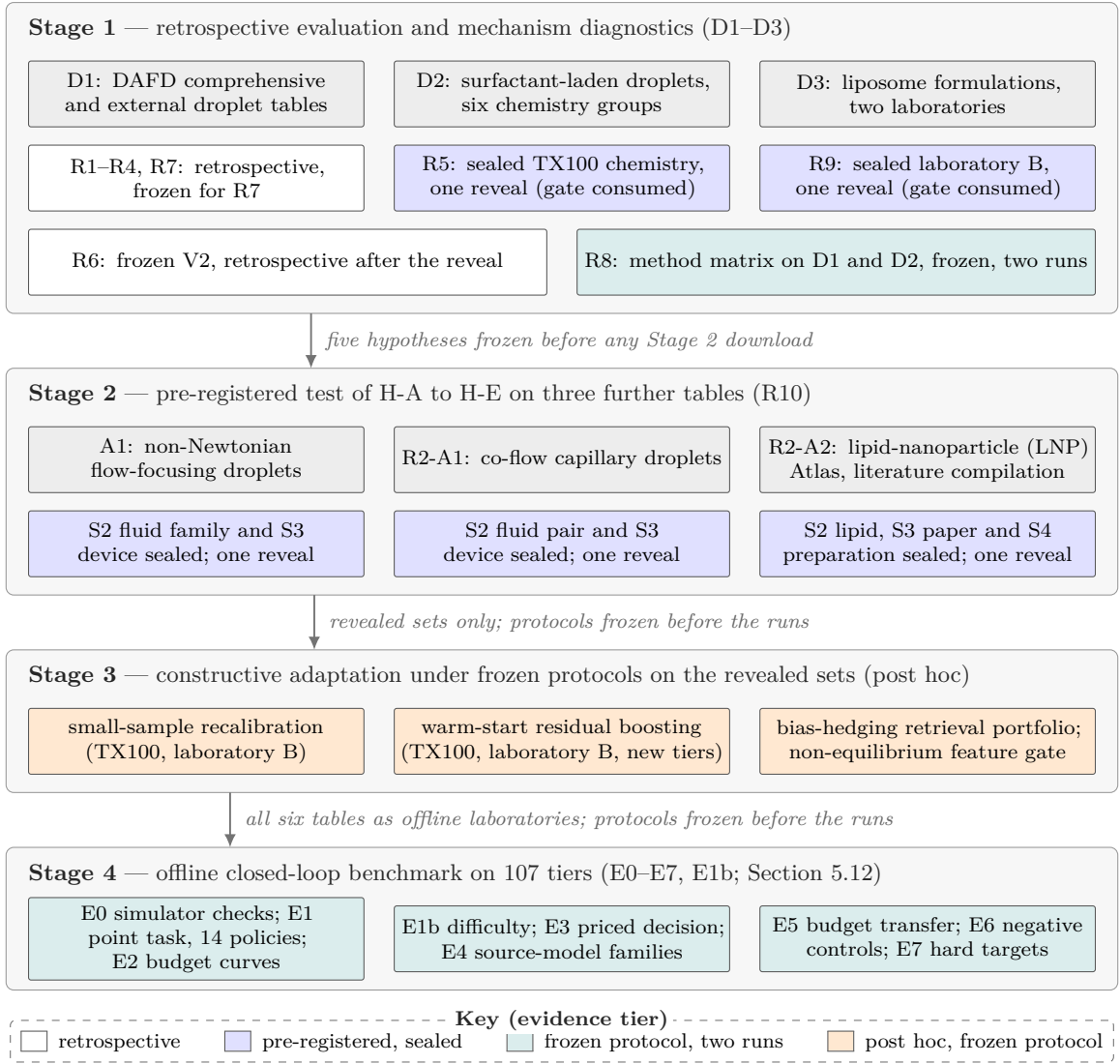

\paragraph{Pre-stated hypotheses.} The project record of 16~July~2026 fixed
four falsifiable hypotheses before the expanded analysis was implemented
(translated here from the original record; wording otherwise unchanged):
\begin{description}[leftmargin=*,itemsep=1pt]
\item[H1] Random 80/20 splitting overestimates model reliability under
  cross-literature, cross-device and cross-fluid conditions.
\item[H2] Uncertainty intervals that achieve nominal coverage on
  in-distribution random splits lose coverage markedly under independent
  literature shift.
\item[H3] A label-free applicability domain built only from the training-input
  distribution can identify a subset of high-risk external candidates and
  raise the fraction of retained candidates that meet a diameter tolerance.
\item[H4] Coupling the droplet-diameter model with the volume-conservation
  relation unifies geometry and flow-rate search into an actionable
  diameter--frequency design target.
\end{description}
H1--H3 are evaluated directly; H4 is addressed only at the level of the
mass-conservation frequency re-implementation check (Section~\ref{sec:r1}) and is
otherwise left open.
\paragraph{Research path.} The work proceeded in four stages and the paper
follows them (Figure~\ref{fig:design}; Table~\ref{tab:resultsmap} maps every
result to its dataset, protocol status and question). Stage~1 is a
retrospective evaluation with mechanism diagnostics on D1--D3 (R1--R9): it
establishes the ordering of shift effects, two sealed negative validations,
the source-level nature of label-free pruning, and the limits of hard
constraints and uncertainty-aware families. Stage~2 is a pre-registered test
of five hypotheses distilled from Stage~1 on three further public tables
(A1, R2-A1 and R2-A2; R10): hypotheses, tiers, thresholds and decision rules
were frozen before any Stage~2 data file was downloaded, and each table's
sealed labels were revealed once. Stage~3 is a set of constructive adaptation
experiments under frozen protocols on the revealed sets; it asks what a
practitioner who can measure a few target-domain rows recovers, is post hoc,
and is reported as a set of candidates for pre-registered evaluation rather
than as recommendations. Stage~4 replays all six tables as offline
laboratories under frozen protocols (E0--E7 and E1b; Section~\ref{sec:eloop}),
where loop policies are compared on 107 target tiers and the decisions of
Stage~3 are tested as policies rather than as candidates.

\section{Related Work}
\label{sec:related}

\paragraph{ML for droplet and device design.} Design automation of
flow-focusing droplet generators, which builds on the droplet-formation
physics of flow-focusing and T-junction devices~\cite{anna2003,garstecki2006},
has progressed from DAFD~1.0
\cite{lashkaripour2021} through versatility/stability
optimisation~\cite{mcintyre2023} to DAFD~3.0's single- and double-emulsion
predictor with constrained search and reported wet-lab
tests~\cite{lashkaripour2024}. Simulation-trained surrogates
(DesignFlow~\cite{samari2025}), deterministic-lateral-displacement (DLD) design
networks~\cite{vatandoust2022}, modular chip-architecture
design~\cite{li2026pathchip}, ML-automated microfluidic circuit
design~\cite{mufg2026} and flow-programming networks~\cite{ceyehao2026}
cover inverse design across device classes. We treat this literature as the
object of evaluation, not as a gap to fill.

\paragraph{Physics-guided modelling and hard constraints.} Physics-guided
microfluidic models mostly impose physics as soft penalties or as
physics-informed neural networks (PINNs) that evaluate differential-equation residuals
\cite{raissi2019,karniadakis2021,bpgnn2026,flexpinn2026}; a Bayesian-optimised physics-guided
network for electromagnetic micromixers explicitly lists the efficient
embedding of macroscopic constraints without full Navier--Stokes solution as
unresolved~\cite{bpgnn2026}. Architecture-level hard constraints appear in a
periodicity-enforcing network for DLD devices~\cite{lee2025}, and
constraint-aware optimisation with repair operators appears in autonomous
micromixer design~\cite{micromixer2026}. That hard constraints are a concept
distinct from PINN penalties, and that they can outperform soft penalties
under extrapolation, has been examined in general
settings~\cite{naturecs2021,lu2021,beucler2021,baty2023,prnet2026}, and the
failure modes of soft-penalty training are themselves
documented~\cite{krishnapriyan2021}; we do not re-litigate it. A
physics-constrained boiling surrogate reports that constraints help inside
the training envelope but hurt outside it~\cite{filmpinn2026}, which is the
kind of regime-dependent question our evaluation quantifies for a specific
microfluidic surrogate. The V2 model below is therefore called a physics-guided regressor with
hard structural constraints, not a classical PINN: the public tables contain
no velocity or pressure fields, interface coordinates, time coordinates or
boundary-condition observations at which a partial-differential-equation (PDE) residual could be evaluated.

\paragraph{Dimensionless learning versus native variables.} A substantial
line of work embeds dimensional invariance into learning, discovering
governing dimensionless groups from scarce data~\cite{xie2022,oppenheimer2023},
and a recent droplet study that combines dimensional analysis with learning
reports $R^2$ of 0.80--0.83 on unseen viscosity curves in its training device
but 0.46 for a strongly shear-thinning fluid and 0.48 for a new channel
depth~\cite{labchip2025dimless}, a degradation under rheology and device
shift rather than a failure of extrapolation as such. We do not
claim that modelling in native recorded variables is new or better; the
tabular microfluidic models cited above already operate on raw quantities. Our position is an
\emph{interface discipline}: V2 trusts only natively recorded quantities
(flow rates, equilibrium interfacial tension, concentration, CMC, a
no-surfactant indicator), and recomputes flow ratio, capillary and Reynolds
numbers deterministically from fixed units and device constants; an
optimiser or downstream user cannot supply a dimensionless number
independently. The only comparative
evidence we offer is an ablation under one protocol and one split: a
native-input XGBoost baseline (B2) has a chemistry-macro MAPE of 9.13\%
compared with 10.27\% for the published four-input baseline (B1)
(Section~\ref{sec:r6}). This is a within-protocol ablation, not a reliability
conclusion about native variables under chemistry shift, and it is not a
property of V2, which does not beat B1.

\paragraph{Reliability evaluation, preregistration and negative results.}
Assessing pointwise reliability after training was framed by Schulam and
Saria~\cite{schulam2019}; preregistration for predictive modelling has been
proposed for ML~\cite{hofman2023}; publishing negative results is an
established practice with its own venues and a recent position
paper~\cite{karl2024}; leakage is a recognised driver of inflated
claims~\cite{kapoor2023}. Applicability-domain screening has a long history
in quantitative structure--activity relationship (QSAR) modelling, where it is recommended for flagging unreliable
predictions~\cite{tropsha2003,netzeva2005,sahigara2012}, and selective
prediction supplies the risk--coverage formalism used
here~\cite{elyaniv2010}. Conformal prediction gives finite-sample marginal
coverage under exchangeability~\cite{vovk2005,romano2019}, and coverage under
covariate shift requires additional structure such as a known or estimated
density ratio~\cite{tibshirani2019,barber2023}; we claim neither for
chemistry-held-out transfer. A pre-training identifiability analysis for physics-informed surrogates
in mechanics-governed flow~\cite{cieslak2026} is the closest methodological
relative outside microfluidics. Within microfluidics, reviews name the lack of
standardised benchmarks, cross-laboratory generalisation and explicit
external validation as open problems~\cite{labchip2025review,das2025,
delgiudice2026}, and a curated cross-laboratory liposome dataset addresses the
data layer~\cite{smartmicrofluidics2026}; we found no implementation of a
sealed-prediction validation on a microfluidic design surrogate
(Appendix~\ref{app:search}).

\paragraph{Self-driving laboratories and the surrogate inside the loop.} Self-driving
laboratories couple automated synthesis or fabrication, characterisation and a
learning algorithm that chooses the next experiment~\cite{macleod2020}; for reviews
see~\cite{abolhasani2023,tom2024sdl}.
In those platforms the learned model is a component of a closed loop, not an
end product, and its value is measured by how many experiments the loop needs
rather than by a test-set error; the reviews name the choice of surrogate,
the transfer of a model from one campaign or instrument to the next, and the
decision of when to trust it as open questions. The framework of
Section~\ref{sec:framework} is a statement of those questions for one domain
with measurable answers, and the offline instantiation on public tables is a
way of answering them before an instrument is committed. We know of no prior
work that replays public experimental tables as closed-loop target
laboratories to compare acquisition, adaptation, proposal and decision rules
for microfluidic surrogates; the nearest practice is the retrospective
benchmark of acquisition functions on tabulated datasets common in materials
informatics~\cite{liang2021bo}, which the row-indexed oracle of Section~\ref{sec:fw-offline}
shares, while the sealed and pre-registered evidence tiers are taken from the
preregistration literature cited above.

\paragraph{Relation to Vepa (2026).} Closest to our work, Vepa~\cite{vepa2026}
pre-registered a validation of machine-learned sodium-cathode voltages compared with
experimentally anchored literature values and reported its failure. We share
the preregistration ethos but differ in domain (droplet microfluidics rather
than materials informatics), in the evaluated model (a frozen physics-guided,
hard-constrained design surrogate evaluated alongside the published-feature
baseline), and in the search-space endpoints (precision and recall of label-free
pruning, and retrospective target-retrieval coverage). Where Vepa pre-registered a follow-up
calibration check, we state that any new blind claim for V2 requires a
different, never-inspected dataset, and we carried out one such further
sealed test, for the published-feature baseline, on a third dataset, whose
negative outcome is reported in Section~\ref{sec:r9}.

\section{A Decision Framework for Learned Surrogates in a Self-Driving Loop}
\label{sec:framework}

A self-driving laboratory couples a learned surrogate to an instrument. The
surrogate $f_S$ is trained on a source table $S$; the instrument measures a
proposed condition $x$ in a target domain $T$ at unit cost and returns the
property $y$. Between proposals the loop updates the surrogate with what it
has measured and decides what to do next. For a surrogate in such a loop, four
decisions recur, and the evidence of this paper is organised around them:
\begin{description}[leftmargin=*,itemsep=1pt]
\item[Q1 Competence.] When is $f_S$ trustworthy on $T$, and what, if
  anything, computable before any target measurement says so?
\item[Q2 Budget.] How many target measurements does the loop need before the
  surrogate is usable on $T$, and how does that number depend on the kind of
  shift between $S$ and $T$?
\item[Q3 Acquisition.] Which conditions should be measured first?
\item[Q4 Decision.] After the first measurements, should the loop keep
  adapting the source model, stop adapting it, abandon it for a model fitted
  to the target rows alone, or escalate to a full recalibration at a cost?
\end{description}
The framework (Figure~\ref{fig:loop}) fixes the objects and quantities below so that each question is
answered by a measurable number, offline on public tables here, and is
designed to be applied prospectively in a laboratory with the same code.

\begin{figure}[htbp]
\centering\footnotesize
\begin{tikzpicture}[x=1cm,y=1cm,
  bx/.style={draw=black!65,rounded corners=2pt,align=center,inner sep=3pt,font=\footnotesize,minimum height=10mm,text width=29mm},
  qx/.style={bx,fill=blue!8},
  lp/.style={-{Latex[length=1.8mm,width=1.6mm]},line width=0.7pt,black!60}]
\node[bx,fill=black!6] (src) at (0,0) {Source table $S$\\[1pt]surrogate $f_S$};
\node[qx] (q1) at (3.6,0) {\textbf{Q1} competent?\\[1pt]{\scriptsize $r = \mathrm{error}_T(f)/$dev-ref}};
\node[qx] (acq) at (7.2,0) {\textbf{Q3} acquire $k_0$\\[1pt]first rows ($k_0 \ge 0$)};
\node[bx] (meas) at (10.8,0) {measure on $T$\\[1pt]{\scriptsize instrument\\(offline: a table row)}};
\node[bx] (adapt) at (10.8,-2.5) {adapt $f$\\[1pt]{\scriptsize warm-start / affine / scratch}};
\node[qx] (dec) at (7.2,-2.5) {\textbf{Q4} decide\\[1pt]{\scriptsize continue / stop / switch / escalate ($\lambda$)}};
\node[bx] (prop) at (3.6,-2.5) {propose next\\[1pt]condition};
\draw[lp] (src) -- (q1);
\draw[lp] (q1) -- (acq);
\draw[lp] (acq) -- (meas);
\draw[lp] (meas) -- (adapt);
\draw[lp] (adapt) -- (dec);
\draw[lp] (dec) -- (prop);
\draw[lp] (prop.south) -- ++(0,-0.38) -| ($(meas.east)+(0.5,0)$) -- (meas.east);
\node[font=\scriptsize\itshape,text=black!60,anchor=north] (looplab) at ($(dec.south)+(0,-0.42)$) {closed loop on target domain $T$};
\coordinate (mid) at (5.6,0);
\node[anchor=north,font=\scriptsize,text=black!75,text width=12.4cm,align=center] at ($(looplab.south -| mid)+(0,-0.1)$) {stop when a measured row lies within $\tau$ of $y^{*}$ (time-to-target $T^{*}$) or after $B$ measurements\\\textbf{Q2} budget $k^{*}(\varepsilon)$: the smallest $k$ whose error is at most $\varepsilon\,{=}\,1.25\times$ dev-ref};
\end{tikzpicture}
\caption{The decision framework. A surrogate trained on a source table enters a loop on a new target domain; the four questions sit at the points where the loop must decide. Offline, the instrument is replaced by the rows of a held-out tier (Section~\ref{sec:fw-offline}); prospectively, by the instrument itself (Section~\ref{sec:outlook}).}
\label{fig:loop}
\end{figure}

\subsection{Objects}
\label{sec:fw-objects}
The source $S$ is one public table or a development partition of it; the
target $T$ is a group of rows the development set never contained (a device,
a fluid pair, a surfactant chemistry, a laboratory, a lipid, a source paper or
a preparation method). The measurement oracle $O$ answers a proposed condition
with its property; in the offline instantiation it is the table itself
(Section~\ref{sec:fw-offline}), in a prospective one the instrument. A loop
policy $\pi = (\text{acquire}, \text{adapt}, \text{propose}, \text{decide})$
has four components: an acquisition rule choosing which rows to measure before
searching ($k_0$ rows; random, farthest-point, $k$-means or widest source
interval); an adaptation method mapping the $k$ measured rows to an updated
surrogate (none, two-parameter log-affine correction, warm-start residual
boosting of the source model, or a model trained from scratch on the measured
rows; Section~\ref{sec:adapt}); a proposal rule choosing the next row for a
target property (greedy nearest prediction, or a retrieval portfolio that
hedges the proposal by the source model's residual scale); and a decision
rule mapping diagnostics of the measured rows to one of the actions of Q4.

\subsection{Quantities}
\label{sec:fw-quantities}
\emph{Development reference} (dev-ref): the surrogate's grouped
cross-validation error on $S$ (grouped by the key that defines $T$ where one
exists), the error a careful developer would report without any target label.
\emph{Competence ratio}: $r = \mathrm{error}_T(f)/\text{dev-ref}$; the
surrogate is usable on $T$ when $r \le 1.25$ and replicates its development
error when $r \le 1$ (the thresholds of Section~\ref{sec:d3protocol}), and the pre-registered
hypotheses of Section~\ref{sec:hprotocol} are statements about $r$.
\emph{Diagnostics}: statistics computable from $(f_S, X_T)$ and the $k$
measured rows; label-free when $k=0$. \emph{Measurement budget}
$k^{*}(\varepsilon)$: the smallest $k \ge 1$ at which the adapted surrogate's median
error over random draws of the $k$ measured rows, on the unmeasured rows of
$T$, falls to $\varepsilon = 1.25 \times$ dev-ref, the
usable level fixed before the second sealed validation
(Section~\ref{sec:d3protocol}); ``not reached within the budget'' is a value.
\emph{Point task}: for a target property $y^{*}$ and tolerance $\tau$, the
time-to-target $T^{*}$ is the number of measurements until a measured row
lies within $\tau$ of $y^{*}$, capped at the budget $B$ plus one, and the
cumulative regret sums, over the rows measured before success, their relative
error in excess of $\tau$. \emph{Decision cost}: escalation to a full recalibration is charged
$\lambda$ measurement-equivalents, and a decision rule is judged by its cost
against the best action chosen after the fact.

\subsection{Evidence tiers and decision forms}
\label{sec:fw-evidence}
Every result carries one of five tiers, coded as in Table~\ref{tab:findings}
and shaded in Figure~\ref{fig:design}: \emph{sealed} (S; the target labels
were split off before any analysis and revealed once through an exclusive
gate), \emph{pre-registered} (P; hypotheses, thresholds and decision rules
were frozen and hashed before the data were downloaded), \emph{frozen
protocol} (F; the protocol and code were hashed before the run and the run was
repeated in a separate directory with a byte-identical artifact manifest),
\emph{post hoc} (H; analyses on revealed labels under a protocol frozen
before a single run, labelled as such wherever their numbers appear) and
\emph{retrospective} (R; every analyst choice made with the labels
available). Comparative statements use one decision form: the mean
difference against a named reference policy or model over the units (tiers or
cells), a table-level block-bootstrap interval, and the labels ``better'' or
``worse'' when the interval excludes zero, otherwise ``not distinguishable'',
with Benjamini--Hochberg control of two-sided block-bootstrap $p$-values
(twice the smaller share of resampled means on either side of zero) where
several comparisons share a question; a finding whose sign changes when every
oracle label is scaled by $1 \pm 0.2$ is marked fragile. With six tables these
intervals are the decision criterion of the frozen protocols, not an inference
to a population of tables. Both signs are reported; nothing is dropped for
being negative.

\subsection{Offline instantiation: tables as laboratories}
\label{sec:fw-offline}
No wet experiment is run. Every leave-one-group tier of the six tables with
at least twelve rows is one target domain (107 tiers in all: D1 31, D2 6,
A1 24, R2-A1 7, R2-A2 38, and the laboratory-B tier of D3); its rows are
$X_T$ with hidden labels, while the remaining rows of its table are $S$.
Proposals are rows of the tier; the oracle returns that row's label and
consumes it, so a recorded condition can be realised exactly and the
distortion between a proposal and its measurement is zero by construction.
What the replay cannot do is realise a condition the table never recorded, so
the density of a tier's sampling (median relative gap between nearest rows
$\le 0.05$: ``dense'', else ``sparse'') is recorded and every closed-loop
result is reported on all tiers and by density. Seven frozen protocols instantiate the framework:
E0 checks the simulator (sampling density, replicate noise floor, label
drift, a shuffled-label control, and floor and ceiling policies); E1 replays
the point task for 14 policies on every tier (three targets, ten seeds,
$\tau = 0.10$, $B = 20$); E2 measures $k^{*}(\varepsilon)$ by shift type; E1b
repeats E1 at a halved tolerance, a halved budget and both; E3 learns a decision
rule from two-row diagnostics by five-fold table rotation and prices
escalation at $\lambda \in \{0, 5, 10\}$; E4 repeats three policies with six
source-model families; E5 tests whether a planned budget transfers across
tables; E6 replays every policy on shuffled labels and collects every gain with
its drift and shuffled status; and E7 repeats the closed-loop questions on tail
targets at a 3\% tolerance (Sections~\ref{sec:eloop},
\ref{sec:q2}--\ref{sec:robust}).

\subsection{Map of the evidence}
\label{sec:fw-map}
Table~\ref{tab:resultsmap} lists every evidence unit with its tables, its
tier and the question it bears on. Units R1--R10 belong to the first three stages, the
sealed and pre-registered programme (Figure~\ref{fig:design}); H-J (the
few-shot meta-analysis of Section~\ref{sec:q2}), E0--E7 and E1b are the
offline closed-loop instantiation. Results keep their unit
labels and are presented by question in Section~\ref{sec:results}; the
framework-level findings they support are collected in
Section~\ref{sec:synthesis}.

\begin{table}[htbp]
\centering\footnotesize
\caption{Map of the evidence units. ``Retrospective'' means every analyst
choice was made with the labels available; ``sealed'' means labels were split
off before any analysis and revealed once through an exclusive gate;
``frozen'' means the protocol was hashed and committed before the run. Q1
competence, Q2 budget, Q3 acquisition, Q4 decision; P provenance; M method.}
\label{tab:resultsmap}
\begin{tabular}{p{0.07\linewidth}p{0.13\linewidth}p{0.26\linewidth}p{0.06\linewidth}p{0.34\linewidth}}
\toprule
Unit & Table(s) & Protocol status & Q & What it establishes \\
\midrule
R1 & D1 & retrospective re-implementation; fidelity check (Appendix~\ref{app:fidelity}) & Q1, P & the published configuration is reproduced up to one library default \\
R2 & D1 (external) & retrospective & Q1 & error grows across literature sources \\
R3 & D1 & retrospective & Q1 & nominal intervals lose coverage under shift \\
R4 & D1 & retrospective, with negative controls & Q1 & label-free pruning gains are source-level, not candidate-level \\
R5 & D2 & pre-registered, sealed, one reveal & Q1 & an unseen chemistry inside the marginal ranges fails \\
R6 & D2 & frozen protocol, retrospective after the reveal & Q4 & hard bounds remove invalid predictions but do not beat the XGBoost baseline \\
R7 & D1 & frozen protocol, two runs & Q1 & shift types order; screening is source-level \\
R8 & D1, D2 & frozen protocol, two runs & Q1, Q4 & which model and interval families survive which shift \\
R9 & D3 & pre-registered, sealed, one reveal; disclosed scoring correction & Q1 & a new laboratory fails; the support gap is visible from inputs \\
R10 & A1, R2-A1, R2-A2 & pre-registered multi-table protocol, sealed tiers, one reveal each & Q1 & no fixed shift multiplier; random-split rankings do not transfer \\
R10, constructive & revealed tiers; D2, D3 & frozen post-hoc protocols & Q2, Q3, Q4 & a few target labels recover chemistry shift; warm-start, portfolio and gates compared \\
H-J & 106 tiers of five tables & meta-analysis under a frozen protocol, two runs & Q4, Q2 & the two-row bias ratio predicts the few-shot gain; a threshold rule on it helps at $k=2$ only and fails its two-of-three criterion \\
E0 & 107 tiers of six tables & frozen protocol, two runs & M & simulator validity: density, noise floor, drift, controls \\
E1 & 107 tiers & frozen protocol, two runs & Q2, Q3, Q4 & point task: 14 loop policies against a reference \\
E2 & 107 tiers & frozen protocol, two runs & Q2 & $k^{*}(\varepsilon)$ by shift type and adaptation \\
E1b & 107 tiers & frozen protocol, two runs & Q2--Q4 & the E1 decisions at a halved tolerance, budget, or both \\
E3 & 107 tiers & frozen protocol, two runs & Q4 & a decision rule with escalation cost under table rotation \\
E4 & 107 tiers & frozen protocol, two runs & Q1, Q4 & six source-model families inside the loop; rankings \\
E5 & 107 tiers & frozen analysis protocol, two runs & Q2 & transfer of planned budgets across tables \\
E6 & 107 tiers & frozen protocol, two runs & M & shuffled-label replay of every policy; fragility of every gain \\
E7 & 107 tiers & frozen protocol, two runs & Q2--Q4 & the closed-loop questions on hard tail targets \\
\bottomrule
\end{tabular}
\end{table}

\section{Data and Provenance}
\label{sec:data}
Six public tables are used. D1, D2 and D3 carry Stage~1: D1 and D2 the
droplet-diameter evaluation, D3 the second sealed validation (and, after its
reveal, one closed-loop tier); none is pooled
with another. A1, R2-A1 and R2-A2 enter only in Stage~2 (Section~\ref{sec:r10})
and are introduced at the end of this section.

\paragraph{D1: DAFD~3.0 (OSF 938rs).} The comprehensive table has 868 rows and
the external generalisability table 64 rows from four literature sources
(Galogahi et al., $n=24$; Lashkaripour et al., $n=20$; Roumpea et al.,
$n=14$; Mazutis et al., $n=6$)~\cite{lashkaripour2024}. Eight published
inputs are used: orifice width, viscosity ratio, capillary number, flow-rate
ratio, and four normalised geometry descriptors; the target is droplet
diameter. The 868 rows contain 474 single-emulsion rows and 197
double-emulsion conditions represented by paired inner/outer rows (394 rows);
the external 64 rows contain 44 single-emulsion rows and 10 double-emulsion
conditions (20 rows). SHA-256 digests of the four downloaded workbooks are
listed in Appendix~\ref{app:hashes}. The OSF deposit is not redistributed; its
per-file redistribution terms remain to be verified before any mirror.

\paragraph{D2: Chagot et al.\ (Zenodo 7055018, CC~BY~4.0).} Two MATLAB tables
(392 and 76 rows) give 468 configurations on a fixed oval flow-focusing
junction (195~\um{} $\times$ 190~\um{}) with continuous and dispersed flow
rates, equilibrium interfacial tension, surfactant concentration and critical
micelle concentration (CMC) as inputs and droplet diameter (mm) as
output~\cite{chagot2022}. Chemistry groups are defined only by the recorded
CMC value. The row arithmetic is as follows. The complete \tox{} group
(CMC~$=3.5$~mM, 87 rows) was the sealed holdout, so the V1 development set is
the remaining $468-87=381$ rows. Two rows with CMC~$=0.60$~mM, a value not
listed among the paper's surfactants, are included there because V1 was
sealed before any quarantine rule existed. Every later analysis (V2 and the method matrix)
quarantines those two rows and never reassigns them, leaving 466 rows in six
groups (CMC 0, 2, 3.5, 4.33, 11 and 20~mM with 63, 30, 87, 15, 178 and 93
rows), of which the five non-\tox{} groups hold 379 rows.

\paragraph{D3: Smart Microfluidics liposome dataset (Zenodo 17867478,
CC~BY~4.0).} Used for the second sealed validation
(Section~\ref{sec:d3protocol}) and, after its reveal, as the laboratory-B tier
of the closed-loop benchmark (Section~\ref{sec:eloop}). The record documents microfluidic liposome
production on a micromixer chip with four lipid components (ESM, HSPC, CHOL,
DSPE-PEG2000; mg/mL), total flow rate (mL/min), flow-rate ratio and aqueous
medium (MQ or PBS) as inputs, and liposome size (nm), polydispersity index
and a formation flag as outputs~\cite{smartmicrofluidics2026}. The cleaned
table released by the authors carries no laboratory field, so we source the
data directly from their quality-checked raw files, which record it. The
development set is the 246 laboratory-A rows (Latina, operator A,
equipment A) that remain after quarantining Droplet-chip rows and rows with
missing inputs or size; the sealed set is the 57 laboratory-B rows (Rome,
operator A, equivalent equipment B). A 12-row operator-B set from laboratory A was sealed as a
secondary set outside the primary estimand. The target differs from D1 and D2
(liposome size rather than droplet diameter), which is one reason D3 results
are reported apart from them.

\paragraph{A1, R2-A1 and R2-A2: Stage~2 tables.} Three further tables were
admitted by a two-step candidate screen that required a peer-reviewed source
or its DOI repository, a licence permitting re-analysis, a directly readable
row-level table with inputs and target, traceable authors and no shared source
with D1--D3; the screen, the header-only intake manifests and every exit are
in the repository. A1 is the non-Newtonian flow-focusing droplet study of
Hormozinezhad et al.~\cite{labchip2025dimless} (ESI, CC~BY~3.0): droplet length
against flow rates, viscosities, interfacial tension and etched channel depth
and width on two glass devices, 901 rows after the binding's rule removed rows
without a measurable droplet. R2-A1 is the co-flow capillary table of Wang et
al.~\cite{wang2024coflow} (Zenodo, CC~BY~4.0): 830 rows of droplet diameter
against flow rates, fluid properties and capillary dimensions for eleven fluid
pairs on two devices; the observed flow regime is kept as a descriptor and
never used as an input. R2-A2 is the LNP Atlas of Song et
al.~\cite{lnpatlas2025} (Scientific Data; Zenodo, CC~BY~4.0), a
peer-reviewed literature compilation of lipid-nanoparticle formulations rather
than one laboratory's measurements: of its 1092 rows, 919 carry a parsable
reported size; lipid identities, molar ratios, payload type and preparation
keywords are the inputs, and the 571 rows prepared by microfluidic mixing form
the main table. The three tables are described further, with their sealed
tiers, in Section~\ref{sec:r10}.

\paragraph{Licensing boundary.} Manuscript text and original figures are
CC~BY~4.0 and code is Apache-2.0; neither relicenses the source papers or
datasets. No raw third-party data from any of the six tables are included in
the release tree.

\section{Methods}
\label{sec:methods}

\subsection{Baseline re-implementation}
\label{sec:e0}
We re-implemented the published XGBoost configuration from the method
parameters stated in~\cite{lashkaripour2024} without copying upstream code
(100 trees, squared error, learning rate 0.3, maximum depth 6, L2
regularisation 1, minimum child weight 1) and repeated 15 random 80/20 splits
of the 868-row table. Generation frequency is derived from mass conservation,
$F = 6Q_d/(\pi d^3)$, and compared with the published frequency error.
External evaluation trains on all 868 rows and predicts the 64 external rows.
Frozen library versions are recorded (XGBoost~3.3.0, scikit-learn~1.9.0,
Python~3.12.13).
\paragraph{Provenance of the baseline.} The baseline is an independent
re-implementation written from the Methods of~\cite{lashkaripour2024}:
xgboost~3.3.0 with explicit parameters (100 trees, depth 6, learning rate
0.3, L2 penalty 1, minimum child weight 1), the eight published inputs, no
input scaling, and no reuse of upstream code or of the deployed model. After
the fact, a frozen protocol compared it with the upstream repository named in
the paper's code-availability statement (\texttt{CIDARLAB/dafd-website},
commit \texttt{4902f8ce}; Zenodo 10156784 = tag v3.0.0 = commit
\texttt{d1e6296}; the digests of its training script and deployed model are
in Appendix~\ref{app:fidelity}) run as is: on the same random splits the two
give 5.268\% and 5.247\% MAPE, and the only difference is the library's
\texttt{base\_score} default (a fixed 0.5 in xgboost~1.3.3, a data-estimated
intercept in 3.3.0), worth $-0.444$ percentage points on the 64 literature
rows with no change of direction or conclusion. The upstream repository
carries no licence file, so neither its code nor its model is redistributed.
The pre-registered endpoints called ``reproduction criterion'' (V1) and
``reproduction threshold'' (D3) ask whether our re-implementation reaches a
published error; they do not refer to upstream code. The deployed upstream
model, run through the same protocol, equals the upstream default
constructor fitted with a \texttt{StandardScaler} on all 932 rows, that is
the 868 comprehensive rows together with the 64 literature rows; for that
model the literature rows are in-sample (MAPE 0.313\%), whereas our external
evaluation holds them out.

\subsection{Expanded retrospective protocol on DAFD}
\label{sec:expanded}
Because the 64-row external set had been inspected during the pilot, the
expanded DAFD analysis is retrospective and exploratory; a frozen protocol
fixed every method, threshold, endpoint and negative control before
implementation so that no outcome-driven change could be made afterwards.
Evaluation protocols are: 15 row-random splits (re-implementation check only); 15
exact-input grouped splits; 30 splits grouped by device geometry; 30 splits grouped by geometry plus viscosity ratio (geometry-plus-fluid); and full training followed by the
fixed external set. Group identifiers are connected components of the grouping
relation and the double-emulsion inner/outer pair links, so a pair never
crosses a fit/test boundary. Models are the published XGBoost configuration
(focal), random forest and histogram gradient boosting (robustness only).
Intervals compare split conformal, conformalised quantile regression (CQR) and
normalised split conformal, calibrated without external labels under both
row-random and geometry-grouped protocols; grouped calibration aggregates
non-conformity by dependency-group maximum. Applicability scores are
per-feature min--max, standardised $k$-nearest-neighbour distance
($k\in\{3,5,10,20\}$), OAS-shrinkage Mahalanobis distance, Isolation Forest,
Local Outlier Factor, one-class SVM, bootstrap-ensemble disagreement and an
out-of-fold residual-risk model; thresholds are the 0.90/0.95/0.975/0.99
quantiles of held-out calibration scores, and continuous risk--coverage curves
summarised by a tie-invariant area under the risk--coverage
curve~\cite{elyaniv2010} (AURC) are
primary. Mandatory negative controls are random
scores and random retention at matched budgets, within-source and
within-source-context physical-cluster permutations, hydraulic-diameter-only
and predicted-size-only scores, and a shuffled-label predictor. The
retrospective target-search replay uses a separate candidate pool per external
source, a fixed target grid of 30--230~\um{}, hidden outcomes during ranking,
and reports precision among answered queries only together with query coverage
and successful-target recall. Benjamini--Hochberg (BH) adjustment is applied
within each test family.

\subsection{Sealed chemistry holdout (V1)}
\label{sec:v1protocol}
The V1 protocol was written and hashed before either MATLAB file was
downloaded or opened (lock 2026-07-16T12:27:58Z). A custodian program split
the combined 468 rows by the published \tox{} CMC value, wrote a labelled
381-row development table, an input-only 87-row blind table and a separately
sealed label file with owner-only permissions, and printed only counts and
hashes. The point
model re-instantiates the XGBoost configuration described by the source
paper (100 estimators, depth 3, learning rate 0.3, seed 42) on the four
published inputs, fitted once on the development set with no blind-label
tuning; the paper's Zenodo deposit contains data only, so no upstream code
exists to reuse, and on the deposit's own 392/76 split this configuration
reaches 3.17\% test MAPE against the published 3.9\%
(Appendix~\ref{app:fidelity}; the source of the difference awaits a reading
of the paper's Methods). A 90\% split-conformal interval was
calibrated on development chemistry groups. Four applicability screens
(strict min--max, 5-NN at the 95th development percentile, OAS Mahalanobis,
Isolation Forest) and six target-search strategies over 13 fixed targets
(0.06--0.18~mm) were frozen. Pre-declared endpoints were blind MAPE compared with a
reproduction criterion ($<11.4\%$, the published semi-empirical benchmark) and
a strong-replication criterion ($\le 6.0\%$), plus four conditions under which
an applicability rule may be called beneficial. The complete pre-reveal
prediction bundle (87 predictions, 78 query--strategy pairs, 240 ranked
candidates) was hashed (digest in Appendix~\ref{app:hashes}) before the one-time
reveal gate was consumed. The two thresholds deserve a note on provenance.
The 11.4\% figure is the error of the source paper's physics-based
benchmark (the Kalli et al.\ correlation described in
Section~\ref{sec:v2protocol}) on the source paper's random 76-row test
split, and 6.0\% is a tolerance around the source paper's 3.9\% random-split
XGBoost result~\cite{chagot2022}. Both are therefore anchored to
in-distribution performance, which this paper argues overstates reliability
under shift; a development-side leave-one-chemistry-out reference in the
style later used for D3 was not pre-registered for V1 and is computed
retrospectively (Sections~\ref{sec:posthoc}, \ref{sec:r5}).

\subsection{Reveal incident and serialisation-only recovery}
\label{sec:incident}
The reveal gate was consumed at 2026-07-16T12:30:53Z. The program verified all
frozen hashes, joined predictions to labels by opaque row identifier, computed
every locked endpoint, and then failed at the final write because unanswered
target-search queries carried IEEE \texttt{NaN} fields that a strict JSON
writer correctly rejected. The intended result file was never created. A
separate recovery program, added without modifying the frozen reveal
implementation (digest in Appendix~\ref{app:hashes}), re-validated all hashes,
re-verified exact opaque-ID set equality, recomputed the same locked metrics
from the already-unsealed labels, and converted only undefined \texttt{NaN}
query fields to JSON \texttt{null}. Recovery completed at 2026-07-16T12:32:58Z;
the recovered result file's digest is listed in Appendix~\ref{app:hashes}. No model
was refit, no ranking recomputed, no metric redefined, and no pre-reveal
artifact modified. The validation therefore remains prospective with respect
to all predictions, masks, rankings and decision rules, but we describe it as
a sealed prediction followed by a disclosed serialisation-only recovery, not
as an operationally flawless one-shot reveal. A postmortem traced the root
cause to Python's permissive \texttt{allow\_nan=True} default and the absence
of a single strict artifact-boundary writer; the pre-fix expanded summary
contained 279 non-finite fields. A shared strict writer and cross-parser
regression tests were added, and all expanded outputs generated before the fix
were excluded and regenerated.

\subsection{Frozen V2 protocol}
\label{sec:v2protocol}
The V2 protocol was locked before any V2 fitting (digest in
Appendix~\ref{app:hashes}). V2 is a native-variable physics-guided
constrained regressor. The only externally supplied state is
\[
x_{\mathrm{native}}=[Q_c,\;Q_d,\;\gamma_{\mathrm{eq}},\;c,\;\mathrm{CMC},\;I_{\mathrm{no\,surf}}];
\]
a non-trainable physics layer recomputes flow ratio, capillary and Reynolds
numbers and normalised concentration from fixed device constants
($D=195$~\um, oval area $\pi\cdot195\cdot190/4$~\um$^2$, published phase
densities and viscosities). The locked semi-empirical anchor is
\[
y_{\mathrm{phys}} = \frac{d_{\mathrm{phys}}}{D} = 0.642\, r_Q^{0.188}\,(1/Ca_c)^{0.161},
\]
and the bounded output parameterisation, written as the frozen code
implements it, is
\begin{gather*}
\hat d = D\bigl[\epsilon + (1-2\epsilon)\,\sigma(z)\bigr],
\qquad
z = \mathbf{1}\{0<y_{\mathrm{phys}}<1\}\,\mathrm{logit}_\epsilon(y_{\mathrm{phys}}) + \delta_\theta(x),\\
\mathrm{logit}_\epsilon(y) = \mathrm{logit}\!\left(\mathrm{clip}_\delta\!\left(\frac{\mathrm{clip}(y,\epsilon,1-\epsilon)-\epsilon}{1-2\epsilon}\right)\right),
\end{gather*}
with $\epsilon=10^{-4}$ and $\mathrm{clip}_\delta$ a second clip to
$[\delta,1-\delta]$ at double-precision machine epsilon $\delta=2^{-52}$.
For $\epsilon<y<1-\epsilon$ the map is the exact inverse of the bounded
output map, so a zero residual returns the anchor diameter. The frozen code
clips $y$ before normalising, so an anchor-valid value within $\epsilon$ of
either bound is first sent to 0 or 1 and then, by the inner clip, to a finite
saturated logit of $\mp 36.04$ rather than $\mp\infty$. On the 466-row table
no anchor-valid row lies within $\epsilon$ of a bound: $y_{\mathrm{phys}}$
spans 0.376--0.994 and the anchor logits $-0.51$ to 5.18 (evaluated with the
frozen function in a retrospective check, Section~\ref{sec:posthoc}), so the
saturation is never exercised, and the two-stage clip is recorded as a known
limitation of the frozen code rather than corrected, because the V2 results
are frozen. The indicator is the anchor-validity mask: on rows where the
anchor equation returns $y_{\mathrm{phys}}\ge 1$ (or a non-finite value) the
anchor term is replaced by zero, the logit of the bounded midpoint
$\hat d=D/2$, and the residual network alone carries the prediction. This
fallback is a discontinuous guard: as $y_{\mathrm{phys}}$ crosses 1 the
anchor term drops from at least 9.21 (its value at
$y_{\mathrm{phys}}=1-2\epsilon$) to 0. Six of the 466 rows (1.29\%) have
$y_{\mathrm{phys}}\ge 1$ (1.030--1.384), which is why the deterministic
anchor B0, which reports $D\,y_{\mathrm{phys}}$ without the bounded map,
returns point-invalid predictions for exactly these rows
(Table~\ref{tab:v2}) while the V2 variants cannot. The residual penalty
$\lambda_p\,\mathrm{mean}(\delta_\theta^2)$ is likewise applied to
anchor-valid rows only. The residual $\delta_\theta(x)$ is a two-branch
multilayer perceptron: a native branch on the standardised transforms
$[\log Q_c,\log Q_d,\log\gamma_{\mathrm{eq}},\log(1+c),\log(1+\mathrm{CMC}),I_{\mathrm{no\,surf}}]$
(6--32--32, SiLU) and a physics branch on the standardised logarithms of the
recomputed flow ratio, $Ca_c$, $Ca_d$, $Re_c$ and $Re_d$ together with
$\log(1+c/\mathrm{CMC})$ (6--16--16, SiLU), concatenated into a 48--16--1
head. Every V2 point prediction therefore satisfies $0<\hat d<D$, and no
input reaches the model unchecked: the physics layer rejects non-positive or
non-finite flow rates and tensions, and the sigmoid is evaluated in its
numerically stable form.

\paragraph{Provenance of the anchor.} The coefficients $k=0.642$,
$\alpha=0.188$ and $\beta=0.161$ are those the source paper adopts as its
physics-based benchmark from an earlier study by the same group on the same
junction~\cite{kalli2022jcis,chagot2022}; they were not fitted on the evaluated
table, and the source paper reports their 11.4\% MAPE on its own test split.
We did not verify the surfactant set of that earlier study; the group's
contemporaneous work used DTAB, SDS and \tox{}, so B0 and the three V2
variants may carry prior information about \tox{} behaviour through three
published coefficients. This is one more reason every V2 result on \tox{} is
labelled retrospective. The sealed V1 model and the D3 model use no anchor
and are unaffected.

\paragraph{What the bound is and is not.} The bound $0<\hat d<D$ is not a
physical law. It is an empirical property of the regime that the source
dataset documents: Chagot et al.\ report that every measured droplet is
smaller than the 0.195~mm channel depth, i.e.\ the data lie in the
spherical-droplet (dripping) regime of one fixed junction; the
dripping-to-jetting transition in co-flowing streams is documented
elsewhere~\cite{utada2007}. Confined, plug or
jetting droplets with $d>D$ exist in other regimes and devices, and this
parameterisation would clip them wrongly; the frozen protocol therefore
restricts the bound to the Chagot table and forbids transferring it to DAFD,
where normalised diameter exceeds one. Throughout this paper ``physically
invalid'' for V2 means ``outside the open interval $(0,D)$ observed for this
dataset's regime'', not ``impossible in general''. A bounded output inside
$(0,D)$ is admissible for this regime; it is not thereby accurate, and the
results keep the two notions apart. The second sealed validation
(Section~\ref{sec:r9}) shows the complementary failure: an unbounded
regressor on an independent dataset returned negative diameters for 38.6\% of 57 sealed rows.

The model matrix is B0 (deterministic anchor), B1 (published four-input
XGBoost), B2 (native-input XGBoost), B3 (native-plus-physics MLP, unconstrained
output), V2-H (B3 with bounded output), V2-P (bounded residual MLP plus anchor
prior; primary) and V2-M (V2-P plus a local flow-monotonicity penalty;
sensitivity). Evaluation is nested leave-one-chemistry-out over the six main
CMC groups with inner-group selection of $\lambda_p\in\{0.01,0.1,1.0\}$ and
early-stopping epoch, five seeds $\{17,42,2024,3407,9001\}$, and equal weight
per held-out chemistry group (chemistry-macro MAPE). V2-P may be called
beneficial only if all six locked conditions hold: (1) lower chemistry-macro
MAPE than B1; (2) lower MAPE in at least four of six groups; (3) worst-group
MAPE no more than 2 percentage points worse than B1; (4) zero point and
interval physical-invalid rates; (5) retrospective search reports coverage,
recall and precision together; (6) no outer outcome used for selection.
Because \tox{} labels had already been revealed, every V2 result on \tox{} is
a post-reveal retrospective analysis and is never called a blind validation.

\subsection{Method-matrix benchmark across shift cells (protocol W3)}
\label{sec:w3protocol}
A separate frozen protocol, W3 (hashed before any run), defines seven
evaluation cells: D1 row-random 80/20 (15 seeds, reference); D1 dependency-group splits
by geometry (30 seeds); D1 comprehensive-table training with the 64-row
external literature test; D1 single emulsion to double emulsion (SE$\to$DE,
474~$\to$~394 rows) and the reverse (DE$\to$SE); D2 row-random 80/20 (15
seeds, reference); and D2 leave-one-chemistry-out (LOCO; six folds, with the
\tox{} fold labelled post-reveal retrospective). In every cell the
training rows are divided into a fit partition and a 20\% calibration
partition (by dependency group for D1, by row for D2). The three point
models are the published XGBoost configuration, a random forest and a
fixed-architecture multilayer perceptron (64--64, early stopping), all
untuned. The four interval methods at nominal 0.90 are split conformal, CQR,
group-maximum (Mondrian~\cite{vovk2005}) conformal and a 20-member cluster-bootstrap
ensemble-variance interval. The five applicability scores are 5-NN distance, OAS Mahalanobis distance, Isolation
Forest, a transductive domain classifier (balanced logistic regression
separating fit-partition inputs from the test inputs, scored as the
probability of belonging to the test batch) and a $k$-NN density ratio (5-NN
distance to the fit set divided by 5-NN distance to the other test inputs).
The controls are a random score and a size-only score,
$|s-\tilde{s}_{\mathrm{fit}}|/\mathrm{IQR}_{\mathrm{fit}}(s)$ (median $\tilde{s}_{\mathrm{fit}}$ and
interquartile range on the fit partition), where the
label-free proxy $s$ is the hydraulic diameter for D1 and the deterministic
anchor-equation diameter for D2. Binary rules threshold at the 0.95
calibration quantile; hyperparameters and software are listed in
Appendix~\ref{app:impl}. A
pre-registered candidate-level rule asks whether a score's within-test-group
AURC beats both controls, with a within-group permutation $p<0.05$ that
survives BH adjustment. Because point models here see only the fit partition, the
reference-cell errors are higher than in Section~\ref{sec:r1}; shift ratios
are therefore computed within this protocol only. Two complete runs in
separate directories produced byte-identical hash manifests.

\subsection{Uncertainty-aware comparison models on the same cells (protocol W4)}
\label{sec:w4protocol}
Because every interval in the method matrix is built around one point model,
a further frozen protocol, W4 (hashed and committed before any run), reruns the
seven cells, with identical seeds and fit/calibration partitions, for three
model families that carry their own predictive uncertainty: a quantile
regression forest (400 trees, minimum leaf 5; Meinshausen leaf
weights~\cite{meinshausen2006},
native interval between the 0.05 and 0.95 weighted quantiles, uncertainty
score the quantile gap), a Gaussian process (GP) on standardised inputs
(anisotropic RBF with a constant factor and a white-noise term, two optimiser
restarts; native interval mean $\pm 1.645\sigma$, score $\sigma$) and a deep
ensemble~\cite{lakshminarayanan2017} of five independently initialised 64--64
perceptrons on the full fit
partition (native interval mean $\pm 1.645$ times the member standard
deviation, score that deviation; no heteroscedastic likelihood head). None is
tuned per cell. Each native interval is reported alongside two
calibration-partition corrections at nominal 0.90: split conformal around the
point prediction and a conformalised version of the native interval widened by
the finite-sample quantile of its conformity score. The native uncertainty,
divided by the absolute point prediction, is also evaluated as a
selective-prediction score compared with the random, size-only and five
applicability scores of Section~\ref{sec:w3protocol}, computed in the same
cells with the same seeds. Pre-registered reading rules are descriptive: a
native interval ``survives'' a cell if its coverage is at least 0.80; the
uncertainty score counts as useful in a cell only if its AURC is below both
controls; no permutation test is run. Two complete runs must produce
byte-identical manifests.

\subsection{Covariate-shift-weighted intervals on the same cells (protocol W5)}
\label{sec:w5protocol}
The coverage losses of Sections~\ref{sec:w3protocol} and \ref{sec:w4protocol}
could have two different causes: a covariate shift, in which the inputs move
but the conditional law does not, and which conformal prediction can in
principle repair by reweighting calibration residuals with the input density
ratio~\cite{tibshirani2019}; or a conditional shift or support mismatch,
which no input-only reweighting can repair. A third frozen protocol, W5
(hashed and committed before any run), tests this on the seven cells with the
reference XGBoost model: calibration residuals are reweighted by two
label-free, transductive density-ratio estimates already computed in the
matrix, namely the odds of the class-balanced domain classifier and a
$k$-nearest-neighbour density ratio $(r_{\mathrm{fit}}/r_{\mathrm{test}})^{d}\,n_{\mathrm{fit}}/n_{\mathrm{test}}$
built from the 5-NN distances of the W3 score ($r_{\mathrm{fit}}$ and $r_{\mathrm{test}}$ the 5-NN distances to
the fit set and to the other test inputs, $d$ the number of inputs). The weighted finite-sample quantile of
Tibshirani et al.\ is used as the radius, with the test point's own mass
placed at infinity so that an interval becomes infinite when the calibration
set has too little mass near the test input. An unweighted branch must
reproduce the matrix's split-conformal coverage cell by cell. Endpoints are
coverage (an infinite interval covers), the fraction of infinite intervals,
coverage and width over finite intervals only, and the effective sample size
of the calibration weights. Pre-registered expectations are descriptive:
weighting should leave the reference cells unchanged, may raise coverage
under SE$\to$DE shift, should recover DE$\to$SE only through infinite
intervals, and should not reach 0.80 in most whole-chemistry folds, where the
\tox{} evidence already points to conditional shift.

\subsection{Second sealed validation under laboratory shift}
\label{sec:d3protocol}
The sealed protocol was instantiated for the Smart Microfluidics dataset and
hashed before the record was downloaded. After download, only file lists,
column headers, metadata dictionaries and the authors' processing logs were
inspected; the resulting parameter binding (which files, which columns,
which rows are laboratory B) was hashed and committed before a custodian
program separated development rows, sealed-set inputs and sealed-set labels
(owner-only permissions, never committed). The locked model is the sealed-V1 configuration
(XGBoost, 100 trees, depth 3, learning rate 0.3, seed 42) on the six numeric
inputs plus one-hot aqueous medium, fitted on all 246 development rows; a 90\%
split-conformal interval is calibrated on a 20\% development partition;
applicability screens are strict min--max, 5-NN, OAS Mahalanobis, Isolation
Forest and a transductive domain classifier with random and predicted-size
controls. Because no published held-out size error for this dataset was
retrieved before the lock, the success criteria are relative to a
development reference frozen before reveal: the mean 5-fold grouped
cross-validation MAPE on laboratory A. Reproduction success requires sealed
MAPE $\le 1.25\times$ this reference and strong replication $\le 1.00\times$.
Two facts were recorded from inputs alone before any fit, and one amendment
followed from them. First, every sealed row uses PBS buffer while the
development set has 231 MQ and 15 PBS rows; the protocol's candidate-level
stratum (buffer) therefore has a single level in the sealed set, and an
amendment adds main lipid (ESM or HSPC, derived from inputs) as the stratum.
Second, the strict min--max screen accepts 31.6\% of sealed rows (PEG within
the development range for 0.316 of rows, HSPC for 0.684). Prediction and
reveal programs were frozen after a dry run on a pseudo-sealed split of
laboratory A rows only; the reveal gate is a single exclusive file and was
consumed once. A scoring-definition defect found in the applicability-benefit
block after the reveal, and its disclosed correction, are reported once, in
the disclosure paragraph of Section~\ref{sec:r9}.

\subsection{Pre-registered multi-table protocol (H-A to H-E)}
\label{sec:hprotocol}
Stage~2 tests five hypotheses distilled from Stage~1 under a protocol
(\texttt{H\_PROTOCOL\_v1}) hashed and committed before any Stage~2 data file
was downloaded; a v2, which changed only the dataset list, was frozen before
the R2-A1 and R2-A2 files were downloaded. H-A: the shift multiplier
$r$ (sealed-tier MAPE divided by a grouped development reference) lies in a
fixed band $[2.2, 3.0]$ on every unseen-fluid (S2) and unseen-device or
unseen-laboratory (S3) tier. \mbox{H-B1}: a positivity-bounded perceptron (log target)
beats its unbounded raw-scale twin on the sealed tiers with no invalid
prediction; \mbox{H-B2}, which needs an anchor equation, asks whether an anchored
residual model improves on the bounded one inside the anchor's domain only.
\mbox{H-C1}: dimensionless inputs beat native inputs across held-out fluid groups;
\mbox{H-C2} compares rows that share dimensionless groups at different absolute
scale and is decided only with at least twenty such pairs. H-D: the answer rate of interval-contained retrieval tracks the
fraction of intervals narrower than the tolerance band ($|A-W|\le 0.10$). \mbox{H-E1}:
six-model rankings under random splits do not predict rankings under shift
(Spearman $\rho\le 0.5$); \mbox{H-E2}: native widths grow under shift while tracking
row error. For each admitted table a binding file fixes inputs, target and
grouping columns before any label is read; a count rule chooses each sealed
level (the level whose row count is closest to the median after the largest
level is excluded); a custodian program splits the sealed labels off; the
model configurations, intervals and tolerance are those of W3/W4; every
development-side analysis is run twice with byte-identical manifests; and each
reveal consumes one exclusive gate. Multi-table success conditions need at
least two admitted tables. The protocols, bindings, programs with SHA-256
sidecars and the deviation register are in the repository
(Section~\ref{sec:repro}).

\subsection{Post-hoc analyses (not pre-registered)}
\label{sec:posthoc}
Five blocks of analyses were added after both Stage~1 reveals to check
mechanisms suggested by the results. None is part of a frozen protocol, every label they touch was already open, and
their outputs are hashed in separate directories with their own environment
records. (i)~A retrospective development reference for V1:
the locked V1 model is refitted leave-one-chemistry-out over the five
development CMC groups of the V1 development table (the \tox{} rows are never
used; the two rows with CMC 0.60~mM stay on the training side), giving the reference that the D3-style criterion would have used.
(ii)~Alternative point metrics for the D3 sealed set from the revealed scored
rows: symmetric MAPE, median APE, absolute log-ratio error over positive
predictions, and errors within quartile bins of observed size. (iii)~A
log-target baseline for D3: the locked configuration and inputs fitted to
$\log(\mathrm{size})$ on the 246 development rows, with the same grouped
five-fold development reference as the frozen model, whose folds and
predictions it reproduces exactly before the transform is applied, then
scored on laboratory B. (iv)~Metric robustness for the method matrix: median
APE, symmetric MAPE and log-ratio error for every cell from the canonical
per-row predictions, ratios to the reference cell under each metric,
row-bootstrap 95\% intervals (2000 resamples) for the single-split cells,
size-quartile bins for the two device-type cells, and the observed-size
distributions of single- and double-emulsion rows. Bins are quartiles of
observed size, a rule fixed before the values were seen. (v)~Mechanism
checks: the anchor-to-logit map of Section~\ref{sec:v2protocol} evaluated
with the frozen V2 function on the 466 rows; the signed error direction per
size quartile and the prediction ranges by model family, in micrometres and
in model-response space, for the two device-type cells; and the per-context
retention of the binary rules behind the within-context permutation of
Section~\ref{sec:r7}; the held-out interval widths of the V2 suite; and the
native interval widths of the uncertainty-aware families on internal versus
external rows. Apart from the refits of (i) and (iii) on development rows,
these read canonical artifacts only; none changes a frozen number. Every number from these analyses is marked post hoc
where it appears.

\subsection{Offline closed-loop benchmark (protocols E0--E7)}
\label{sec:eloop}
\begingroup\emergencystretch=3em\relax
Seven frozen protocols instantiate the framework of Section~\ref{sec:fw-offline} on the
107 tiers: the closed-loop protocol for E0--E2 (revision 1.1; \hashtrunc{022aa19d}{d6fb5f}),
the difficulty-sensitivity protocol E1b (\hashtrunc{7fa2a69c}{826c8e}), the decision-policy
protocol E3 (revision 1.1; \hashtrunc{71166fca}{c1e310}), the model-family protocol E4
(\hashtrunc{0b0b68d4}{7c59ce}), the budget-transfer protocol E5 (\hashtrunc{52fd522e}{34e1bc}),
the negative-control protocol E6 (revision 1.1; \hashtrunc{d28896ae}{2b8a06}) and the
hard-target protocol E7 (revision 1.1; \hashtrunc{130a1509}{f71425}). The runner and analyzer of E0--E2 were
hashed before the first counted run (runner \hashtrunc{645b73b2}{e13776}); every later protocol is run by a
wrapper script that imports that runner unchanged and overrides named constants, adds policies or
replaces a diagnostic function (E-D5; the four-row diagnostics of E7), so the E1 code keeps its recorded hashes, and E5 analyses frozen outputs without new loops.
Every protocol fixes its thresholds and decision rules before the data, every counted run is
repeated in a separate directory and compared file by file, smoke runs with reduced sizes are not
counted, and the registered deviations E-D1--E-D13 are listed in Section~\ref{sec:deviations}.\par\endgroup

\paragraph{Tiers and source models.} The target domains are the 106 leave-one-group tiers of the
few-shot meta-analysis (D1 31, D2 6, A1 24, R2-A1 7,
R2-A2 38) and the D3 laboratory-B tier, 107 in all, each with its table's
source constructor (defined below). By shift type they are 33 chemistry, 30 device, 29 laboratory (three
literature sources of D1, 25 source papers of R2-A2 and laboratory B of D3), 11 composition, two
emulsion-type and two preparation tiers. Per tier, computed once before any loop: the source
model, its grouped out-of-fold predictions on the source rows (giving dev-ref and $q_{\mathrm{src}}$, the
90th percentile of the absolute log residual), a quantile forest's 90\% interval width (log scale) on every tier row,
and the tier inputs standardised by the source mean and standard deviation.

\paragraph{Source models and adaptation methods.}\label{sec:adapt} Each table keeps its locked XGBoost
constructor: for D1 the reproduced DAFD configuration (100 trees, depth 6, learning rate 0.3), for D2 the
four-input B1 configuration and for A1, R2-A1, R2-A2 and D3 the configuration of the multi-table protocol
(both 100 trees, depth 3, learning rate 0.3), fitted to the logarithm of the target for A1, R2-A2 and D3 and to
the raw target for D1, D2 and R2-A1: D1 and D2 as in their original protocols, A1, R2-A1 and R2-A2 as
fixed by the multi-table protocol, and D3 as in the log-target baseline of Section~\ref{sec:posthoc},
item (iii); the sealed validation of D3 (Section~\ref{sec:d3protocol}) used the raw scale. With $k$ measured rows and the source
prediction $f$, clipped at $10^{-6}$ before any logarithm, the adaptation methods are: none ($f$ unchanged);
log-affine, the least-squares fit of $\log y = a \log f + b$ on the measured rows (a shift of $\log f$ when
$k = 1$ or the measured predictions coincide); warm-start residual boosting, an XGBoost model (100 trees, depth 2,
learning rate 0.1) fitted to $\log y$ on the measured rows and the source rows, the source rows weighted 0.1, with
$\log f$ as base margin; and scratch, an XGBoost model (100 trees, depth 3, learning rate 0.3) fitted on the
measured rows alone, on the table's target scale, from the third measured row on, the source prediction being
used before that.

\paragraph{E0: simulator validity.} Four checks precede the loop. (i)~Sampling density: the median
relative gap between a tier row's label and that of its nearest neighbour in standardised input
space; tiers at or below 0.05 are labelled dense, the others sparse. The first frozen
protocol used this gap as an eligibility gate. The uncounted smoke run showed that the gate would
have excluded most of its tiers, among them the D3 laboratory tier, although with row-indexed
proposals the gap measures how densely a tier samples its inputs rather than how far the oracle's
answer is from the request; revision 1.1, registered as deviation E-D1 on the same day and before any
counted run, made the gap a descriptive label, and the E1b protocol pre-specifies the recomputation
of every decision on the dense tiers alone, which is reported beside the all-tier decision.
(ii)~Noise floor: the pooled relative standard deviation of rows with identical inputs, with the rule
that no policy is credited with an error below it (not applied in the E2 results; deviation E-D13,
Section~\ref{sec:e2}). (iii)~Drift: every oracle label is multiplied by $1 + \delta$
and the target is expressed in the drifted scale (deviation E-D2); $\delta = \pm 0.2$ is the
pre-registered fragility check, $\pm 0.5$ descriptive. (iv)~Controls: a shuffled-label oracle on
which the reference policy must not be distinguishable from the floor policy (the simulator is
accepted when the block interval of their mean $T^{*}$ difference contains zero), the floor policy
without adaptation, and a ceiling policy that refits the source constructor on every tier label.

\paragraph{E1: point task.} Three targets per tier (the labels nearest the 20th, 50th and 80th
label percentiles, so every target is reachable), relative tolerance $\tau = 0.10$, budget
$B = 20$; a loop succeeds at the first measured row whose label lies within $\tau$ of the target,
$T^{*}$ is that step or $B + 1$, and the cumulative regret sums, over the rows
measured before success, the relative error in excess of $\tau$. Phase A measures $k_0$ rows chosen by the acquisition rule; phase B adapts
the surrogate with all measured rows, proposes one unmeasured row and measures it. The 14
policies of Table~\ref{tab:e1policy} are run with 10 seeds per target, with the drift and
control loops (80250 loops per run). Decision form: the tier value of a policy is its mean
$T^{*}$; $\Delta$ is the mean over tiers of the difference to the reference policy P1 (two random
rows, warm-start, greedy proposal), with a table-level block-bootstrap interval; ``better'' or
``worse'' when the interval excludes zero, Benjamini--Hochberg over the 13 non-reference
policies; a verdict is fragile when the sign of $\Delta$ changes under $\delta = \pm 0.2$. The
pre-registered questions are acquisition (P4, P5, P6 against P1; P10, P11 against P9, where P10 also
uses five phase-A rows), the number of calibration rows (P2, P3 against P1), the adaptation method
(P7, P8, P12 against P1) and the proposal rule (P9 against P1, P10 against P3, which also changes the
acquisition rule, and P11 against P6).

\paragraph{E2: budget curves.} For every tier and $k \in \{1, 2, 3, 5, 8, 10, 15, 20\}$ with at
least ten unmeasured rows, 100 random draws of $k$ rows are adapted by the log-affine,
warm-start and scratch methods and scored by the MAPE on the unmeasured rows; $k^{*}(\varepsilon)$ is the
smallest $k$ whose median MAPE over draws is at most $\varepsilon = 1.25 \times$ dev-ref, and ``not
reached within 20'' is a value. Reported per shift type and density label as the median and the
90th percentile over tiers; descriptive, with no threshold beyond $\varepsilon$.

\paragraph{E1b: difficulty.} The E1 design is repeated for $(\tau, B) \in \{(0.05, 20), (0.10, 10),
(0.05, 10)\}$ with the same policies, targets, seeds, drift check and control; the budget curves and
the $\pm 0.5$ runs are dropped. A finding of E1 holds in a cell when the same verdict form returns
it; it is graded ``robust to difficulty'' when it holds in all three cells, ``mostly robust'' at two,
and ``specific to the E1 setting'' otherwise; the grade is appended to the finding, which is neither
dropped nor upgraded by it.

\paragraph{E3: decision rule.} Four arms share the same two random phase-A rows per tier and seed:
continue (warm-start), none (unadapted source predictions), switch (the unadapted proposal at $k = 2$
and, from the third row, a model trained on the measured rows alone) and escalate (the ceiling model
after phase A) at a cost $\lambda \in \{0, 5, 10\}$ measurement-equivalents added to
$T^{*}$. Four diagnostics are computed from the two rows and the pre-computed source quantities: the
bias ratio $|\bar r| / \mathrm{SD}(r)$ of the two log residuals (the H-J statistic), the mean absolute
log residual relative to $q_{\mathrm{src}}$, the tier's mean interval width relative to $q_{\mathrm{src}}$
(label-free) and the fraction of tier rows with any standardised input beyond three source standard
deviations (label-free; revision 1.1 replaced an unbounded norm that near-constant source columns
dominated, deviation E-D5). The rule family is one diagnostic, one threshold at a training decile and
one action on each side, plus the four constant rules; it is learned on four tables and applied to
the fifth (the D3 tier is held out with D2 and never trained on). The rule holds at $\lambda$ when its
held-out cost is below both ``always continue'' and ``never adapt'' with block intervals excluding
zero; E3 is positive when it holds at two of the three $\lambda$ levels. Also reported: the headroom
of a perfect per-unit decision over always continuing, the fold rules, and the rule's cost under
$\delta = \pm 0.2$ with the fold rules learned at $\delta = 0$. The H-J threshold of P12 is one member
of this family and is evaluated as such.

\paragraph{E4: source-model families.} Six families serve as the source model of the loop, each
fitted on the tier's source rows on the table's target scale with seed 42: the table's XGBoost
constructor of E1, a random forest (400 trees, minimum leaf 2), a two-layer perceptron (64--64,
early stopping, standardised target), a quantile forest (point prediction), a Gaussian process with
an automatic-relevance kernel (mean prediction) and a five-member deep ensemble (mean prediction).
Inside the loop the family's predictions are the base margin of the same warm-start adaptation; the
family is not refitted. Policies P0, P1 and P2 are replayed for every family with the E1 targets,
seeds, tolerance, budget and drift check. Decision form: the tier-mean difference to XGBoost under
P2 with table-level block intervals and Benjamini--Hochberg control over the five other families; the
same difference under P0 and P1 is descriptive.
The ranking question is answered per tier by Kendall's $\tau_b$ between the families' ranking by
table-level random-split error (5-fold, row-random) and their ranking by loop $T^{*}$ under P2; the
random-split ranking ``predicts'' the loop ranking when the mean $\tau_b$ is positive with a block
interval excluding zero. The zero-shot ranking on the tier (which needs the tier labels) is reported
beside it as a label-informed reference.

\paragraph{E5: transfer of planned budgets.} E5 runs no loop: it takes the warm-start budget
$k^{*}(\varepsilon)$ of every tier from E2 and asks whether a planning rule learned on other tables
gives a useful budget for a new one. A rule maps pre-measurement features to a planned budget
$\hat B$, the 80th percentile of $k^{*}$ over the training tiers sharing the feature, rounded up and
capped at the budget of 20: always the full
budget (R0), shift type (R1), shift type and density (R2), shift type and the label-free support
diagnostic (R3), and shift type and the two-row bias ratio (R4). A plan costs the rows measured beyond
need when $\hat B \ge k^{*}$ and the full budget of 20 when $\hat B < k^{*}$ (the surrogate is not usable
when the loop stops); tiers that never reach $\varepsilon$ are reported separately. Each rule is learned
on four tables and applied to the fifth (D3 with D2), and compared with the same rule learned on the
held-out table itself; a rule transfers when this transfer regret has a block interval containing
or below zero and the transferred rule beats R0 with an interval excluding zero.

\paragraph{E6: negative controls.} All 14 E1 policies are replayed on labels permuted within each
tier, with three targets (the permuted labels nearest the E1 targets) and seeds 0--4 (the ceiling, refitted on the permuted labels, is the positive control). A policy that is
``better'' than the reference under permutation is an artefact candidate. Every ``better'' verdict of
E1, E1b, E3 and E4 (under P2) is then listed with its sign under the $\pm 20\%$ drift and its shuffled verdict,
and classified as confirmed (neither fragile nor an artefact), fragile, artefact or positive control.
The analytic chance level of uniformly random proposals is reported beside every policy.

\paragraph{E7: hard targets.} E7 was designed after E1, E3 and E6 had returned, to ask whether
their answers to Q3 and Q4 hold when targets are hard to hit by chance. Targets are the labels nearest the 5th, 10th, 90th and 95th percentiles
of each tier, the tolerance is 3\% and the budget 20. A unit (tier and target) is hard when at most
10\% of the tier's rows lie within the tolerance of the target; the primary analysis uses the hard
units, and units whose target lies outside the source table's label range are flagged as
extrapolative. The E1 policies are replayed together with an interval-aware proposal (P13: start at
once and propose the row minimising $\max(0, |\log \hat f - \log y^{*}| - w/2)$, with $w$ the row's 90\%
source quantile-forest width on the log scale, so a covering interval counts as distance zero; ties go
to the smaller $|\log \hat f - \log y^{*}|$) and four decision arms after four random rows (continue, stop
adapting, switch to scratch, escalate at cost $\lambda$), whose diagnostics now use four residuals and
add their sign consistency. A shuffled-label replay of every non-arm policy is built in, so every
``better'' verdict carries its own artefact check; the decision forms of E1 and E3 apply.

\subsection{Deviation registers}
\label{sec:deviations}
Implementation defects and protocol amendments found by code,
protocol-to-code, statistical and reproducibility checks are recorded in
numbered registers that are part of the released repository rather than
reproduced here: D01--D19 for the expanded DAFD and V2 suites
(\texttt{PROTOCOL\_DEVIATION\_REGISTER.md}, hashed in the reproducibility
manifest), W3-C1 and W3-D1 for the method matrix, and D3-B1 to D3-R1 for the
second sealed validation, and E-D1 to E-D13 for the closed-loop benchmark
(\texttt{E\_DEVIATION\_REGISTER.md}: the density label that replaced the first
protocol's eligibility gate, the drifted-scale target, an affine overflow in
six budget rows, the wrapper-script design with one reverted exploratory edit,
the bounded support diagnostic of E3, the ceiling as positive control of E6,
the scheduling and host of the E4 runs, the wall-clock column of E4, the
multiplicity family of E7, tie handling in two descriptive statistics, and the
noise-floor rule left unapplied in E2). For the expanded and V2 suites only the run pair
produced after the last registered correction is used in this paper.

\subsection{Statistical reporting}
\label{sec:stats}
Four external sources and six chemistry groups are too few for
population-level inference. Source- and cluster-level bootstrap intervals are
descriptive, and the table-level block intervals of the closed-loop protocols
are their decision criterion rather than population inference
(Section~\ref{sec:fw-evidence}); row-level Wilson intervals are conditional summaries;
seed standard deviations are algorithmic sensitivity, not sampling
uncertainty; $R^2$ is labelled unstable for small or narrow-range groups and
is never used alone to support a conclusion; BH-adjusted and unadjusted
$p$-values are reported together, with effect sizes taking precedence.

\section{Results by Question}
\label{sec:results}
Results keep the unit labels R1--R10 of the first three stages
(Section~\ref{sec:intro}) and are grouped by the question of
Section~\ref{sec:framework} they bear on; Table~\ref{tab:resultsmap} maps
every unit to its dataset, protocol status and question. Section~\ref{sec:q1}
collects the evidence on competence under shift (R1--R5 and R7--R10); Section~\ref{sec:q2} the measurement budget (the
constructive experiments of R10, H-J, E2, E1 and E5); Section~\ref{sec:q3} the
acquisition order (E1); Section~\ref{sec:q4} the decision after the first
measurements (R6, R8, H-J, E1, E3, E4) together with the robustness of the
closed-loop findings to drift and to difficulty (E1b); Section~\ref{sec:robust} subjects every
closed-loop gain to negative controls and to hard targets (E6, E7).

\subsection{Q1 --- When is the surrogate competent?}
\label{sec:q1}
The units below share one reading: the development reference of a surrogate
does not carry over to a target group its development set never contained,
by a factor that depends on the kind of shift, and no statistic computable
from inputs alone identifies the rows on which it fails. Three numbers set the
scale: the re-implemented baseline's diameter error roughly doubles from random
splits to the external literature sources (10.97\% MAPE), its nominal 90\%
conformal intervals cover 55.1\% of external rows, and the frozen model's
intervals cover 44.8\% of the sealed \tox{} rows.

\subsubsection{R1: Baseline re-implementation and fidelity check}
\label{sec:r1}
Table~\ref{tab:repro} compares our re-implementation with the published figures.
Random-split diameter MAPE is $5.268\% \pm 0.421\%$ (published 5.4\%) with
MAE $4.502$~\um{} and $R^2=0.960$; mass-conservation frequency MAPE is
$16.697\% \pm 1.819\%$ (published 16.6\%). The re-implementation is within the
published tolerances, so the degradation reported below is not an artifact of
a weak baseline. Appendix~\ref{app:fidelity} reports the fidelity check
against the upstream code run as is: 5.268\% versus 5.247\% on the same
random splits.

\begin{table}[t]
\centering\small
\caption{Baseline re-implementation on 15 random 80/20 splits of the 868-row
DAFD~3.0 comprehensive table (mean $\pm$ SD over splits).}
\label{tab:repro}
\begin{tabular}{lrr}
\toprule
Metric & Published~\cite{lashkaripour2024} & This re-implementation \\
\midrule
Diameter MAPE & 5.4\% & $5.268\% \pm 0.421\%$ \\
Diameter MAE & --- & $4.502 \pm 0.529$~\um \\
Diameter $R^2$ & --- & $0.960 \pm 0.010$ \\
Frequency MAPE (mass conservation) & 16.6\% & $16.697\% \pm 1.819\%$ \\
\bottomrule
\end{tabular}
\end{table}

\subsubsection{R2: Cross-literature degradation}
\label{sec:r2}
Across the 15 split models, external MAPE is $10.971\% \pm 0.552\%$; a single
model trained on all 868 rows gives external MAPE 11.258\%, MAE 12.227~\um{}
and $R^2=0.917$. Table~\ref{tab:sources} gives the per-source breakdown. The
Mazutis source ($n=6$) has $R^2=-10.18$ because its diameters span a very
narrow range; we report it for completeness and draw no conclusion from
small-sample $R^2$ alone. H1 is supported.

\begin{table}[t]
\centering\small
\caption{External literature set after training on all 868 comprehensive rows. $R^2$ for $n\le 14$ sources is unstable and not interpreted alone.}
\label{tab:sources}
\begin{tabular}{lrrrr}
\toprule
Source & $n$ & MAPE & MAE (\um) & $R^2$ \\
\midrule
Galogahi et al. & 24 & 8.35\% & 10.95 & 0.812 \\
Lashkaripour et al. & 20 & 8.78\% & 3.77 & 0.724 \\
Mazutis et al. & 6 & 19.41\% & 8.53 & $-10.18$ \\
Roumpea et al. & 14 & 16.29\% & 28.08 & 0.272 \\
\midrule
All external & 64 & 11.26\% & 12.23 & 0.917 \\
\bottomrule
\end{tabular}
\end{table}

\subsubsection{R3: Coverage loss of nominal 90\% intervals}
\label{sec:r3}
\begin{figure}[t]
\centering
\includegraphics[width=0.8\linewidth]{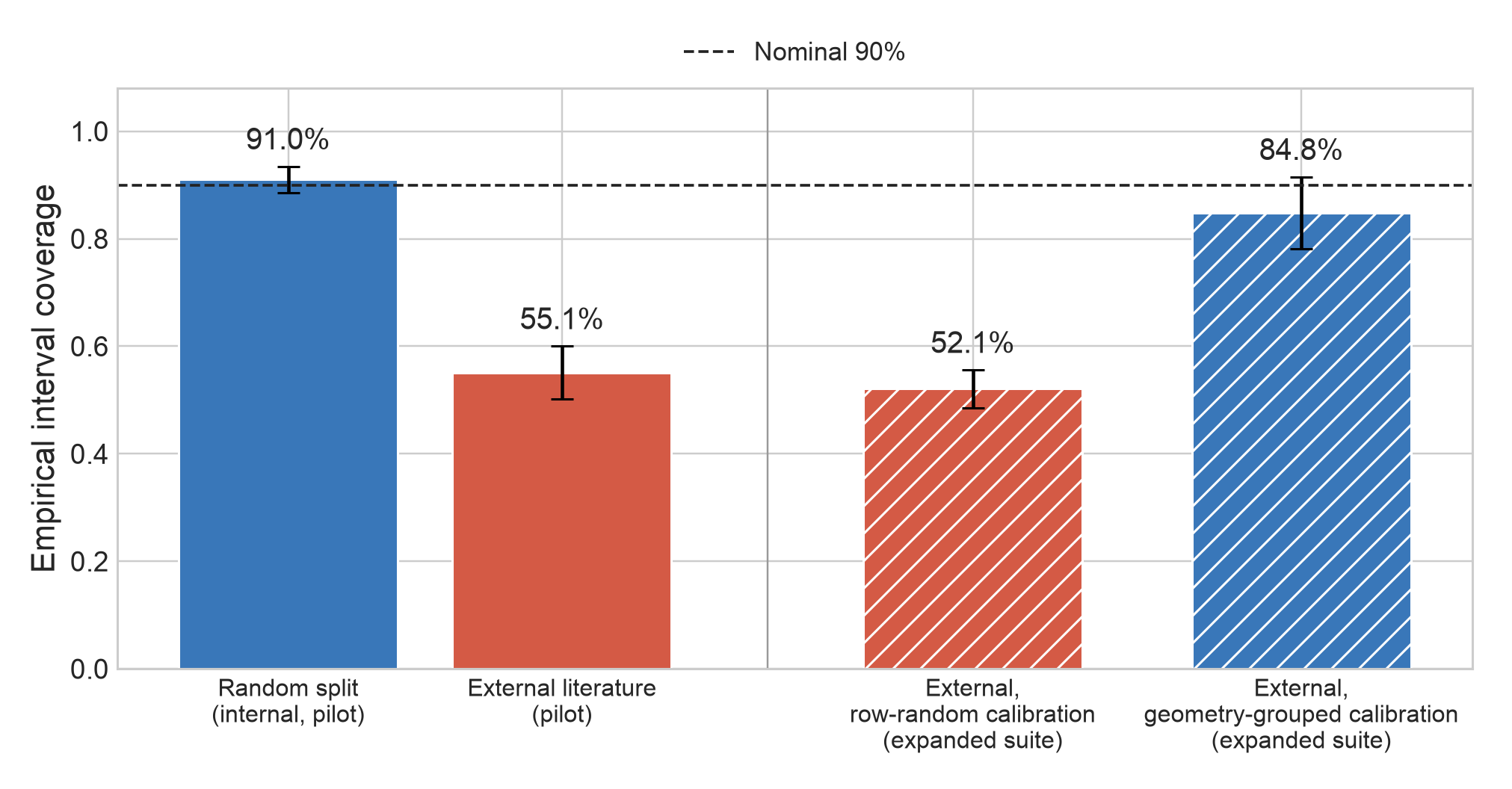}
\caption{Nominal 90\% CQR coverage. Left pair: internal random test rows
versus the external literature set over 15 runs of the pilot analysis.
Right pair: external coverage in the expanded suite under row-random and
geometry-grouped calibration (15 seeds each; Section~\ref{sec:r7}). Error
bars are one standard deviation over runs. Geometry-grouped calibration
recovers 84.8\% coverage at a mean interval width of 89.8~\um{} versus
34.1~\um.}
\label{fig:cqr}
\end{figure}

With CQR calibrated on random internal splits (15 runs of 60/20/20), observed
coverage is $0.9103 \pm 0.0244$ internally (mean width 28.95~\um) and
$0.5510 \pm 0.0492$ on the external set (mean width 37.16~\um;
Figure~\ref{fig:cqr}). This is not a failure of conformal theory: the random
calibration set and the external literature rows are not exchangeable. H2 is
supported. Section~\ref{sec:r7} shows that geometry-grouped calibration
recovers much of the external coverage at the cost of wider intervals
(89.8~\um{} versus 34.1~\um).

\subsubsection{R4: Label-free applicability pruning (pilot and canonical)}
\label{sec:r4}
\begin{figure}[t]
\centering
\includegraphics[width=0.75\linewidth]{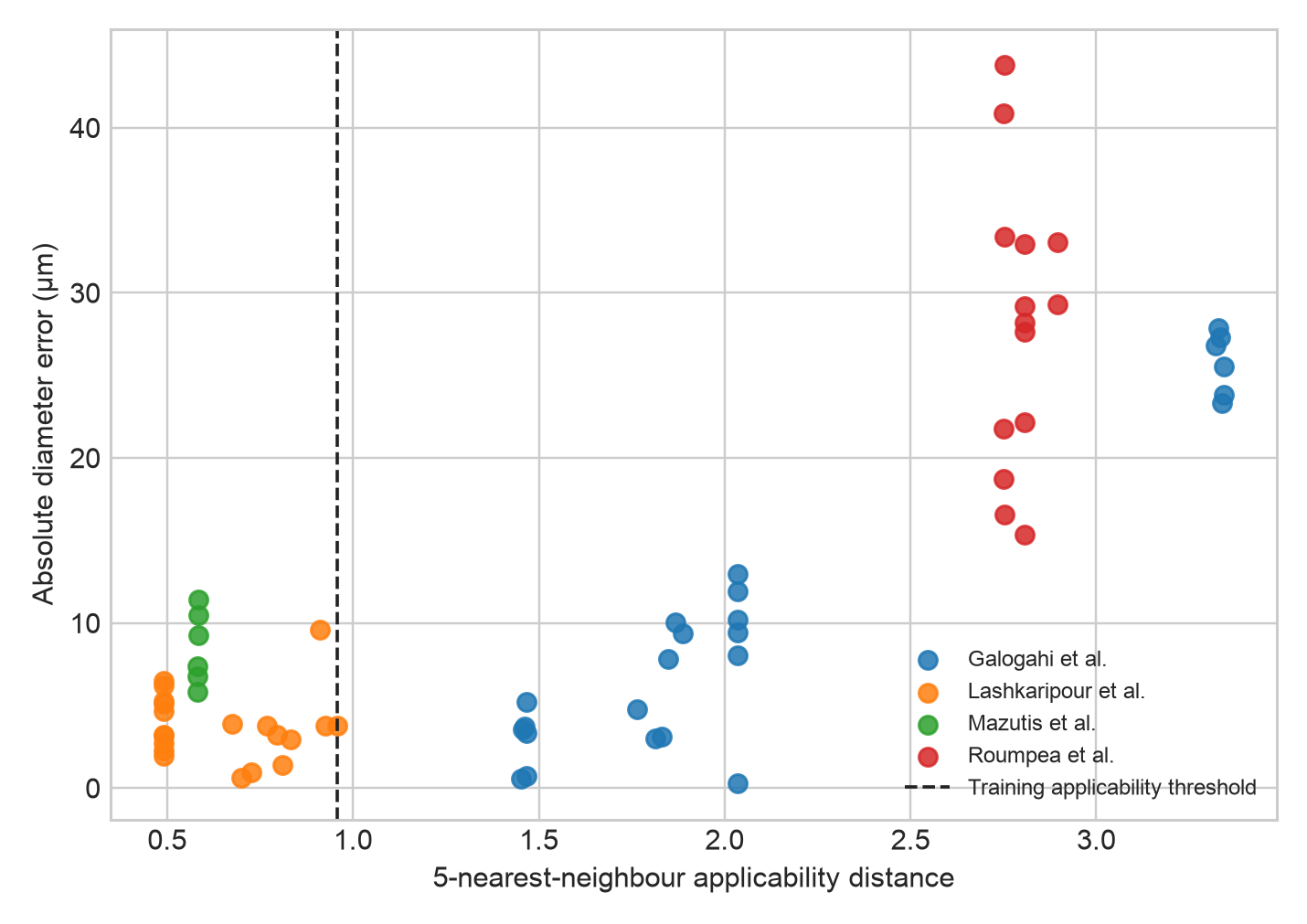}
\caption{Pilot rule: standardised 5-NN distance to the training inputs versus
absolute external error, with the 95th-percentile threshold.}
\label{fig:distance}
\end{figure}

\begin{figure}[t]
\centering
\includegraphics[width=0.75\linewidth]{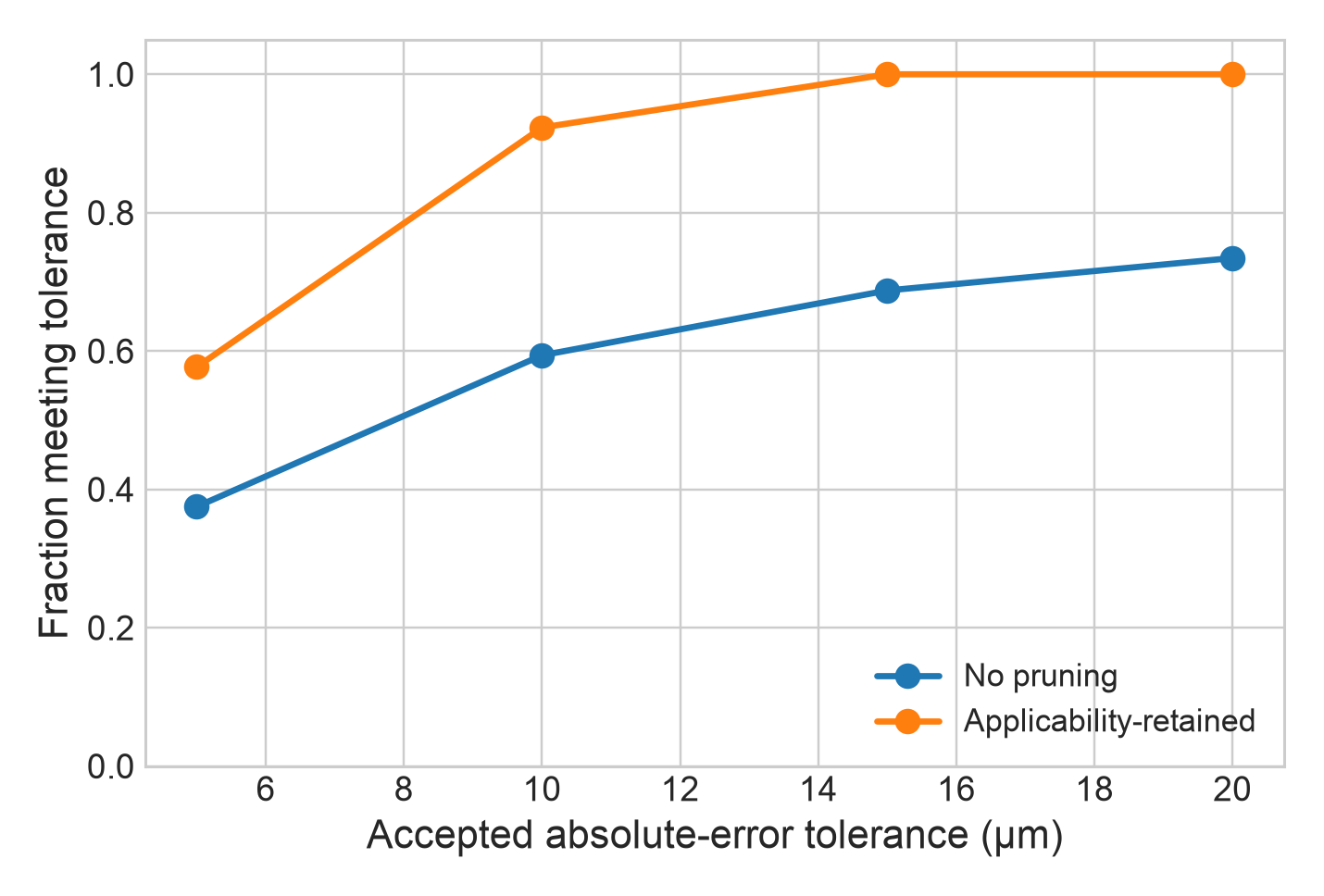}
\caption{Pilot rule: fraction of candidates within tolerance before and after
pruning. Precision rises while recall falls; both must be read together.}
\label{fig:pruning}
\end{figure}

\paragraph{Pilot rule.} Using only the 868 training inputs, we standardised the
eight features, computed leave-one-out 5-NN distances and took their 95th
percentile (0.9572) as the threshold. Of 64 external candidates, 26 were
retained and 38 pruned (pruning rate 59.375\%, a $2.46\times$ reduction).
Distance and absolute error correlate (Spearman $\rho=0.718$,
$p=2.46\times10^{-11}$, row-naive; Figure~\ref{fig:distance}). Retained
candidates have MAE 4.87~\um{} versus 17.26~\um{} for pruned ones. At a
10~\um{} tolerance, 38 of all 64 external candidates succeed (59.4\%); among
the 26 retained candidates, 24 succeed (92.3\%); the recall of successful
candidates is therefore 24/38 = 63.2\%. At 15 and 20~\um{} the retained
precision is 1.0 with recall 0.591 and 0.553. The denominators differ and
must not be conflated: 59.4\% is over all external candidates, 92.3\% over
retained candidates. Retention is entirely blocked by source (Galogahi 0/24,
Roumpea 0/14, Lashkaripour 20/20, Mazutis 6/6), so the pilot rule removes
whole literature sources rather than individual candidates
(Figure~\ref{fig:pruning}).

\paragraph{Canonical rule.} In the frozen expanded suite
(Section~\ref{sec:expanded}), thresholds are calibrated on a dependency-grouped
held-out calibration partition (104 single-emulsion rows) rather than on
leave-one-out training scores, and the 5-NN threshold becomes 1.5647. The rule
then retains 32 of 64 candidates (Galogahi 6/24, Lashkaripour 20/20, Mazutis
6/6, Roumpea 0/14). At the 10~\um{} tolerance its precision is 93.8\% (30/32)
and its recall 78.9\% (30/38); at 10\% relative error its precision is 62.5\%
and recall 71.4\%, relative to a full-pool success rate of 43.8\% (28/64).
Table~\ref{tab:pruning} reports both rules side by side. H3 is supported in
its precision clause, but the dependency-aware controls in
Section~\ref{sec:r7} qualify the mechanism.

\begin{table}[t]
\centering\small
\caption{Label-free 5-NN pruning of the 64 external candidates: pilot rule
(leave-one-out training threshold) and canonical rule (dependency-grouped
calibration threshold). Precision is over retained candidates; recall is over
all successful candidates in the full pool.}
\label{tab:pruning}
\resizebox{\linewidth}{!}{%
\begin{tabular}{lrr}
\toprule
 & Pilot (LOO q95) & Canonical (grouped q95) \\
\midrule
Threshold (standardised distance) & 0.9572 & 1.5647 \\
Retained / pruned & 26 / 38 & 32 / 32 \\
Retention by source (G/L/M/R) & 0/24, 20/20, 6/6, 0/14 & 6/24, 20/20, 6/6, 0/14 \\
Full-pool success, 10~\um & 59.4\% (38/64) & 59.4\% (38/64) \\
Retained precision, 10~\um & 92.3\% (24/26) & 93.8\% (30/32) \\
Successful-candidate recall, 10~\um & 63.2\% (24/38) & 78.9\% (30/38) \\
Retained precision / recall, 10\% relative & --- & 62.5\% / 71.4\% \\
\bottomrule
\end{tabular}}
\end{table}

\subsubsection{R5: Sealed \tox{} validation --- a negative result}
\label{sec:r5}
\begin{figure}[t]
\centering
\includegraphics[width=\linewidth]{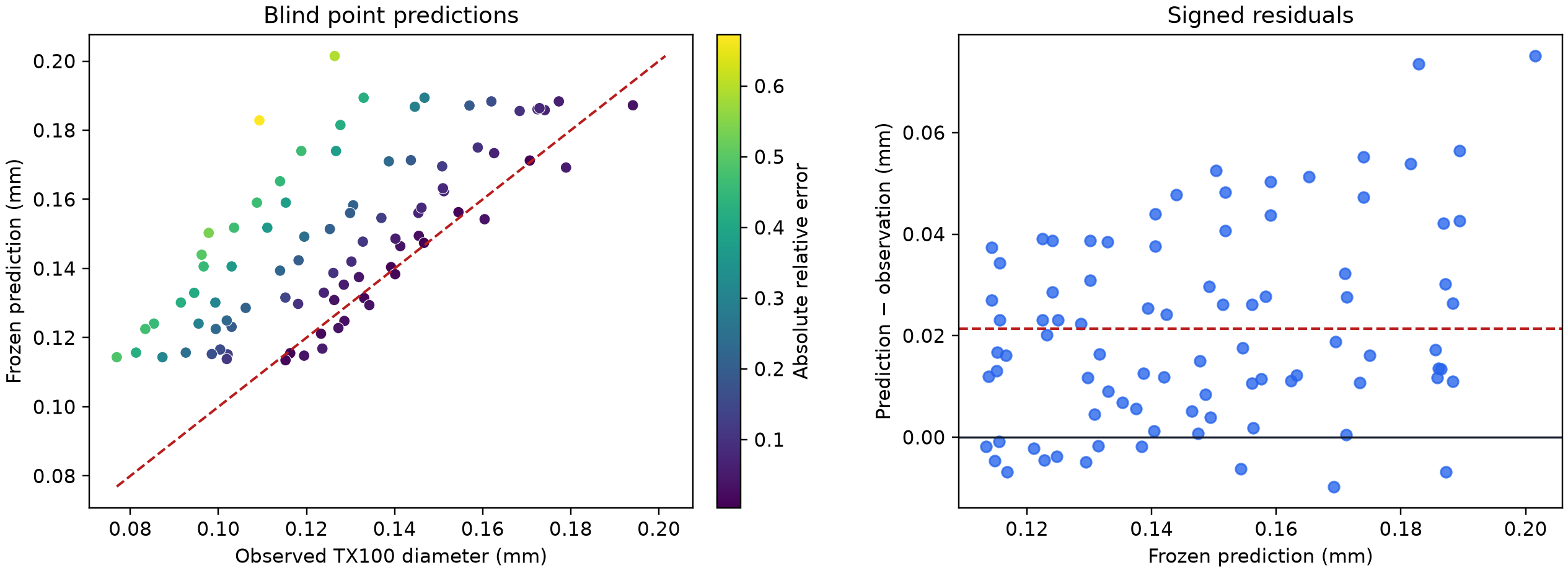}
\caption{Negative result. Left: frozen V1 predictions versus observed \tox{}
diameters (87 rows), coloured by absolute relative error. Right: signed
residuals versus prediction; the dashed line is the mean bias
(0.0213~mm).}
\label{fig:v1diag}
\end{figure}

R1--R4 are retrospective: every analyst choice was made with the D1 labels
available, and the shifts they measure are between rows of one table. R5 asks
whether the same pipeline survives when an entire surfactant chemistry of a
second table is sealed before any label is seen. Table~\ref{tab:v1} lists the headline locked endpoints; the others are given in the text. The frozen point model reached
a blind MAPE of 19.8436\% on the 87 \tox{} rows, failing both pre-declared
criteria (reproduction $<11.4\%$: false; strong replication $\le 6\%$: false),
with MAE 0.022615~mm (22.6~\um), RMSE 0.028681~mm (28.7~\um) and
$R^2=-0.2436$. The 90\% split-conformal interval covered 0.4483 of rows
(mean width 0.03281~mm, 32.8~\um). A
post-reveal descriptive analysis found over-prediction on 85.06\% of rows with
a signed relative bias of 0.1895; conditional on the 87 rows, a row bootstrap
places MAPE in $[16.4\%, 23.5\%]$ and the Wilson interval for coverage in
$[0.348, 0.553]$. The strict min--max screen accepted all 87 rows
(candidate reduction 0.0): the failure is not a marginal-range violation but
a conditional shift: the four recorded predictors overlap training support
while their relationship to diameter changes with the unseen surfactant
(Figure~\ref{fig:v1diag}).

\begin{figure}[t]
\centering
\includegraphics[width=0.8\linewidth]{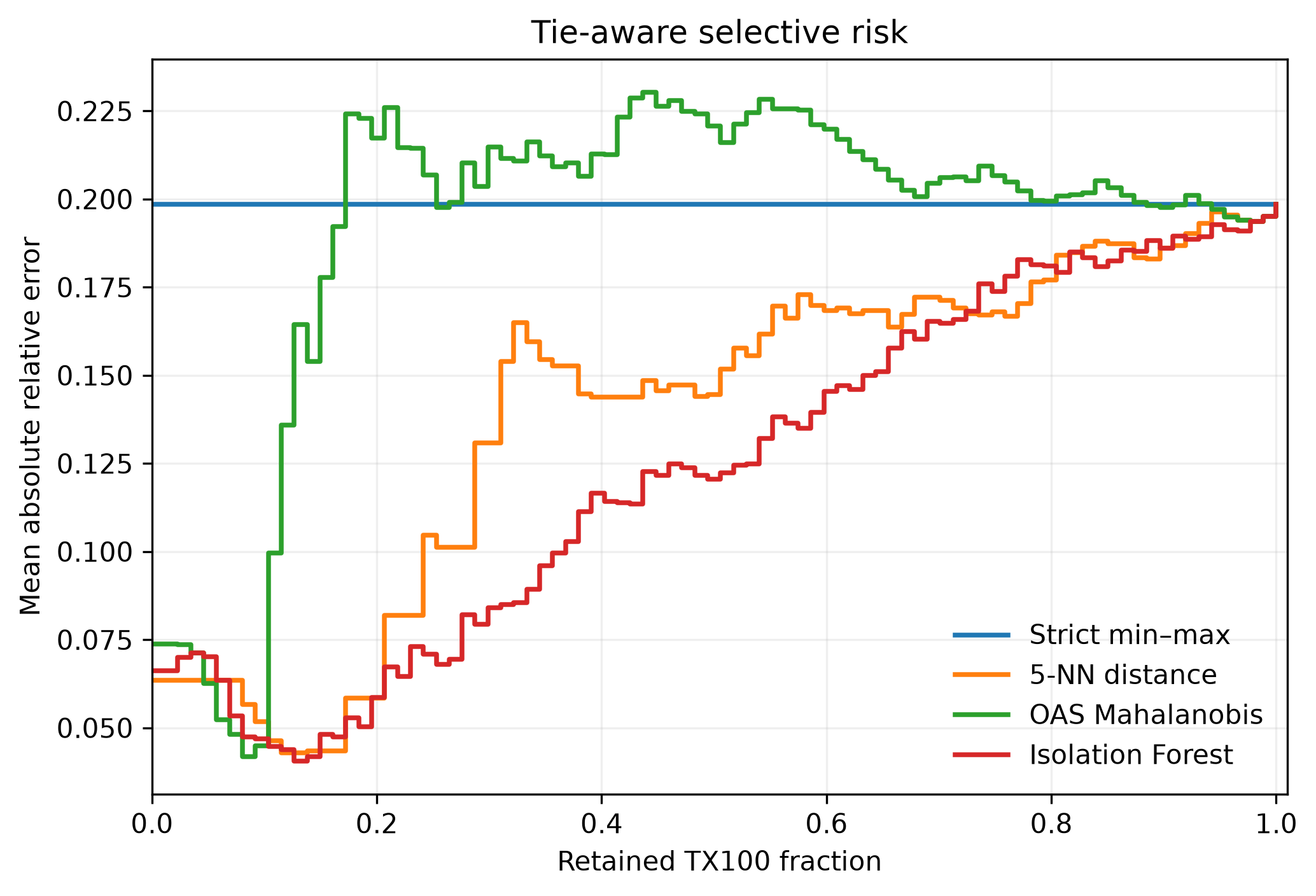}
\caption{Negative result, applicability view. Tie-aware selective risk on the
\tox{} rows for the four locked screens. No screen brings the retained risk below the full-pool risk except at a large loss of retention; the only pre-declared benefit
(Isolation Forest) does not survive BH adjustment.}
\label{fig:v1risk}
\end{figure}

Among the locked applicability rules, only Isolation Forest met all four
pre-declared benefit conditions (candidate reduction 0.1034, retained relative
risk 0.1861 versus 0.1984 for the full pool, matched-budget one-sided
$p=0.0266$, surviving at 0.05); 5-NN ($p=0.111$) and Mahalanobis ($p=0.675$)
did not. After BH adjustment across the four locked tests, however, the
Isolation Forest result is no longer significant (adjusted $p=0.106$), so we
report it as suggestive only (Figure~\ref{fig:v1risk}). All six target-search
strategies answered the same 8 of 13 fixed targets (coverage 0.615), with
success@1 of 0.2308 and success@3/@5 of 0.4615/0.6154 over all queries, and
oracle-achievable recall at 1/3/5 of 0.25/0.50/0.667; screening did not change
which candidate was ranked first except for Mahalanobis on one query. An
independent recomputation agreed with the recovered result to within
$10^{-12}$ on every metric, and all frozen source hashes were unchanged before
and after the post-reveal analysis.

\paragraph{Retrospective development reference (post hoc).} The pre-declared
thresholds are anchored to the source paper's in-distribution figures
(Section~\ref{sec:v1protocol}). Under a development-side criterion of the kind later
used for D3 ($1.25\times$ a grouped reference, here leave-one-chemistry-out), the locked V1 model's leave-one-chemistry-out MAPE over the five
development groups is 6.54\% (CMC 0, $n=63$), 5.01\% (2~mM, 30), 15.76\%
(4.33~mM, 15), 5.46\% (11~mM, 178) and 6.46\% (20~mM, 93): macro 7.85\%,
pooled 6.26\%, with 2 rows of unlisted CMC kept on the training side only.
The $1.25\times$ reproduction threshold would have been 9.81\%; the sealed
19.8436\% is 2.53 times the macro reference and fails this criterion as
well. The result is therefore negative under both the pre-registered and the
retrospective criterion, and the \tox{} error exceeds every development fold,
the nearest being the 15-row 4.33~mM group at 15.76\%.

\begin{table}[t]
\centering\small
\caption{Locked V1 endpoints on the sealed \tox{} group ($n=87$).}
\label{tab:v1}
\begin{tabular}{lr}
\toprule
Endpoint & Value \\
\midrule
MAPE & 19.8436\% \\
Reproduction criterion ($<11.4\%$) & not met \\
Strong-replication criterion ($\le 6.0\%$) & not met \\
MAE / RMSE & 0.022615 / 0.028681~mm \\
$R^2$ & $-0.2436$ \\
Nominal 90\% coverage (mean width) & 0.4483 (0.03281~mm) \\
Over-predicted rows & 85.06\% \\
Signed relative bias & 0.1895 \\
Strict min--max candidate reduction & 0.0 \\
Isolation Forest q95: reduction / $p$ / BH $p$ & 0.1034 / 0.0266 / 0.106 \\
Target queries answered & 8 of 13 \\
Success@1 / @3 / @5 (all queries) & 0.2308 / 0.4615 / 0.6154 \\
\bottomrule
\end{tabular}
\end{table}

\subsubsection{R7: Expanded shift diagnostics and negative controls on DAFD}
\label{sec:r7}
\begin{figure}[t]
\centering
\includegraphics[width=0.85\linewidth]{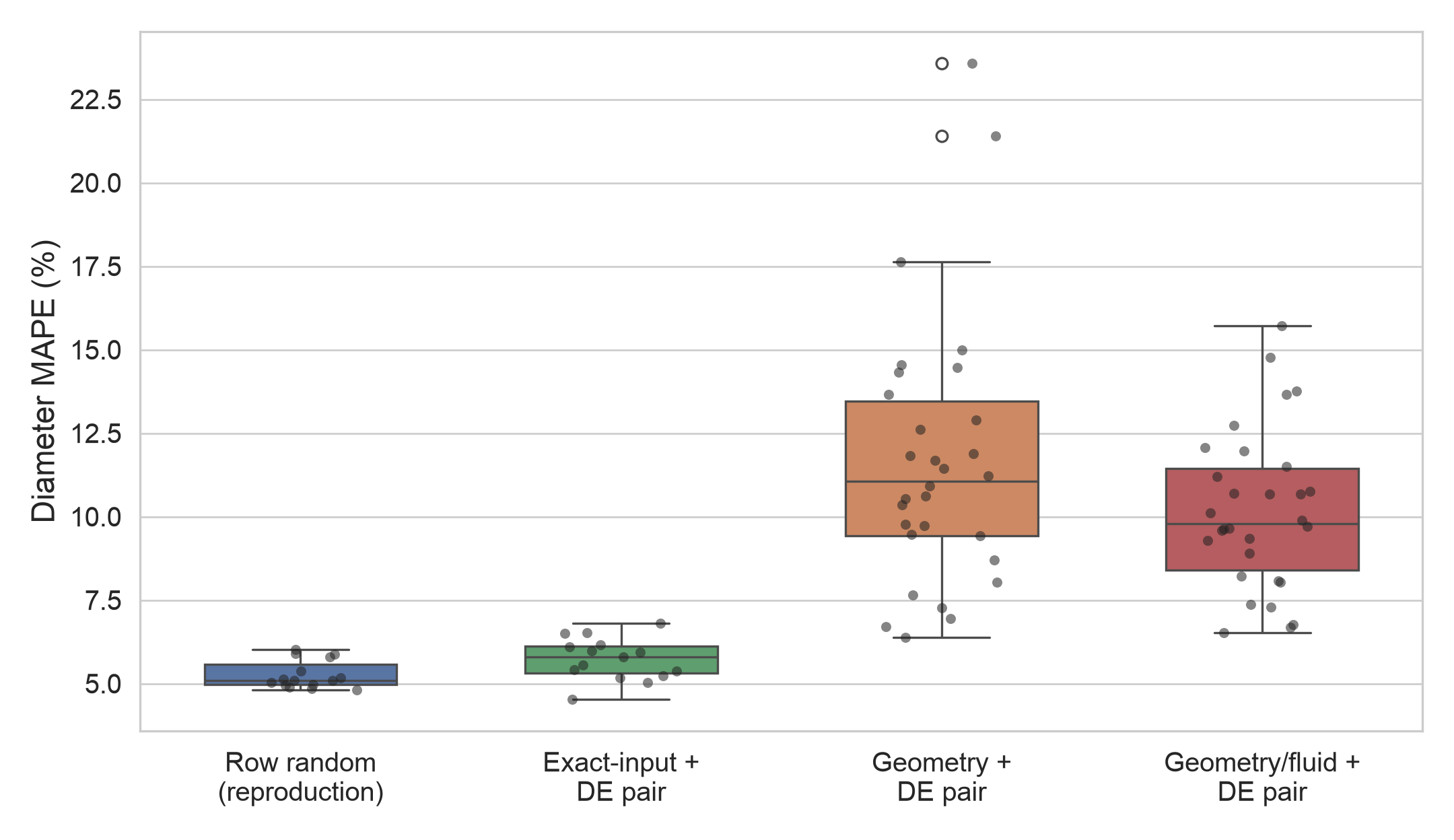}
\caption{Focal XGBoost diameter MAPE under four split protocols on the 868-row
table. Grouped protocols prevent double-emulsion pairs and device geometries
from crossing the fit/test boundary.}
\label{fig:splits}
\end{figure}

R5 located a sealed failure on D2 (R6, Section~\ref{sec:r6}, tests bounded
physics-guided variants on the same table), but neither could say how large a shift D1
tolerates, nor whether applicability screens act at the row or the source
level. R7 returns to D1 with grouped splits, negative controls and a replay
of target search to answer those two questions before the cross-table matrix
of R8.
\paragraph{Grouped splits.} For the focal XGBoost model, mean diameter MAPE is
5.27\% on row-random splits, 5.74\% on exact-input grouped splits, 11.69\% on
geometry-grouped splits (median 11.07\%, range 6.39--23.58\% over 30 splits)
and 10.18\% on geometry-plus-fluid grouped splits (Figure~\ref{fig:splits}).
The geometry-plus-fluid key joins rows only when geometry and viscosity ratio
both match, so it forms finer groups and lets a geometry recur across the
fit/test boundary with a different fluid; its lower 10.18\% describes that
weaker hold-out of geometry.
Row-random splits leak: on average 79\% of double-emulsion pairs cross the
train/test boundary and 99.5\% of test geometries appear in training. Random
forest and histogram gradient boosting show the same pattern (5.81/6.32/10.67/9.39\% and 5.01/5.41/11.33/10.12\%). A shuffled-label control gives external
MAPE 33.4\% (range 23.0--46.8\% over 30 seeds), confirming that the models
learn signal. Equal-weight source-macro external MAPE is 13.21\% for XGBoost
versus 11.26\% row-micro.

\paragraph{Interval transfer.} Under the expanded suite's calibration protocol, with row-level random calibration, nominal 90\% CQR coverage falls from 0.900 internally to 0.521 externally (mean width
28.3 to 34.1~\um); split and normalised split conformal behave alike
(0.908~$\to$~0.469; 0.912~$\to$~0.524). With geometry-grouped calibration,
external CQR coverage is 0.848 with mean width 89.8~\um{}, split conformal
0.890 (66.7~\um) and normalised split conformal 0.936 (104.2~\um); the
equal-weight source-macro external coverage for CQR is 0.750, and simultaneous
dependency-group coverage is 0.714. Grouped calibration therefore recovers
coverage at the cost of wider intervals (89.8~\um{} versus 34.1~\um{} under
row-random calibration); with few calibration groups the attainable
finite-sample coverage is capped at 0.95, above the 90\% target.

\begin{figure}[t]
\centering
\includegraphics[width=0.85\linewidth]{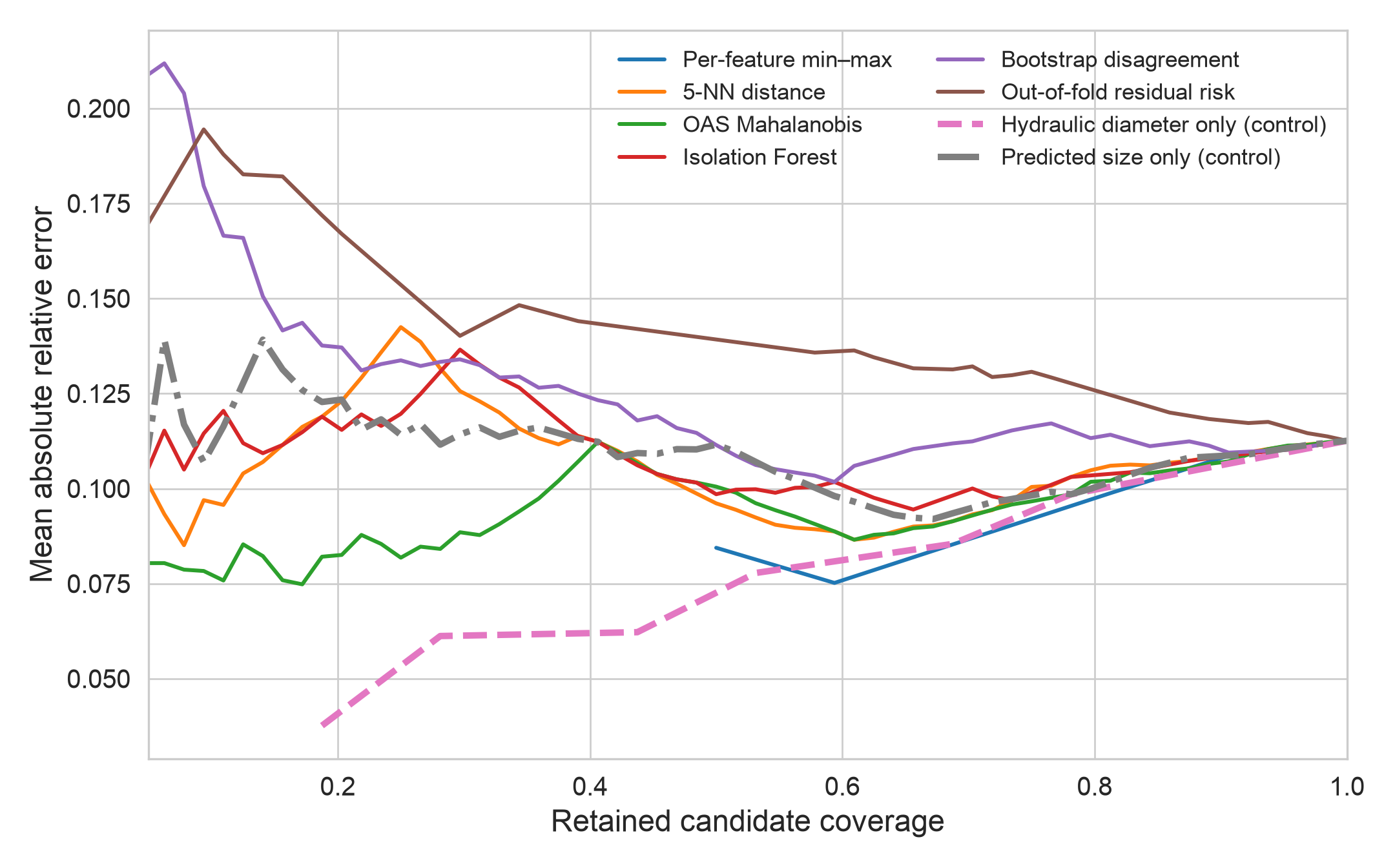}
\caption{Relative-risk versus retained-coverage curves for applicability
scores and controls on the 64 external rows. The hydraulic-diameter-only
control has the lowest AURC; no label-free applicability score beats it.}
\label{fig:riskcov}
\end{figure}

\begin{figure}[t]
\centering
\includegraphics[width=0.85\linewidth]{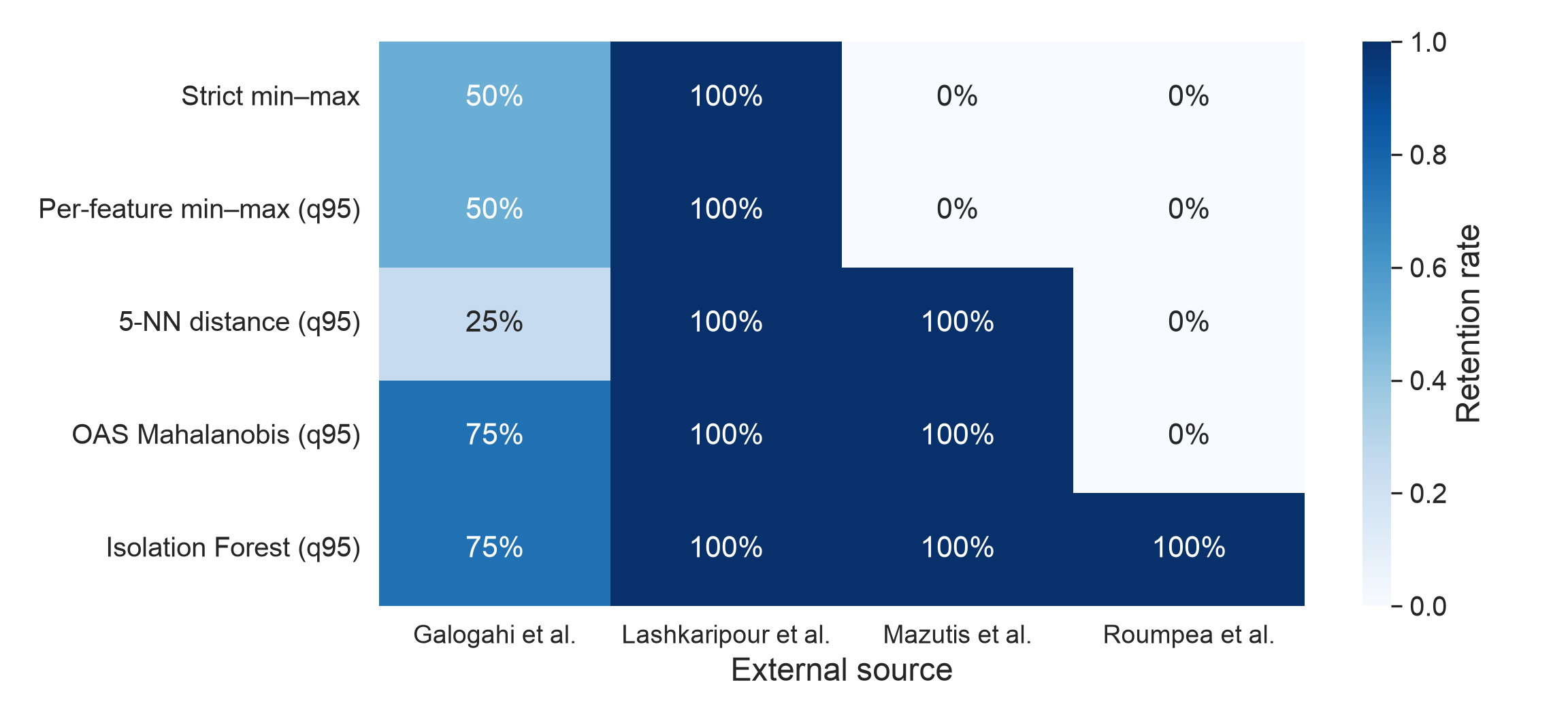}
\caption{Retention rate by external source for binary rules at the 95\%
calibration quantile. For every rule retention is all-or-none within
source--geometry contexts, which is why within-source-context permutation nulls explain the observed
risk reduction.}
\label{fig:retention}
\end{figure}

\paragraph{Applicability scores against controls.} On the continuous
relative-risk curve, the full-pool mean absolute relative error is 0.1126 and
AURC values are 0.1053 (5-NN), 0.0951 (Mahalanobis), 0.1082 (Isolation
Forest) and 0.109 (random score). The hydraulic-diameter-only control reaches
0.0734 and the one-class SVM 0.0882; the out-of-fold residual-risk model
(0.1431) and bootstrap disagreement (0.1273) are worse than random. Because
no applicability score beats the size-only control on continuous relative
risk, none of them supports a candidate-level benefit claim under our
pre-registered criteria (Figure~\ref{fig:riskcov}). The binary 5-NN
rule at q95 does lower retained relative risk (0.0961 versus 0.1126) relative
to matched-budget random retention (global $p=0.020$, BH 0.036;
source-stratified $p=0.0013$, BH 0.0039; within-source physical-cluster
permutation $p=0.0030$, BH 0.0076), but under a within-source-context
permutation that preserves which contexts are retained the effect vanishes
($p=1.0$), and it does so by construction: binary retention is all-or-none
within each of the eight source--geometry contexts, so permuting scores
within contexts leaves the retained row set, and hence the statistic,
unchanged (retrospective check, Section~\ref{sec:posthoc}); the benefit is
explained by which source contexts the rule keeps
(Figure~\ref{fig:retention}). The overall Spearman correlation between 5-NN
distance and relative error is 0.222 ($p=0.078$, row-naive) with opposite
signs within sources (Galogahi 0.79, Lashkaripour $-0.27$, Roumpea $-0.49$),
whereas hydraulic diameter alone correlates 0.69 with relative error.
Source-then-cluster hierarchical bootstrap intervals for retained-minus-full
relative risk include zero for 5-NN ($-0.016$, 95\% interval
$[-0.088, 0.027]$) and Mahalanobis ($-0.021$, $[-0.056, 0.021]$).

\paragraph{Retrospective target-search replay.} Over 84 context--target
queries in per-source pools (30 oracle-achievable at 10\% relative tolerance),
unscreened point ranking answers 26.2\% of queries with success@1 precision
0.50 and successful-target recall 0.367 (success@3: 0.636/0.467). The 5-NN q95
screen raises answered precision to 0.60 but answers only 6.0\% of queries
with recall 0.10; the Mahalanobis and strict min--max screens answer 14.3\%
with precision 0.667 and recall 0.267; conformal target inclusion answers
25.0\% with precision 0.524 and recall 0.367; a random-score screen at the
same quantile answers 26.2\% with precision 0.455 and recall 0.333. Screening
trades query coverage and recall for precision and never dominates the
unscreened or random strategies on all three endpoints at once.

\subsubsection{R8: Method-matrix benchmark across shift cells}
\label{sec:r8}
Two complete runs of the frozen matrix protocol (Section~\ref{sec:w3protocol})
in separate directories produced byte-identical artifact-hash manifests (13 entries each), and all 69 cell--repeat splits (across the seven cells) have zero dependency-unit overlap between fit, calibration and test rows. Table~\ref{tab:w3matrix} gives the
matrix for the reference XGBoost model, the two alternative point models and
the four interval methods. Because the point models see only the fit
partition, the reference cells sit above the R1 errors of
Section~\ref{sec:r1} (D1 6.71\% versus 5.27\%); the ratios in the table are
internal to this protocol.

\begin{table}[t]
\centering\small
\caption{Method-matrix benchmark. XGBoost MAPE and its ratio to the dataset's
random-split reference, the alternative point models, and empirical coverage
of nominal 90\% intervals (XGBoost; means over repeats). Reference cells use
the fit partition and are not comparable with Table~\ref{tab:repro}.}
\label{tab:w3matrix}
\resizebox{\linewidth}{!}{%
\begin{tabular}{llrrrrrrrrr}
\toprule
 & & & \multicolumn{4}{c}{Point error (MAPE; ratio to reference)} & \multicolumn{4}{c}{Nominal 90\% coverage} \\
\cmidrule(lr){4-7}\cmidrule(lr){8-11}
Cell & Shift & Repeats & XGB & Ratio & RF & MLP & Split & CQR & Mondrian & Ens. \\
\midrule
D1 random & reference & 15 & 6.71\% & 1.00 & 6.86\% & 14.52\% & 0.959 & 0.948 & 0.998 & 0.917 \\
D1 geometry & device geometry & 30 & 12.43\% & 1.85 & 11.78\% & 55.67\% & 0.860 & 0.866 & 0.970 & 0.854 \\
D1 external & cross-literature & 1 & 11.66\% & 1.74 & 11.80\% & 16.70\% & 0.453 & 0.484 & 0.734 & 0.547 \\
D1 SE$\to$DE & device type & 1 & 22.12\% & 3.30 & 21.19\% & 266.34\% & 0.330 & 0.414 & 0.525 & 0.586 \\
D1 DE$\to$SE & device type & 1 & 69.18\% & 10.31 & 73.67\% & 3592.49\% & 0.065 & 0.051 & 0.097 & 0.259 \\
D2 random & reference & 15 & 3.91\% & 1.00 & 5.32\% & 5.35\% & 0.904 & 0.911 & 0.989 & 0.891 \\
D2 chemistry LOCO & whole chemistry & 6 & 10.24\% & 2.62 & 9.07\% & 21.16\% & 0.528 & 0.786 & 0.857 & 0.670 \\
\bottomrule
\end{tabular}}
\end{table}

\begin{table}[t]
\centering\small
\caption{Method-matrix benchmark. AURC of the relative-error risk--coverage
curve (lower is better) for each applicability score and control, XGBoost,
means over repeats; bold marks the best entry in each cell.}
\label{tab:w3aurc}
\resizebox{\linewidth}{!}{%
\begin{tabular}{lrrrrrrrr}
\toprule
Cell & Full pool & Random & Size-only & 5-NN & Mahalanobis & Isol.\ Forest & Domain clf. & Density ratio \\
\midrule
D1 random & 0.0671 & 0.0662 & 0.0541 & \textbf{0.0466} & 0.0516 & 0.0535 & 0.0653 & 0.0543 \\
D1 geometry & 0.1243 & 0.1255 & 0.1141 & \textbf{0.1085} & 0.1142 & 0.1225 & 0.1153 & 0.1097 \\
D1 external & 0.1166 & 0.1107 & \textbf{0.0913} & 0.1135 & 0.0957 & 0.1178 & 0.1161 & 0.1039 \\
D1 SE$\to$DE & 0.2212 & 0.2256 & 0.3128 & 0.1442 & 0.1467 & 0.2219 & \textbf{0.1394} & 0.1737 \\
D1 DE$\to$SE & 0.6918 & 0.6734 & 0.6423 & 0.6017 & 0.5874 & \textbf{0.4182} & 0.5729 & 0.5263 \\
D2 random & 0.0391 & 0.0388 & 0.0398 & 0.0365 & 0.0379 & 0.0371 & 0.0368 & \textbf{0.0324} \\
D2 chemistry LOCO & 0.1024 & 0.1102 & 0.0884 & 0.0893 & 0.1089 & \textbf{0.0816} & 0.1013 & 0.1036 \\
\bottomrule
\end{tabular}}
\end{table}

\paragraph{Device-type shift is the largest shift in these data.} Training on the
474 single-emulsion rows and testing on the 394 double-emulsion rows raises
XGBoost MAPE to 22.12\% (ratio 3.30 to the D1 reference); the reverse
direction gives 69.18\% (ratio 10.31). Geometry grouping (12.43\%, ratio
1.85), cross-literature transfer (11.66\%, 1.74) and whole-chemistry
leave-one-out on D2 (10.24\%, 2.62) are all smaller. Random forest tracks
XGBoost in the shift cells (21.19\% and 73.67\% under device-type shift) but
not on the D2 random split (5.32\% versus 3.91\%). The
fixed-architecture multilayer perceptron, whose early stopping triggered
within 75--696 iterations in every fit, extrapolates to 266.34\% and
3592.49\% under device-type shift and to 55.67\% under geometry grouping: an
unbounded regressor has nothing to hold it inside the physical range once it
leaves the training support. The gap of one to two orders of magnitude between
the tree models and the perceptron has a structural reading: a random forest's
prediction in model-response space is an average of training targets and so
lies within the training response range, gradient boosting sums leaf values
fitted to that range and is effectively bounded by it, whereas a ReLU
perceptron with a linear output extrapolates affinely once it leaves the
training support; on typical tabular data tree ensembles are also
reported to outperform neural networks~\cite{grinsztajn2022}. The canonical per-row predictions bear this out
(retrospective check, Section~\ref{sec:posthoc}): in DE$\to$SE the XGBoost
and random-forest predictions of normalised diameter span 0.94--1.29 and
0.98--1.29, inside the double-emulsion training response range of 0.84--1.60
for 1.000 of the test rows, while the perceptron's span 21.1--22.4 lies
outside it for every row (fraction inside 0.000). The matrix also reproduces the quantities of
Sections~\ref{sec:r2}, \ref{sec:r5} and \ref{sec:r6} up to the fit-partition
difference: the \tox{} fold of the leave-one-chemistry-out cell gives 20.75\%
($n=87$, signed relative bias 0.203) compared with the sealed 19.84\%, the other
five folds range from 5.22\% to 16.15\%, the six-fold macro is 10.24\% versus B1's 10.27\%, and the external cell is 11.66\% (Galogahi 7.71\%,
Lashkaripour 9.03\%, Mazutis 17.87\%, Roumpea 19.52\%; source-macro
13.53\%) compared with 11.26\% for the 868-row model.

\paragraph{Interval transfer.} On the reference cells the three row-level
methods sit near nominal (D1: split conformal 0.959, CQR 0.948, ensemble
variance 0.917; D2: 0.904, 0.911, 0.891) and group-maximum Mondrian
conformal over-covers (0.998, 0.989). Under geometry grouping the row-level
methods fall to 0.854--0.866, while Mondrian reaches 0.970 row-level coverage at a mean width of 78.4~\um{} versus 43.9~\um{} for split conformal (though with few calibration groups its finite-sample group-level guarantee has an attainable ceiling of 0.857 against the 90\% target). Under cross-literature
shift coverage is 0.453 (split), 0.484 (CQR), 0.547 (ensemble) and 0.734
(Mondrian, 36.8~\um{} versus 26.3~\um). Under whole-chemistry leave-one-out,
CQR holds 0.786 where split conformal holds 0.528 and ensemble variance
0.670; Mondrian reaches 0.857 only by widening to 87.2~\um{} versus 21.6~\um. Under device-type shift every interval calibrated inside the
training device type is near failure (0.330--0.586 for SE$\to$DE;
0.051--0.259 for DE$\to$SE).

\paragraph{Applicability scores versus controls.} Table~\ref{tab:w3aurc}
gives the AURC of each score and control in each cell. The cross-literature
cell agrees qualitatively with Section~\ref{sec:r7}: the size-only control (0.0913) beats
every applicability score, the nearest being Mahalanobis (0.0957). Under
device-type shift the picture inverts: the size-only control is worse than
random (0.3128 versus 0.2256 for SE$\to$DE) while the domain classifier
(0.1394), 5-NN (0.1442) and Mahalanobis (0.1467) lie well below the
full-pool risk of 0.2212, so droplet size is not a proxy for a change of
device type. Binary rules thresholded at the 0.95 calibration quantile retain
almost no double-emulsion rows (5-NN retention 0.000 in both directions;
Isolation Forest 0.023 in SE$\to$DE, with precision 1.000 and recall 0.043
at 10~\um). This is the formal counterpart of a warning already in our own
deviation register: the canonical pruning thresholds of Section~\ref{sec:r4}
are calibrated on single-emulsion rows and support no claim about
double-emulsion candidates. Under whole-chemistry leave-one-out the global 5-NN
rule lowers precision (means over the six held-out chemistries: 0.416 at 10~\um{}
relative to a full-pool rate of 0.514;
retention 0.785, recall 0.824) while Isolation Forest raises it (0.541;
retention 0.894, recall 0.945): being near the training chemistries is not
the same as being reliable on a new one. In the cross-literature cell the
5-NN rule retains 0.688 of rows with precision 0.886 and recall 0.975
compared with a full-pool rate of 0.625 for the fit-partition model.

\paragraph{Candidate-level rule.} The pre-registered within-group criterion
is met by no score in the cross-literature cell (0 of 5 scores), in agreement
with Section~\ref{sec:r7}. In the geometry-grouped cell the transductive
domain classifier meets it in 12 of 30 seeds (within-group AURC 0.1126 versus 0.1238 for random and 0.1244 for size-only), 5-NN in 1, Mahalanobis
in 2, density ratio in 3 and Isolation Forest in 0; the domain classifier
sees the test inputs and so describes batch screening, not
candidate-by-candidate arrival. In the two reference cells the criterion
is triggered in 1--2 of 15 seeds per score, the order of the 5\% false-positive
rate. Although the device-type cells nominally pass the within-group test, these
passes are vacuous in practice: SE$\to$DE contains only 2 test groups, and in
DE$\to$SE (35 groups) the reference model has already failed (full-pool risk
0.6918), so within-group ordering ranks a failed model. The leave-one-chemistry-out cell has a single test group per fold
and is not evaluable under the criterion.

\paragraph{Metric robustness and the size confound (post hoc).} The two
device-type cells are single fixed splits, so their uncertainty is over rows,
not seeds: row-bootstrap 95\% intervals for XGBoost MAPE are 20.66--23.56\%
(SE$\to$DE), 65.38--73.28\% (DE$\to$SE) and 9.75--13.62\% (cross-literature).
Table~\ref{tab:w3robust} recomputes every cell under median APE, symmetric
MAPE and absolute log-ratio error. The ranking of the five shift cells by
ratio to the reference cell is identical under MAPE, symmetric MAPE and
log-ratio error (DE$\to$SE, SE$\to$DE, whole chemistry, geometry,
cross-literature); under median APE the cross-literature cell (2.89) moves
above chemistry (2.66) and geometry (2.21), and the two device-type cells
remain first (15.02 and 4.92). Device type is confounded with droplet size:
single-emulsion rows have observed diameters of 28.11--245.1~\um{} (median
85.71) and double-emulsion rows 15.57--54.21~\um{} (median 35.12); 0.698 of
double-emulsion rows lie inside the single-emulsion size range but only 0.141
of single-emulsion rows inside the double-emulsion range. Size-quartile bins
show that the error is not confined to out-of-range sizes. For SE$\to$DE the
MAPE rises with droplet size: 6.43\% (15.6--27.2~\um, $n=99$), 12.78\%
(27.2--35.1~\um), 33.02\% (35.1--40.6~\um) and 36.25\% (40.6--54.2~\um). The largest errors therefore fall on double-emulsion rows whose sizes the
single-emulsion training set covers. For DE$\to$SE the smallest
single-emulsion quartile (28.1--63.3~\um, $n=119$), which overlaps the
double-emulsion range, has the highest MAPE at 104.20\%, falling to 77.90\%
(63.3--85.7~\um, $n=118$), 61.46\% (85.7--118.3~\um, $n=118$) and 33.15\%
(118.3--245.1~\um, $n=119$) for the larger quartiles. Device-type shift therefore
degrades predictions inside the overlapping size range as well as outside
it. The direction of these errors is one-sided in DE$\to$SE and depends on size
in SE$\to$DE (retrospective check,
Section~\ref{sec:posthoc}). In DE$\to$SE every quartile is over-predicted,
with 1.000, 0.992, 0.924 and 0.798 of rows above the observed size and median
predicted-to-observed ratios of 2.03, 1.72, 1.65 and 1.31: the
double-emulsion training responses (normalised diameter 0.84--1.60) lie
above the observed responses of 0.823 of the single-emulsion rows, and a tree
model cannot predict below its training floor, so the falling MAPE across
quartiles is this upward offset divided by a growing observed size, not an
under-prediction ceiling. In SE$\to$DE the two smaller quartiles are nearly
unbiased (0.646 and 0.551 of rows over-predicted) while the two larger ones
are under-predicted (0.000 and 0.091 over-predicted; median ratios 0.65 and
0.63).

\begin{table}[t]
\centering\small
\caption{Post-hoc metric robustness of the method matrix (XGBoost; means over
repeats; not part of the frozen protocol). Median APE (MdAPE) and symmetric
MAPE (sMAPE, $100\cdot\mathrm{mean}\,|\hat d-d|/((|\hat d|+|d|)/2)$) in percent; log-ratio error is the mean absolute $\log(\hat d/d)$. Ratios are to
the same dataset's random-split reference cell under the same metric.}
\label{tab:w3robust}
\resizebox{\linewidth}{!}{%
\begin{tabular}{lrrrrrr}
\toprule
Cell & MdAPE & sMAPE & Log-ratio & Ratio MdAPE & Ratio sMAPE & Ratio log \\
\midrule
D1 random & 4.39\% & 6.80\% & 0.068 & 1.00 & 1.00 & 1.00 \\
D1 geometry & 9.71\% & 12.45\% & 0.125 & 2.21 & 1.83 & 1.84 \\
D1 external & 12.69\% & 11.66\% & 0.117 & 2.89 & 1.71 & 1.71 \\
D1 SE$\to$DE & 21.60\% & 25.89\% & 0.263 & 4.92 & 3.81 & 3.85 \\
D1 DE$\to$SE & 65.92\% & 47.47\% & 0.492 & 15.02 & 6.98 & 7.20 \\
D2 random & 2.88\% & 3.83\% & 0.038 & 1.00 & 1.00 & 1.00 \\
D2 chemistry LOCO & 7.69\% & 10.21\% & 0.103 & 2.66 & 2.67 & 2.69 \\
\bottomrule
\end{tabular}}
\end{table}

\paragraph{Uncertainty-aware families: point error and shift ordering.}
Table~\ref{tab:w4} reports the quantile forest, the Gaussian process (GP) and the
deep ensemble of Section~\ref{sec:w4protocol} on the same cells, seeds and
partitions, from two runs with byte-identical 11-entry manifests. The shift
ordering is a property of the data, not of XGBoost: for every family the
DE$\to$SE cell has the largest error ratio to its own reference (9.37, 8.93
and 226.33) and SE$\to$DE the second largest for the forest and the GP
(2.77 and 3.96). The deep ensemble, like the matrix's single perceptron,
diverges severely under device-type shift (302.83\% and 3271.87\%).

\paragraph{Ranking under random splits versus under chemistry shift
(descriptive).} Because the matrix and the uncertainty-aware protocol share
cells, seeds and partitions, the six point models can be ordered on D2 under
both protocols. By random-split MAPE the order is GP 3.65\%, XGBoost
3.91\%, deep ensemble 5.12\%, random forest 5.32\%, perceptron 5.35\% and
quantile forest 6.39\%; under whole-chemistry leave-one-out it is random
forest 9.07\%, quantile forest 9.85\%, XGBoost 10.24\%, GP 16.60\%, deep
ensemble 20.89\% and perceptron 21.16\%, with inflation ratios of 1.71,
1.54, 2.62, 4.55, 4.08 and 3.95 in the same order (exact ratios of the
artifact values). In D2 the ranking under
random splits does not predict the ranking under chemistry shift: the two
best random-split models are fourth (GP) and third (XGBoost) under shift, and the two forest
variants, fourth and sixth at random, are first and second. We report the
ordering as observed and leave its explanation to a pre-registered test.

\paragraph{Uncertainty-aware families: interval coverage and the
Gaussian-process exception.} No interval calibrated inside the training
device type holds coverage in the DE$\to$SE direction: native coverage is
0.072 (forest), 0.652 (GP) and 0.000 (ensemble), and conformalisation
does not help (0.072, 0.629, 0.000). The one departure from the
pre-registered expectation that no native interval would survive either
device-type cell is the Gaussian process in SE$\to$DE, whose native interval
covers 0.959 of double-emulsion rows (0.982 after conformalisation) because
its predictive standard deviation grows outside the training support, at a
mean width of 49.2~\um{} versus 9.7~\um{} for the forest, on rows whose
median diameter is 35.12~\um; the same mechanism fails in the opposite
direction (0.652 at 162.0~\um). Under whole-chemistry shift the forest's
native interval reaches 0.830 (width 51.4~\um), above the matrix's CQR at
0.786 though below Mondrian's 0.857 at 87.2~\um{} in the matrix; under
cross-literature shift the conformalised forest reaches 0.812 and the
conformalised ensemble 0.906 at 84.2~\um{} where the XGBoost split-conformal
interval covered 0.453. The ensemble's and the forest's native intervals are
wider on the external rows than on the internal reference rows (mean native
width 41.2~\um{} versus 16.0~\um{} for the ensemble and 39.2 versus
27.5~\um{} for the forest; retrospective check, Section~\ref{sec:posthoc}),
consistent with member disagreement growing on inputs away from the training
set; whether that disagreement tracks error is a separate question that
these cells do not settle. Conformalisation restores near-nominal coverage on
both reference cells for all three families (0.903--0.953 on D1,
0.894--0.896 on D2; the GP over-covers D1 at 0.953, marginally outside
the pre-registered 0.05 band), whereas the deep ensemble's native interval is
badly under-dispersed even in distribution (0.403 and 0.535), the known
behaviour of mean--variance ensembles without a likelihood head.

\paragraph{Uncertainty-aware families: native uncertainty as a score.}
Model-native uncertainty is a good in-distribution ranking score (it beats
both controls in 0.93 of the D1 random-split repeats for the forest and for
the GP), but, as the pre-registered expectation anticipated, every
family is beaten by the size-only control or by the random score in at least
one shift cell: the forest's quantile gap ranks double-emulsion candidates
worse than random in SE$\to$DE (0.3072 versus 0.2132) and is beaten by
size-only in the cross-literature cell (0.1115 versus 0.0994); the GP
is beaten by size-only in both device-type cells (0.3760 versus 0.3145;
0.8487 versus 0.7968); the ensemble's score fails in DE$\to$SE. The interval
findings of the matrix are therefore not XGBoost-specific: the collapse is a
property of the shift, with the one qualification that a model whose
uncertainty grows with distance from the training inputs can secure coverage
under one direction of device-type shift by widening to the scale of the
quantity it predicts.

\begin{table}[t]
\centering\small
\caption{Uncertainty-aware model families on the method-matrix cells (frozen
W4 protocol; means over repeats; two runs byte-identical; QRF: quantile
regression forest). Ratio is MAPE
relative to the same model's random-split reference cell. Coverage of
nominal 90\% intervals: native, split conformal around the point prediction,
and the conformalised native interval. AURC of each model's relative-error risk--coverage curve (lower is better) for its relative uncertainty score and for the size-only control.}
\label{tab:w4}
\resizebox{\linewidth}{!}{%
\begin{tabular}{llrrrrrrr}
\toprule
Cell & Model & MAPE & Ratio & Native & Split & Conformalised & AURC uncert. & AURC size-only \\
\midrule
D1 random & QRF & 7.65\% & 1.00 & 0.908 & 0.947 & 0.903 & 0.0540 & 0.0622 \\
 & GP & 9.51\% & 1.00 & 0.860 & 0.974 & 0.953 & 0.0596 & 0.0679 \\
 & Deep ens. & 14.46\% & 1.00 & 0.403 & 0.960 & 0.914 & 0.0784 & 0.1065 \\
D1 geometry & QRF & 12.00\% & 1.57 & 0.859 & 0.863 & 0.873 & 0.1047 & 0.1115 \\
 & GP & 20.81\% & 2.19 & 0.709 & 0.862 & 0.885 & 0.1847 & 0.1907 \\
 & Deep ens. & 39.81\% & 2.75 & 0.608 & 0.847 & 0.885 & 0.3328 & 0.3838 \\
D1 external & QRF & 11.95\% & 1.56 & 0.781 & 0.531 & 0.812 & 0.1115 & 0.0994 \\
 & GP & 20.51\% & 2.16 & 0.734 & 0.625 & 0.734 & 0.1500 & 0.2180 \\
 & Deep ens. & 17.80\% & 1.23 & 0.328 & 0.906 & 0.906 & 0.1202 & 0.1226 \\
D1 SE$\to$DE & QRF & 21.18\% & 2.77 & 0.525 & 0.411 & 0.525 & 0.3072 & 0.3064 \\
 & GP & 37.60\% & 3.96 & 0.959 & 0.122 & 0.982 & 0.3760 & 0.3145 \\
 & Deep ens. & 302.83\% & 20.95 & 0.330 & 0.030 & 0.391 & 1.4859 & 4.5986 \\
D1 DE$\to$SE & QRF & 71.65\% & 9.37 & 0.072 & 0.044 & 0.072 & 0.5954 & 0.6664 \\
 & GP & 84.87\% & 8.93 & 0.652 & 0.017 & 0.629 & 0.8487 & 0.7968 \\
 & Deep ens. & 3271.87\% & 226.33 & 0.000 & 0.000 & 0.000 & 35.0508 & 32.0161 \\
D2 random & QRF & 6.39\% & 1.00 & 0.928 & 0.907 & 0.894 & 0.0532 & 0.0636 \\
 & GP & 3.65\% & 1.00 & 0.884 & 0.899 & 0.896 & 0.0332 & 0.0364 \\
 & Deep ens. & 5.12\% & 1.00 & 0.535 & 0.894 & 0.896 & 0.0417 & 0.0522 \\
D2 chemistry LOCO & QRF & 9.85\% & 1.54 & 0.830 & 0.702 & 0.758 & 0.0825 & 0.0967 \\
 & GP & 16.60\% & 4.55 & 0.699 & 0.359 & 0.716 & 0.1382 & 0.1333 \\
 & Deep ens. & 20.89\% & 4.08 & 0.274 & 0.359 & 0.447 & 0.1751 & 0.1877 \\
\bottomrule
\end{tabular}}
\end{table}

\paragraph{Covariate-shift weighting does not rescue the intervals.}
Table~\ref{tab:w5} reports the reweighted split-conformal intervals of
Section~\ref{sec:w5protocol} for the reference XGBoost model (two runs,
byte-identical 7-entry manifests; the unweighted branch reproduces the matrix's split-conformal coverage across all 69 cell--repeat splits to machine precision). On
the reference cells the domain-classifier weights stay close to one and move
coverage by at most 0.005 (0.959 to 0.955 on D1, 0.904 to 0.899 on D2), as
expected. The $k$-NN density ratio is too crude in eight dimensions (its
effective sample size is 0.137 of the calibration set even in
distribution) and lowers D1 coverage to 0.863, outside the pre-registered
0.03 band. Under device-type and cross-literature shift both weightings drive
every test weight orders of magnitude above the calibration weights, so every
interval becomes infinite: the reported coverage of 1.000 is abstention, not
prediction, and no finite-interval coverage exists. This is the method's
correct behaviour when the test input lies outside the calibration support,
and it agrees with the applicability screens that retain no double-emulsion
rows; it also means that the covariate-shift repair has nothing to offer in
exactly the cells where coverage was lost. Under geometry shift, the shift with the
smallest coverage loss here, domain weighting restores nominal coverage (0.908 versus 0.860)
at the price of 0.209 of intervals infinite and finite intervals of
52.1~\um{} versus 43.9~\um. Under whole-chemistry shift it reaches 0.848 versus 0.528, but 0.326 of intervals are infinite and the finite ones cover
only 0.662 at 77.5~\um{} versus 21.6~\um; by the pre-declared coverage
endpoint domain weighting survives 4 of 6 folds, which is more than the
pre-registered expectation allowed, and it does so largely by abstaining (on
the \tox{} fold about three quarters of its intervals are infinite). The
chemistry failure is therefore not repaired by either input-only reweighting tested here,
consistent with the conditional-shift reading of the \tox{} result, and the
device-type failure is a support mismatch that reweighting can name but not
fix.

\begin{table}[t]
\centering\small
\caption{Covariate-shift-weighted split conformal for the reference XGBoost
model (frozen W5 protocol; means over repeats; two runs byte-identical).
``Cov.'' counts infinite intervals as covering; ``inf.'' is the fraction of
infinite intervals; ``fin.\ cov.'' and ``width'' are over finite intervals
only; ESS is the effective sample size of the calibration weights divided by
the calibration size: it describes how evenly the calibration weights are
spread among themselves, not the test weight relative to them, so an
interval can be infinite (test-point weight share above $1-0.90$) while ESS
is near one, as in D1 SE$\to$DE. Dashes: every interval infinite.}
\label{tab:w5}
\resizebox{\linewidth}{!}{%
\begin{tabular}{lrr|rrrrr|rrrrr}
\toprule
 & \multicolumn{2}{c|}{Unweighted} & \multicolumn{5}{c|}{Domain-classifier odds} & \multicolumn{5}{c}{$k$-NN density ratio} \\
Cell & Cov. & Width & Cov. & Inf. & Fin.\ cov. & Width & ESS & Cov. & Inf. & Fin.\ cov. & Width & ESS \\
\midrule
D1 random & 0.959 & 33.0 & 0.955 & 0.000 & 0.955 & 32.4 & 0.841 & 0.863 & 0.002 & 0.862 & 24.1 & 0.137 \\
D1 geometry & 0.860 & 43.9 & 0.908 & 0.209 & 0.880 & 52.1 & 0.420 & 0.886 & 0.553 & 0.788 & 42.0 & 0.168 \\
D1 external & 0.453 & 26.3 & 1.000 & 1.000 & --- & --- & 0.076 & 1.000 & 1.000 & --- & --- & 0.281 \\
D1 SE$\to$DE & 0.330 & 8.9 & 1.000 & 1.000 & --- & --- & 0.998 & 1.000 & 1.000 & --- & --- & 0.416 \\
D1 DE$\to$SE & 0.065 & 21.2 & 1.000 & 1.000 & --- & --- & 0.836 & 1.000 & 1.000 & --- & --- & 0.018 \\
D2 random & 0.904 & 23.8 & 0.899 & 0.000 & 0.899 & 23.7 & 0.963 & 0.917 & 0.004 & 0.917 & 26.7 & 0.431 \\
D2 chemistry LOCO & 0.528 & 21.6 & 0.848 & 0.326 & 0.662 & 77.5 & 0.243 & 0.852 & 0.568 & 0.546 & 37.6 & 0.151 \\
\bottomrule
\end{tabular}}
\end{table}

\subsubsection{R9: Second sealed validation under laboratory shift --- a negative result}
\label{sec:r9}
The reveal gate for the Smart Microfluidics sealed set was consumed once
(2026-10-17T03:41:48Z), after the reveal program had re-verified the digests
of the protocol, binding, deviation register, custodian manifest, prediction
bundle and three frozen program files. Table~\ref{tab:d3} lists the locked
endpoints. The frozen XGBoost size surrogate, fitted on 246 laboratory-A
rows, reached a MAPE of 240.39\% on the 57 sealed laboratory-B rows compared with a frozen development reference of 39.41\% (reproduction threshold 49.27\%);
both pre-declared criteria fail. No sealed row (0/57, fraction 0.000) is
within 10\% relative error; the median absolute percentage error (MdAPE,
descriptive) is 206.74\%, MAE 595.3~nm, RMSE 1335.0~nm, $R^2=0.026$, and the
signed relative bias is $-0.700$ with over-prediction on 0.456 of rows. The nominal 90\% split-conformal interval
(calibration radius 344.19~nm from 50 calibration rows; 196 fit rows) covered
0.719 of rows at a mean width of 688.4~nm. Observed sizes have median
98.8~nm (range 64.4--7932.7~nm); predictions have median 200.0~nm but range
from $-336.3$ to 4036.1~nm: 22 of 57 point predictions (38.6\%) are negative
diameters and 46 of 57 interval lower bounds (80.7\%) are at or below zero.
Nothing in the frozen configuration bounds the output, and these rows are
physically impossible.

\paragraph{Why it failed.} Two facts were on record from the inputs before
any fit (Section~\ref{sec:d3protocol}): every laboratory-B row uses PBS
buffer where the development set has 231 MQ and 15 PBS rows, and only 18 of
57 rows (31.6\%) fall inside the per-feature development range (in-range fraction
per feature: PEG 0.316, HSPC 0.684, all other features in range). After reveal, the 18 in-range rows have median
absolute relative error 0.884 and the 39 out-of-range rows 2.755. Range
violation is therefore strongly associated with error, but a median error of
88\% inside the marginal range is, as with \tox{}, a conditional shift as
well: the same PBS formulations give different sizes on another laboratory's
equipment. Unlike the \tox{} case, the marginal screens did flag the problem
before reveal; what they could not do is turn a flagged batch into reliable
candidates.

\paragraph{Applicability rules.} In the corrected applicability block
(see the disclosure at the end of this section) the full-pool mean absolute relative error is
2.404 and the number of reliable rows (within 10\%) is zero under any rule.
The 5-NN rule at the 0.95 quantile retains 7 of 57 rows (0.123) at a
retained risk of 1.079, with matched-budget random $p=0.0046$ and
BH-adjusted $p=0.018$ across the four non-control rules; it meets the four locked conditions
formally, but with zero reliable candidates and compared with a predicted-size-only
control that retains 23 of 57 rows (0.404) at risk 1.204 ($p=0.0001$), we
make no benefit claim. Mahalanobis retains 2 rows (risk 1.295, $p=0.169$, BH
0.225), the domain classifier 1 row (0.237, $p=0.053$, BH 0.106), Isolation
Forest none (0/57) and strict min--max (a control) 18 rows (1.614, $p=0.0027$). At the
candidate level the pre-registered buffer stratum is not evaluable (one level
in the sealed set); under the main-lipid stratum adopted before any fit (ESM 27, HSPC 30;
Section~\ref{sec:d3protocol})
all four scores beat the random control (within-stratum AURC 2.105): 5-NN
1.926 ($p=0.0025$, BH 0.0033), Mahalanobis 1.850 ($p=0.0005$, BH 0.0010),
Isolation Forest 2.023 ($p=0.016$), domain classifier 1.806 ($p=0.0005$).
None beats the predicted-size-only control (1.562), so the verdict is no
candidate-level benefit, as in Sections~\ref{sec:r7} and \ref{sec:r8}.

\paragraph{Alternative metrics and a log-target baseline (post hoc).} The
headline MAPE is not a tail artifact: on the same 57 rows the symmetric MAPE
is 129.60\%, the median APE 206.74\%, and the mean absolute log-ratio error
over the 35 positive predictions is 0.969; every quartile of observed size
carries a large error and some negative predictions. The quartiles are Q1
64.4--92.9~nm ($n=14$), Q2 92.9--98.8~nm ($n=14$), Q3 98.8--130.2~nm
($n=14$) and Q4 130.2--7932.7~nm ($n=15$); their MAPEs are 249.9\%, 306.5\%,
241.1\% and 169.1\%, their median APEs 208.7\%, 402.4\%, 201.6\% and 102.6\%,
and their negative-prediction fractions 0.286, 0.571, 0.357 and 0.333.
Refitting the locked configuration to
$\log(\mathrm{size})$ changes the picture. Its grouped five-fold development
reference is 16.89\% versus 39.41\% for the raw-scale model (whose folds and
predictions the refit reproduces exactly), so the raw-scale target was
already costly inside laboratory A. On laboratory B the log-target model
gives MAPE 43.05\%, median APE 34.57\%, symmetric MAPE 52.94\%, mean absolute
log-ratio 0.644, 0.070 of rows within 10\% and no negative predictions; its
predictions span 56.5--366.3~nm compared with observed sizes up to 7932.7~nm, and
its largest-size quartile remains the worst (65.8\% versus 38.7\%, 25.8\%
and 40.2\% for the smaller quartiles). This 43.05\% lies below the frozen
reproduction threshold of 49.27\% but is 2.55 times the model's own
development reference (threshold 21.11\%), so it fails the like-for-like
criterion. We read the frozen result as two separable components: a
target-scale choice inherited from the D2 source-paper configuration, which
produced the negative predictions and most of the magnitude, and a residual
laboratory shift of about two and a half times the development error, which
survives the correction. Because the log-target model was chosen after the
reveal, it is a diagnosis, not a validation result.

\paragraph{Secondary endpoints.} PDI is predicted with MAPE 41.24\% (MdAPE
31.44\%; 0.105 of rows within 10\%). The formation classifier reaches
accuracy 0.912 and F1 0.954, but 53 of 57 sealed rows are positive, so the
majority-class baseline is 0.930 and the classifier does not beat it. The
12-row secondary sealed set (operator B, laboratory A, same equipment),
outside the primary estimand and reported descriptively only, gives MAPE
220.27\%, MdAPE 71.87\%, 0.25 of rows within 10\%, coverage 0.833 and 3/12
negative predictions; the median error under operator shift is far below that
under laboratory shift, but 12 rows support no stronger statement.

\paragraph{Disclosure.} After the reveal, the applicability-benefit block of
the frozen reveal program was found to have received signed rather than
absolute relative error, which is why the frozen result file shows a
full-pool ``risk'' of $-0.700$. The primary endpoint, both criteria, the
interval block, the candidate-level block and every point metric use absolute
error and were unaffected. The block was recomputed by a separate program on
the already-revealed rows into a separate file; no frozen artifact was
modified. We describe the D3 reveal as \emph{a sealed prediction followed by
a disclosed post-reveal scoring correction of the applicability block}. The
gate is consumed: the laboratory-B and operator-B labels are no longer blind,
and any further analysis of them is retrospective.

\begin{table}[t]
\centering\small
\caption{Locked endpoints of the second sealed validation (Smart
Microfluidics, laboratory A $\to$ laboratory B, $n=57$).}
\label{tab:d3}
\begin{tabular}{p{0.74\linewidth}r}
\toprule
Endpoint & Value \\
\midrule
Development reference MAPE (laboratory A, grouped 5-fold CV, frozen pre-reveal) & 39.41\% \\
Reproduction ($\le 1.25\times$) / strong-replication ($\le 1.00\times$) thresholds & 49.27\% / 39.41\% \\
Sealed laboratory-B MAPE & 240.39\% \\
Reproduction / strong-replication criteria & not met / not met \\
MdAPE (descriptive) & 206.74\% \\
MAE / RMSE & 595.3 / 1335.0~nm \\
$R^2$ & 0.026 \\
Rows within 10\% relative error & 0/57 (0.000) \\
Signed relative bias / over-predicted fraction & $-0.700$ / 0.456 \\
Nominal 90\% coverage (mean width) & 0.719 (688.4~nm) \\
Negative point predictions & 22/57 (38.6\%) \\
Interval lower bound $\le 0$ & 46/57 (80.7\%) \\
Strict min--max acceptance (recorded pre-reveal) & 18/57 (31.6\%) \\
Candidate-level benefit, main-lipid stratum (4 of 4 scores beat random; 0 of 4 beat size-only) & none \\
\bottomrule
\end{tabular}
\end{table}

\subsubsection{R10: Pre-registered validation of five hypotheses on three further public tables}
\label{sec:r10}

R1--R9 revealed descriptive regularities beyond their pre-declared
endpoints: an ordering of shift effects, positivity bounds that remove invalid
outputs without buying accuracy, no clear advantage of dimensionless inputs,
intervals too wide for tolerance-based retrieval, and model rankings that
changed under shift. Asserting any of them as a general conclusion after
D1--D3 had been inspected would violate the discipline of this evaluation. They
were therefore restated as the five falsifiable hypotheses H-A to H-E of
Section~\ref{sec:hprotocol} and tested under a Stage~2 protocol frozen before
any Stage~2 data file was downloaded (\texttt{H\_PROTOCOL\_v1}; v2, which changed
only the dataset list, was frozen before the R2-A1 and R2-A2 files were
downloaded). Of the candidates screened, three tables were admitted,
four exited at intake because no row-level table existed, and the rest were
excluded as sharing a source with D1--D3; the multi-table success conditions
need at least two admitted tables.

The admitted tables are A1, the non-Newtonian flow-focusing droplet study of
Hormozinezhad et al.~\cite{labchip2025dimless} (ESI, CC~BY~3.0; 901 rows after
the binding's rule removed rows without a measurable droplet; 13 fluid
families on one device form the development set, one Tween family is sealed
as S2 and the whole second device, with nine fluid families of its own, as
S3); R2-A1, the co-flow capillary table of Wang et al.~\cite{wang2024coflow}
(Zenodo, CC~BY~4.0; 830 rows and eleven fluid pairs on two capillary devices;
the count rule seals a 164-row fluid pair as S2 and the 109-row minority
device, whose eight fluid pairs are all absent from development, as S3, which
leaves two fluid pairs and 557 rows for development); and R2-A2, the LNP
Atlas of Song et al.~\cite{lnpatlas2025} (Scientific Data; Zenodo,
CC~BY~4.0), a peer-reviewed literature compilation of lipid-nanoparticle
formulations rather than one laboratory's measurements, of which 919 of the
1092 rows carry a parsable size; the 571 rows prepared by microfluidic
mixing form the main table (S2: one ionizable lipid, 32 rows; S3: one source
paper, 16 rows; 523 development rows from 33 papers) and the 348
non-microfluidic rows are sealed as a preparation-method tier S4.
Table~\ref{tab:r10tiers} gives the sealed results and
Table~\ref{tab:r10decisions} the pre-registered decisions. Every deviation
(a file encoding, a fold-count fallback when only two dependency groups
remained, a single-group calibration fallback, and an implementation defect
in the bounded-perceptron contrast found after the R2-A1 reveal and corrected
by a re-run that left every other model's sealed prediction byte-identical to
the pre-reveal bundle) is in the register.

\begin{table}[t]
\centering\footnotesize
\caption{R10 sealed tiers. dev-ref is the grouped cross-validation MAPE of
the locked XGBoost on the development rows (2 folds on R2-A1, 5 folds
otherwise); $r$ is the sealed MAPE over dev-ref with row-bootstrap 95\%
intervals; coverage is split-conformal coverage at nominal 0.90; the
perceptron pair is the log-target (bounded) and raw-scale (unbounded) MLP
with the number of non-positive predictions; $\rho$ is the Spearman
correlation between the six-model ranking on random splits and on the tier.
A1 device-tier MLP values are the pre-registered run; R2-A1 bounded-MLP
values are from the corrected re-run.}
\label{tab:r10tiers}
\resizebox{\textwidth}{!}{%
\begin{tabular}{llrrrrrr}
\toprule
Table & Tier ($n$) & dev-ref & XGBoost MAPE [95\% CI] & $r$ [95\% CI] & Cov. & log-/raw-MLP MAPE (invalid) & $\rho$ \\
\midrule
A1 & S2 fluid family (40) & 21.82 & 28.78 [25.0, 32.7] & 1.32 [1.15, 1.50] & 0.900 & 15.6 / 18.0 (0 / 0) & $-0.77$ \\
A1 & S3 device (200) & 21.82 & 109.84 [95.1, 125.6] & 5.03 [4.36, 5.76] & 0.225 & $1.66\times10^{6}$ / 1134 (0 / 0) & $-0.14$ \\
R2-A1 & S2 fluid pair (164) & 13.62 & 25.28 [22.4, 28.5] & 1.86 [1.65, 2.10] & 0.817 & 26.0 / 27.7 (0 / 0) & 0.37 \\
R2-A1 & S3 device (109) & 13.62 & 67.48 [62.3, 72.8] & 4.95 [4.57, 5.34] & 0.330 & 97.5 / 2286 (0 / 103) & $-0.71$ \\
R2-A2 & S2 ionizable lipid (32) & 40.87 & 32.63 [23.9, 42.0] & 0.80 [0.58, 1.03] & 0.938 & 31.8 / 36.7 (0 / 0) & $-0.54$ \\
R2-A2 & S3 source paper (16) & 40.87 & 35.26 [19.5, 58.3] & 0.86 [0.48, 1.43] & 0.938 & 29.6 / 36.0 (0 / 0) & 0.20 \\
R2-A2 & S4 non-microfluidic (348) & 40.87 & 25.37 [23.1, 27.8] & 0.62 [0.57, 0.68] & 0.974 & 36.5 / 34.2 (0 / 0) & --- \\
\bottomrule
\end{tabular}}
\end{table}

\begin{table}[t]
\centering\footnotesize
\caption{R10 pre-registered decisions (protocol section 5, computed by the
frozen reveal and combination programs). ``Supported'' and ``falsified'' are
the protocol's terms for these data; ``not determinable'' means that the
protocol's own evaluability condition failed.}
\label{tab:r10decisions}
\begin{tabular}{p{0.20\linewidth}p{0.16\linewidth}p{0.16\linewidth}p{0.16\linewidth}p{0.18\linewidth}}
\toprule
Hypothesis & A1 & R2-A1 & R2-A2 & Protocol level \\
\midrule
H-A band $[2.2,3.0]$ & falsified (1.32 / 5.03) & falsified (1.86 / 4.95) & falsified (0.80 / 0.86; S4 0.62) & falsified on all three \\
\mbox{H-B1} bounded vs unbounded MLP & falsified (S3) & supported & supported & falsified (A1 fails) \\
\mbox{H-B2} anchored residual & no anchor & no anchor & no anchor & not evaluable \\
\mbox{H-C1} / \mbox{H-C2} dimensionless vs native & not supported ($+0.28$ pp, $p=0.45$) / 0 pairs & not supported (2 groups) / 0 pairs & not evaluable & not determinable \\
H-D $|A-W|\le 0.10$ & $A=W=0$ & $A=W=0$ & $A=W=0$ & supported (degenerate) \\
\mbox{H-E1} rankings & $\rho\le 0.5$ & $\rho\le 0.5$ & $\rho\le 0.5$ & supported \\
\mbox{H-E2} width growth & QRF, ensemble & QRF, ensemble & none & QRF and ensemble supported; GP falsified \\
\bottomrule
\end{tabular}
\end{table}

\paragraph{Reading.} The multiplier band is falsified in three different
ways. On the two droplet tables the unseen-fluid tier costs 1.32 and 1.86
times the development reference and the unseen device, which in both tables
also carries fluids absent from development, 5.03 and 4.95 times; on the LNP
compilation, whose development reference is itself a cross-paper grouped
error of 40.87\%, the unseen lipid, paper and preparation-method tiers cost
no more than that reference (0.80, 0.86 and 0.62). There is no fixed
multiplier: its size depends on the type of shift and on the heterogeneity
already present in the development reference. The retrospective
leave-one-group-out reference inside the development set (0.77, 0.99 and
0.75 of dev-ref) understates both sealed device tiers. The unbounded
raw-scale perceptron returns negative diameters on 103 of the 109 R2-A1
device rows; the log-target twin avoids them there (97.5\% versus 2286\%)
and on the LNP tiers, but on the A1 device tier it extrapolates in log space
to $1.66\times10^{6}$\% against 1134\% for the raw-scale model, so H-B is
falsified at the protocol level although it holds on both new tables. H-C
stays undetermined: the paired difference between native and dimensionless
inputs is $+0.28$~pp ($p=0.45$) across the 13 A1 families and rests on two
pairs on R2-A1, where the dimensionless set is 20~pp better on the sealed
device tier (47.2\% versus 67.48\%) only descriptively; no table has twenty
input pairs that share dimensionless groups at different absolute scale, so
\mbox{H-C2} is not determinable. Interval-contained retrieval answers no query on any
of the six S2/S3 tiers ($A=W=0$). The six-model ranking under random splits
fails to predict the sealed ranking in all three further tables (Spearman $\rho$ of
$-0.77$ / $-0.14$, $0.37$ / $-0.71$ and $-0.54$ / $0.20$ on S2/S3), the
pre-registered success condition of \mbox{H-E1}.

\subsection{Q2 --- How many measurements does the loop need?}
\label{sec:q2}
Q2 is answered twice: once on the revealed sealed tiers with the constructive
protocols of Stage~3 and the pre-registered few-shot meta-analysis, and once
inside the offline closed loop of Section~\ref{sec:fw-offline}, where the
measurement budget is read off budget curves and the number of calibration
rows is itself a policy choice.

\subsubsection{R10, constructive: what a few target rows recover}
\label{sec:r10constructive}
\paragraph{Constructive experiments with frozen protocols (post hoc).} A
zero-shot evaluation shows how far a frozen surrogate drifts; in
practice a few target-domain conditions can usually be measured, and Stage~3
asks what such labels recover. Four protocols were frozen before their runs
and are reported whatever their sign. (i) Small-sample recalibration on the two revealed sealed sets:
on \tox{} a one-parameter log shift fitted on $k=2$ target rows lowers the
median MAPE over 200 draws from 19.86\% (the uncalibrated model on the rows left
unmeasured) to 13.91\% but never below the 11.4\%
semi-empirical line, whereas an XGBoost trained from scratch on $k\ge 10$
target rows reaches 10.78\% ($k=10$) and 6.83\% ($k=20$); on laboratory B no
calibration helps at any $k\le 20$ (scratch training 60.03\% at $k=20$
against 43.4\% for the uncalibrated model on the same rows). (ii) Warm-start residual boosting (shallow
trees on the source model's log prediction as base margin, with
0.1-weighted source rows): on \tox{} the median MAPE is 8.93\% at $k=2$,
7.60\% at $k=5$ and 4.21\% at $k=20$, below affine calibration and scratch
training at every comparable $k\ge 3$; on laboratory B its median falls below the
uncalibrated model's at some $k$ (36.63\% versus 43.27\% at $k=5$) but with a
wide tail, and the paired comparison finds it better than no calibration,
affine calibration or scratch training at no $k$; on
the five new sealed tiers the only statement the rule allows is that it beats
scratch training on the R2-A1 fluid-pair tier, while on the R2-A1 device tier
scratch training is best (15.39\% versus 19.69\% at $k=20$). (iii) A
conformal bias-hedging portfolio that spreads $K$ retrieval sub-targets over
the grouped conformal radius: on the A1 fluid-family tier Success@3 rose from
0.30 (greedy top-$K$) to 0.70 ($\Delta=+0.40$ [0.23, 0.58]) and Success@5
from 0.55 to 0.775; in the other three replay units, and in four further
units screened by the pre-registered bias condition
$|\overline{\log y-\log\hat y}|\le q$, it did not meet the margin, on two LNP
tiers it was worse than greedy ($\Delta=-0.16$ and $-0.28$ at $K=5$), and the
pre-registered cross-unit test against bias dominance was null
($\rho=0.00$), so no usability claim is made. (iv) A non-equilibrium
interfacial-tension feature for D2 exited at its feasibility gate: the
adsorption parameters are not available for the six chemistry groups. Taken
together, two target rows recovered development-level error on one chemistry
shift, no adaptation recovered it on the laboratory or device shifts, and the
methods are reported as candidates for pre-registered evaluation rather than
as recommendations.

\begin{table}[t]
\centering\small
\caption{Meta-analysis of few-shot adaptation under a frozen protocol over 106
leave-one-group tiers in five tables. Top: Spearman correlation across tiers
between each decision-time diagnostic and the warm-start gain (median over
100 draws of the MAPE reduction against the uncalibrated source model); an
asterisk marks a table-block bootstrap interval excluding zero after
Benjamini--Hochberg with the same sign within every table. Bottom: mean
held-out regret (percentage points above the post-hoc best method) of the
bias-ratio threshold rule fitted on four tables and applied to the fifth,
against two baselines. A tier enters at a given $k$ when at least ten of its
rows remain unmeasured.}
\label{tab:hj}
\begin{tabular}{lrrr}
\toprule
 & $k=2$ (106 tiers) & $k=5$ (81) & $k=10$ (67) \\
\midrule
Bias ratio of the $k$ log residuals & 0.796* & 0.717* & 0.667* \\
Sign consistency of the $k$ residuals & 0.509* & 0.645* & 0.626* \\
Residual magnitude / source CV error & 0.387* & 0.488* & 0.548* \\
5-NN support overlap (label-free) & $-0.071$ & $-0.096$ & $-0.131$ \\
Quantile-interval width ratio (label-free) & $-0.026$ & $-0.117$ & $-0.083$ \\
\midrule
Regret: never calibrate & 11.30 & 15.89 & 15.71 \\
Regret: always warm-start & 1.97 & 2.60 & 1.63 \\
Regret: bias-ratio rule & 1.75 & 2.80 & 1.80 \\
\bottomrule
\end{tabular}
\end{table}

\paragraph{When does few-shot adaptation help? (post hoc meta-analysis under a frozen protocol).}
The seven Stage~3 tiers cannot say which target domains benefit from a few
labels, so a protocol frozen before the run
(\path{scaleup/validation_plan/H_J_FEWSHOT_RULE_PROTOCOL_v1.md}) constructed
every leave-one-group tier with at least 12 rows across D1, D2, A1, R2-A1 and
R2-A2 (106 tiers: literature pools, geometry groups and emulsion types on
D1, chemistry groups on D2 and A1, devices on A1 and R2-A1, fluid pairs on
R2-A1, lipids, source papers and preparation classes on R2-A2), fitted the
table's locked XGBoost on the remaining rows, and drew $k\in\{2,5,10\}$
target rows 100 times per tier for the four methods of the constructive
experiments (uncalibrated, log-affine, warm-start residual boosting,
scratch). The gain of a method is the median over draws of its MAPE
reduction against the uncalibrated source model. Warm-start gains were
positive in 70.8\%, 85.2\% and 88.1\% of tiers at $k=2$, 5 and 10 (median
gains 1.53, 3.15 and 4.60 percentage points), log-affine gains in 37.7\%,
67.9\% and 77.6\%, and scratch gains in 44.4\% and 52.2\% at $k=5$ and 10.
Five decision-time diagnostics were pre-registered, computed only from the
$k$ drawn rows, the source model and the target inputs. Three of them (the
bias ratio of the $k$ log residuals ($|$mean$|$/SD), their sign consistency
and their magnitude relative to the source model's cross-validated error)
correlate with the warm-start gain across tiers at every $k$
(Table~\ref{tab:hj}; Spearman 0.796, 0.717 and 0.667 for the bias ratio,
table-block bootstrap intervals excluding zero after Benjamini--Hochberg,
same sign within each table); the two label-free diagnostics (5-NN support
overlap and quantile-interval width ratio) do not. The pre-registered rule
test then asked whether a threshold on a diagnostic, fitted on four tables
and applied to the fifth, lowers the regret relative to the post-hoc best
method against both ``always warm-start'' and ``never calibrate''. It does so
only at $k=2$ (bias-ratio rule: mean regret 1.75 versus 1.97 and 11.30
percentage points; difference to always-warm-start $-0.22$ [$-0.45$,
$-0.03$]); at $k=5$ and $k=10$ always-warm-start already has a mean regret
of 2.60 and 1.63 points and no rule on any of the five diagnostics improves
on it. By the pre-registered
criterion (two of three $k$) the rule does not hold: in these five tables a
decision-time diagnostic computed from the $k$ measured rows predicted the
size of the warm-start gain, but choosing between warm-start and no
calibration by a threshold on it beat always applying warm-start only with
two labels. Pooled pre-training across the four droplet tables, planned as a
further experiment, was stopped at its feasibility gate
(\path{scaleup/validation_plan/H_I_POOL_FEASIBILITY_GATE.md}): A1 reports
plug lengths rather than droplet diameters and the R2-A1 deposit states no
flow-rate or diameter units, so only D1 and D2 align on the intersection
features.

\subsubsection{E0--E2: simulator validity, budget curves and the number of calibration rows}
\label{sec:e2}
\paragraph{Simulator (E0).} Of the 107 tiers, 18 are dense by the
median nearest-row gap of 0.05 and the rest sparse. Replicate rows (identical inputs)
give relative noise floors of D1 0.035, D2 0.095, D3 0.102, A1 0.374 and R2-A2 0.201 (R2-A1 has no
replicate rows). The floors of A1 and D2 rest on 3 and 4 replicate groups and lie above most of the
adapted errors of their tables. The protocol's rule that no policy is credited with an error below the
floor was not applied in E2 (deviation E-D13): read strictly, it would count as unreached the 22
chemistry tiers of A1 and D2 whose $\varepsilon$ lies below the floor, and the warm-start share of
chemistry tiers reaching $\varepsilon$ would fall from 94\% to 27\% (all tiers: 76\% to 55\%).
Under a shuffled-label oracle the reference policy is not distinguishable from the floor policy (mean
$T^{*}$ difference $-0.05$ [$-0.63$, $+0.38$] measurements), so the simulator passes its validity
control; whether an individual gain is an artefact is tested in Section~\ref{sec:robust}. The density label replaced an eligibility gate of the
first frozen protocol after the uncounted smoke run and before any counted run (deviation E-D1,
Section~\ref{sec:eloop}); every E1 decision is therefore also recomputed on the 18 dense tiers alone
(the full table is in the repository; the decisions that change are reported below and in
Table~\ref{tab:e1bgrades}), and the E2 budget is reported by density.

\begin{figure}[htbp]
\centering
\includegraphics[width=0.98\linewidth]{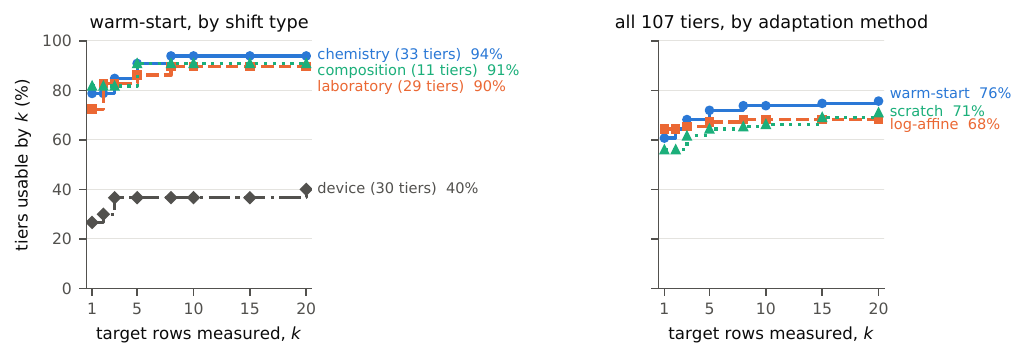}
\caption{Share of tiers whose warm-start surrogate is usable ($\le 1.25\times$ dev-ref on the unmeasured rows) after $k$ target rows: left, by shift type (number of tiers in brackets); right, all tiers by adaptation method. The values at $k = 20$ are those of Table~\ref{tab:e2budget}. Steps are read off the E2 budget curves (100 random draws per $k$, median over draws); the noise-floor rule of E0 is not applied (Section~\ref{sec:e2}).}
\label{fig:e2budget}
\end{figure}

\paragraph{Budget curves (E2).} Table~\ref{tab:e2budget} and Figure~\ref{fig:e2budget} give, per stratum, the fraction of tiers
whose adapted surrogate returns to $\varepsilon = 1.25 \times$ dev-ref within 20 target rows and the
90th percentile of the budget $k^{*}(\varepsilon)$. With warm-start residual boosting
76\% of the 107 tiers reach $\varepsilon$ (log-affine
68\%, scratch 71\%). The budget depends on the kind of shift:
94\% of the 33 chemistry tiers reach it with a 90th-percentile budget of
5.0 rows, 90\% of the 29 laboratory tiers
(10.6 rows) and 91\% of the 11 composition
tiers (5.0 rows), but only 40\% of the 30 device tiers,
whose median $k^{*}$ lies beyond the budget. Two features of the distribution matter for a
practitioner. It is bimodal: many tiers are already within $\varepsilon$ at $k = 1$, because their
uncalibrated error is near the usable level, while the device tiers mostly never reach it within
20 rows, partly because the D1 geometry tiers have a small development reference (median
5.4\% against 25.6\% for chemistry), so $\varepsilon$ is strict in absolute terms.
And density matters as much as shift type: 94\% of dense tiers reach $\varepsilon$
(90th percentile 8.0 rows) against 72\% of sparse tiers.

\paragraph{Calibration rows before searching (E1).} The reference loop (measure two random rows,
warm-start, propose greedily) reaches a target within $\tau = 0.10$ in 3.31 measurements on
average (median 2.87; success 99.5\%); the ceiling, a model refitted on every
tier label, needs 1.23, the floor without adaptation 3.79
(Table~\ref{tab:e1policy}). Spending measurements on calibration before searching does not pay:
starting the warm-started search at once ($k_0 = 0$, P2) needs 2.80 measurements,
$-0.50$ [$-0.87$, $-0.10$] relative to the reference and better after multiplicity control, while five
calibration rows (P3) cost $+0.44$ [$-0.07$, $+0.95$], not distinguishable. The gain of $k_0 = 0$ is
carried by the sparse tiers: there P0, P1 and P2 need 3.73, 3.48 and
2.90 measurements, on the dense tiers 4.06, 2.45 and 2.31,
and on the dense tiers alone the P2 difference is $-0.14$ [$-0.67$, $+0.45$], not
distinguishable and fragile under drift (Table~\ref{tab:e1bgrades}). On densely sampled tiers the
two calibration rows cost little; elsewhere the loop should start searching immediately, and the
same holds on hard tail targets (Section~\ref{sec:robust}).

\begin{table}[htbp]
\centering\footnotesize
\caption{E2 model task: fraction of tiers whose adapted surrogate returns to $\varepsilon = 1.25 \times$ dev-ref within 20 target rows, and the 90th percentile over tiers of the budget $k^{*}(\varepsilon)$ (rows), by stratum and adaptation method (100 random draws per $k$; median over draws). A 90th percentile of 21.0 means that at least a tenth of the tiers did not reach $\varepsilon$ within the budget. The noise-floor rule of E0 is not applied (Section~\ref{sec:e2}). Emulsion-type and preparation strata have two tiers each.}
\label{tab:e2budget}
\begin{tabular}{lrccc}
\toprule
Stratum & tiers & log-affine: reached / $k^{*}_{90}$ & warm-start: reached / $k^{*}_{90}$ & scratch: reached / $k^{*}_{90}$ \\
\midrule
all tiers & 107 & 68\% / 21.0 & 76\% / 21.0 & 71\% / 21.0 \\
chemistry & 33 & 88\% / 17.8 & 94\% / 5.0 & 94\% / 15.0 \\
laboratory & 29 & 83\% / 21.0 & 90\% / 10.6 & 86\% / 21.0 \\
composition & 11 & 82\% / 21.0 & 91\% / 5.0 & 82\% / 21.0 \\
device & 30 & 30\% / 21.0 & 40\% / 21.0 & 30\% / 21.0 \\
emulsion type & 2 & 50\% / 19.2 & 50\% / 20.4 & 50\% / 19.2 \\
preparation & 2 & 50\% / 19.0 & 50\% / 19.0 & 50\% / 19.0 \\
dense & 18 & 89\% / 9.8 & 94\% / 8.0 & 94\% / 9.5 \\
sparse & 89 & 64\% / 21.0 & 72\% / 21.0 & 66\% / 21.0 \\
\bottomrule
\end{tabular}
\end{table}

\subsubsection{E5: do planned budgets transfer across tables?}
\label{sec:e5}
A laboratory facing a new target domain has to plan how many measurements to allow before it has
any. E5 asks whether a planning rule learned on other tables does that job
(Table~\ref{tab:e5transfer}). Of the 107 tiers, 81 reach $\varepsilon$ within 20 rows and
26 do not; the cost is computed on the former. Planning the full budget costs 18.07 rows per tier
beyond need. A budget planned from the shift type alone, learned on four tables and applied to the
fifth, costs 4.86 rows against 2.78 when learned on the held-out table itself: the transfer
regret $+2.09$ [$-0.48$, $+7.62$] has an interval containing zero, and the rule saves
13.21 rows per tier against the full budget. Refining the rule by density, by the label-free
support diagnostic or by the two-row bias ratio fits the training tables better but does not transfer:
each regret interval lies above zero, although each refined rule still beats the full budget. The
verdicts are unchanged when the planning quantile is 0.50 or 0.90. The shift-type rule also flags
77\% of the 26 tiers that never reach $\varepsilon$ (mostly device tiers) as not reachable, at
28\% false alarms among the others. The quantity that transfers is therefore the coarse one: what
kind of shift the new domain represents.

\begin{table}[htbp]
\centering\footnotesize
\caption{E5: planned budgets under five-fold table rotation. Cost = rows measured beyond the budget the
tier needed, or the full budget of 20 when the plan fell short (mean over the 81 tiers that reach
$\varepsilon$ within 20 rows). ``Transferred'' rules are learned on four tables and applied to the fifth;
``table-specific'' rules are learned on the held-out table itself. Regret = transferred $-$
table-specific; intervals are table-level block bootstraps.}
\label{tab:e5transfer}
\setlength{\tabcolsep}{3pt}
\resizebox{\linewidth}{!}{%
\begin{tabular}{@{}lrrllrl@{}}
\toprule
Rule & transferred & table-specific & transfer regret [95\%] & transferred $-$ R0 [95\%] & under-planned & verdict \\
\midrule
R0 full budget & 18.07 & 18.07 & --- & --- & 0\% & no advantage \\
R1 shift type & 4.86 & 2.78 & $+2.09$ [$-0.48$, $+7.62$] & $-13.21$ [$-15.71$, $-6.50$] & 14\% & transfers \\
R2 shift type $\times$ density & 4.77 & 2.31 & $+2.46$ [$+0.50$, $+7.53$] & $-13.31$ [$-15.71$, $-7.19$] & 14\% & does not transfer \\
R3 shift type $\times$ support diagnostic & 4.32 & 2.48 & $+1.84$ [$+0.08$, $+6.05$] & $-13.75$ [$-15.74$, $-8.18$] & 17\% & does not transfer \\
R4 shift type $\times$ bias ratio & 5.28 & 2.36 & $+2.93$ [$+0.44$, $+9.17$] & $-12.79$ [$-15.50$, $-5.55$] & 19\% & does not transfer \\
\bottomrule
\end{tabular}}
\end{table}
\subsection{Q3 --- What should be measured first?}
\label{sec:q3}
The acquisition arm of E1 asks whether a label-free rule for choosing the first rows beats
random choice (Table~\ref{tab:e1policy}). None does by more than a tenth of a measurement on the 107
tiers: farthest-point seeding costs $+0.12$ [$-0.01$, $+0.29$], seeding at the rows with the widest
source interval $+0.21$ [$-0.15$, $+0.53$], both not distinguishable from random; $k$-means seeding
is nominally better ($-0.075$ [$-0.137$, $-0.027$] measurements), but the sign reverses under a
$\pm 20\%$ oracle drift. With the portfolio proposal the ordering is the same or worse
(farthest-point seeding with five rows $+0.71$ and widest interval $+0.12$, both relative to random
seeding with two rows). On the dense tiers $k$-means seeding is better ($-0.34$ [$-0.51$, $-0.19$]
measurements) and stable under drift, and farthest-point seeding is worse ($+0.42$). The negative control of E6 removes this
gain: replayed on permuted labels, where no rule can know where the target lies, $k$-means
seeding is again ``better'' than random ($-0.22$ [$-0.51$, $-0.09$]), so by the pre-registered rule
every $k$-means gain is classed as an artefact of the replay (Section~\ref{sec:robust}). A post-hoc
check with three further permutations gives $+0.05$, $+0.24$ and $+0.13$, none better, so
the first permutation was one draw; the pre-registered classification is kept. On hard tail targets
no acquisition rule helps either (Section~\ref{sec:robust}).

The proposal rule is part of the same question. Hedging the proposal with a retrieval portfolio
that cycles the target through the source model's residual scale is worse than the greedy
nearest-prediction proposal: $+0.39$ [$+0.11$, $+0.89$] with random seeding and $+0.30$ with
widest-interval seeding (not distinguishable); farthest-point seeding with five rows and the
portfolio needs $+0.66$ against random seeding with five rows and greedy proposals, a contrast that
also changes the acquisition rule. When a single recorded row can hit the target, spreading
proposals wastes measurements; the portfolio was designed for the retrieval setting of
Section~\ref{sec:r10constructive}, where it helped in one unit only.

\begin{table}[htbp]
\centering\scriptsize
\caption{E1 point task on the 107 tiers: the 14 loop policies, their mean and median time-to-target $T^{*}$ (measurements; three targets $\times$ 10 seeds per tier; $\tau = 0.10$, $B = 20$), the success rate within the budget, and the difference to the reference policy P1 with its table-level block-bootstrap interval. Verdicts after Benjamini--Hochberg control over the 13 non-reference policies; ``fragile'' marks a sign change under a $\pm 20\%$ oracle drift. Adaptation methods as in Section~\ref{sec:adapt}; the preview rule ($\dagger$) is the H-J threshold of Table~\ref{tab:hj}: warm-start when the two-row bias ratio exceeds 0.63, otherwise no adaptation. CEIL is the ceiling policy (Section~\ref{sec:eloop}).}
\label{tab:e1policy}
\setlength{\tabcolsep}{3pt}
\resizebox{\linewidth}{!}{%
\begin{tabular}{@{}lclllrrrll@{}}
\toprule
Policy & $k_0$ & acquisition & adaptation & proposal & mean $T^{*}$ & median & success & $\Delta$ vs P1 [95\%] & verdict \\
\midrule
P0 & 0 & --- & none & greedy & 3.79 & 2.67 & 98.8\% & $+0.48$ [$-0.38$, $+1.47$] & n.d. \\
P1 (reference) & 2 & random & warm-start & greedy & 3.31 & 2.87 & 99.5\% & --- & --- \\
P2 & 0 & --- & warm-start & greedy & 2.80 & 2.00 & 100.0\% & $-0.50$ [$-0.87$, $-0.10$] & better \\
P3 & 5 & random & warm-start & greedy & 3.75 & 3.43 & 99.8\% & $+0.44$ [$-0.07$, $+0.95$] & n.d. \\
P4 & 2 & farthest-point & warm-start & greedy & 3.43 & 3.00 & 99.7\% & $+0.12$ [$-0.01$, $+0.29$] & n.d. \\
P5 & 2 & $k$-means & warm-start & greedy & 3.23 & 2.67 & 100.0\% & $-0.075$ [$-0.137$, $-0.027$] & better, fragile \\
P6 & 2 & widest interval & warm-start & greedy & 3.51 & 3.00 & 99.7\% & $+0.21$ [$-0.15$, $+0.53$] & n.d. \\
P7 & 2 & random & log-affine & greedy & 3.79 & 3.20 & 99.0\% & $+0.49$ [$+0.33$, $+0.86$] & worse \\
P8 & 5 & random & scratch & greedy & 4.10 & 3.80 & 99.6\% & $+0.79$ [$+0.13$, $+1.62$] & worse \\
P9 & 2 & random & warm-start & portfolio & 3.70 & 2.97 & 99.1\% & $+0.39$ [$+0.11$, $+0.89$] & worse \\
P10 & 5 & farthest-point & warm-start & portfolio & 4.41 & 4.00 & 98.1\% & $+1.11$ [$+0.93$, $+1.43$] & worse \\
P11 & 2 & widest interval & warm-start & portfolio & 3.81 & 3.00 & 100.0\% & $+0.51$ [$-0.03$, $+0.98$] & n.d. \\
P12 & 2 & random & preview rule$^{\dagger}$ & greedy & 3.48 & 2.97 & 99.1\% & $+0.17$ [$+0.04$, $+0.31$] & worse \\
CEIL & 0 & --- & refit on all tier labels & greedy & 1.23 & 1.00 & 100.0\% & $-2.07$ [$-2.76$, $-1.54$] & better, fragile \\
\bottomrule
\end{tabular}}

\end{table}
\subsection{Q4 --- Keep adapting, stop, switch or escalate?}
\label{sec:q4}
Q4 has three parts: whether a change of model family or an output constraint
helps under shift (R6, R8), which adaptation method the loop should run once it
has target rows (E1), and whether a rule computed from the first measured rows
should change the action (H-J, E3). The robustness of the closed-loop answers
to oracle drift and to a harder task closes the section (E1b).

\subsubsection{R6: V2 retrospective results --- a negative/mixed verdict}
\label{sec:r6}
\begin{figure}[t]
\centering
\begin{tikzpicture}
\begin{axis}[
  width=0.8\linewidth, height=6cm, ybar, bar width=14pt,
  ylabel={V2-P minus B1 MAPE (pp)},
  xlabel={Held-out chemistry group (CMC, mM)},
  xtick=data, xticklabels={{0\\($n=63$)}, {2\\($n=30$)}, {3.5 (\tox{})\\($n=87$)}, {4.33\\($n=15$)}, {11\\($n=178$)}, {20\\($n=93$)}},
  x tick label style={font=\small, align=center},
  ymajorgrids, grid style={dashed,gray!30},
  nodes near coords, nodes near coords style={font=\scriptsize,/pgf/number format/fixed,/pgf/number format/precision=2},
  enlarge x limits=0.12,
]
\addplot+[fill=gray!60, draw=black] table[col sep=comma, x expr=\coordindex,
  y=V2_P_minus_B1_mape_percentage_points]{data/paired_group_differences_v2p_vs_b1.csv};
\draw[black] (axis cs:-0.5,0) -- (axis cs:5.5,0);
\end{axis}
\end{tikzpicture}
\caption{Negative/mixed result. Paired difference in held-out-group MAPE,
V2-P minus B1, for the six chemistry groups (negative favours V2-P). The
\tox{} group is a post-reveal retrospective analysis, not a blind test. V2-P
wins only two groups, failing locked condition~2.}
\label{fig:v2paired}
\end{figure}
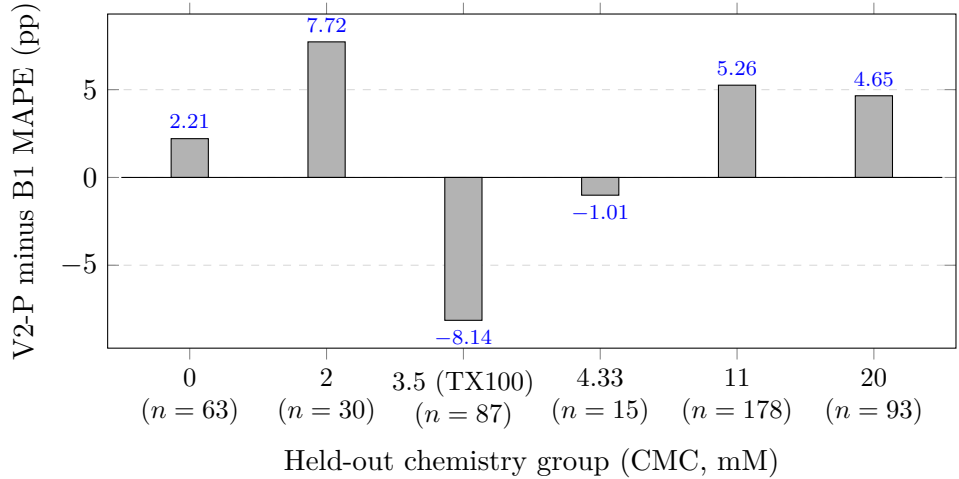

Two complete canonical runs in separate directories produced byte-identical
artifact-hash manifests. Table~\ref{tab:v2} gives the model matrix. Against
the frozen decision rule, condition~1 fails: V2-P chemistry-macro MAPE is
12.0496\% versus 10.2680\% for B1. Condition~2 fails: V2-P beats B1 in only
2 of 6 held-out groups (CMC 3.5~mM, the post-reveal \tox{} group,
$-8.138$~pp; CMC 4.33~mM, $-1.011$~pp; four groups required;
Figure~\ref{fig:v2paired}). Conditions 3--6 hold: V2-P's worst group is
14.847\% versus 19.960\% for B1; V2-H, V2-M and V2-P have zero point and
interval physical-invalid rates compared with 3.22\% for B1 and 13.73\% for the
unconstrained B3; the retrospective search reports coverage, recall and
precision together; no outer outcome was used for selection. The locked decision file
records \texttt{beneficial\_claim\_allowed = false}: V2-P does not meet the
pre-registered criteria for overall superiority, so the verdict is mixed.
Hard structural bounds do not reduce chemistry-macro error relative to B1,
but they eliminate all physically invalid point predictions in these
retrospective analyses and narrow the worst-group error.

Read descriptively, Table~\ref{tab:v2} orders the variants as follows: B1
(published four-input XGBoost) 10.27\% chemistry-macro MAPE with worst group
19.96\%; B2 (native-input XGBoost) 9.13\% and 16.94\%; B3 (unbounded
native-plus-physics perceptron) 16.64\%, 32.50\% and 13.73\% invalid
points; V2-H (the same perceptron with the $(0,D)$ bound) 10.59\%, 14.63\%
and no invalid point; V2-P (bound plus anchor and residual penalty) 12.05\%,
14.85\% and no invalid point. V2-P's largest gain over B1 is the post-reveal
\tox{} group ($-8.138$~pp), to which the anchor-provenance caveat of
Section~\ref{sec:v2protocol} applies. We report this progression as
observed; which ingredient accounts for which step is not identified by this
matrix and is left to a pre-registered test rather than inferred here.

The native-input XGBoost ablation B2 reaches 9.13\% chemistry-macro MAPE
compared with B1's 10.27\%; this is reported as an ablation under one protocol and
is not attributed to V2. V2's bounded LOCO out-of-fold uncertainty heuristic
reaches a mean held-out-group coverage of 0.744 (V2-P) as a transfer stress
test with no nominal guarantee. The retrospective constrained retrieval
abstained on almost every query: under the 10\% relative tolerance B3 and V2-H
answered 1 of 78 queries and every other model answered none; under the
0.01~mm absolute tolerance no model answered any query. We report this as a
null search endpoint rather than omitting it.

This near-complete abstention reflects a
width--tolerance tension in these data. The locked rule answered a query only
when the top-ranked candidate's point prediction lay within the tolerance and
its interval lay entirely inside the tolerance band (a containment rule
recorded in the frozen code); V2-P's held-out intervals average 67.5~\um{} in
width (B3 84.0~\um, V2-H 72.7~\um; retrospective check,
Section~\ref{sec:posthoc}), several times the $\pm 0.01$~mm band, so
certification by interval inclusion could succeed almost nowhere, and the
near-zero query coverage follows from the interval widths rather than from the
point predictions alone.

\begin{table}[t]
\centering\small
\caption{Frozen V2 model matrix under nested leave-one-chemistry-out
evaluation (466 rows, six groups, five seeds; \tox{} is post-reveal
retrospective).}
\label{tab:v2}
\resizebox{\linewidth}{!}{%
\begin{tabular}{lrrrrr}
\toprule
Model & Macro MAPE & Worst group & Row MAPE & Point invalid & Interval invalid \\
\midrule
B0 anchor equation & 18.38\% & 28.31\% & 13.99\% & 1.29\% & --- \\
B1 four-input XGBoost & 10.27\% & 19.96\% & 9.12\% & 3.22\% & --- \\
B2 native-input XGBoost & 9.13\% & 16.94\% & 8.86\% & 2.58\% & --- \\
B3 native+physics MLP (unbounded) & 16.64\% & 32.50\% & 14.45\% & 13.73\% & 0 \\
V2-H bounded MLP & 10.59\% & 14.63\% & 10.72\% & 0 & 0 \\
V2-M V2-P + monotonic penalty & 12.05\% & 14.84\% & 11.30\% & 0 & 0 \\
V2-P bounded residual + anchor (primary) & 12.05\% & 14.85\% & 11.30\% & 0 & 0 \\
\bottomrule
\end{tabular}}
\end{table}

\subsubsection{E1 and E3: adaptation methods, the preview rule and a priced decision}
\label{sec:e3}
\paragraph{Which adaptation the loop should run (E1).} Once target rows exist, warm-start
residual boosting of the source model dominates the alternatives in the point task: the two-parameter
log-affine correction costs $+0.49$ [$+0.33$, $+0.86$] measurements and a model trained from scratch on the
measured rows $+0.79$ [$+0.13$, $+1.62$], both worse (Table~\ref{tab:e1policy}). The H-J threshold
rule, run as a preview with its in-sample threshold (warm-start if the two-row bias ratio exceeds
0.63, otherwise leave the source model alone), is worse than always warm-starting by
$+0.17$ [$+0.04$, $+0.31$]. On the dense tiers alone the log-affine correction stays worse
($+0.54$ [$+0.22$, $+1.11$]) while scratch training is not distinguishable
($+0.30$ [$-0.08$, $+0.84$]).

\paragraph{A priced decision after two measurements (E3).} E3 asks the question the preview could
not. Four actions are available after the two phase-A rows: continue warm-starting, stop
adapting, switch to a model fitted on the target rows alone, or escalate to a full
recalibration (the ceiling model, refitted on every tier label and so an oracle-informed, favourable
version of escalation) at a cost of $\lambda$ measurement-equivalents. Does any threshold
rule on the two-row diagnostics (bias ratio, residual magnitude, source interval width, support
overlap) beat the constant actions once it is learned on four tables and applied to the fifth?
By Table~\ref{tab:e3lambda}, no rule does so except when escalation is free. At $\lambda = 0$ every fold learns
``always escalate'' and the rule is better than always-continue ($-0.96$ [$-1.37$, $-0.48$]
measurements), but the verdict is fragile: under a $\pm 20\%$ oracle drift the non-updating ceiling
model falls behind the adaptive loop ($+2.00$ / $+1.10$), as in
Section~\ref{sec:e1b}. At $\lambda = 5$ and 10 three folds learn ``always continue'' and two a high
bias-ratio threshold that switches to scratch training; the rotated rule costs 3.32
against 3.31 for always-continue, $+0.01$ [$+0.00$, $+0.03$], and the rule
learned on all 107 tiers without rotation is ``always continue''; under a
$\pm 20\%$ oracle drift the rotated rule stays slightly behind always continuing
($+0.08$ / $+0.06$). The rule does beat never
adapting ($-0.59$ [$-1.09$, $-0.24$]) and always switching ($-0.39$
[$-0.91$, $-0.09$]), because it mostly continues. E3 is therefore negative: the pre-registered
form (the rule holds at two of three $\lambda$ levels) is met at one level only, the trivial one.

These diagnostics do not reach the available headroom. A perfect per-unit decision would
save 0.54 measurements at $\lambda = 5$ and 0.49 at $\lambda = 10$ relative to always
continuing, and the after-the-fact best action differs from ``continue'' in 17.4\% of
the 3210 units. In those units stopping adaptation and switching to scratch training are about
equally often best: each attains the minimum in 11.7\% and 11.2\% of all units, tied with
each other in 7.3\%; the share of units with a better action than continuing is largest for
preparation (45.0\%) and emulsion-type (31.7\%) units, and escalation at $\lambda = 5$ attains
the minimum in 1.3\% to 5.0\% of units by type. The information that would tell those units apart is not in the bias ratio, the residual
magnitude, the interval width or the support overlap of two rows. Evaluated as one member of the
family, the preview threshold itself costs $+0.17$ [$+0.04$, $+0.30$] relative to always
continuing (E3's own resampling; [$+0.04$, $+0.31$] in E1), confirming the E1 preview. In this loop, on these tables, the decision after two
measurements is to keep adapting the source model; no cheap diagnostic tested here indicates
otherwise.

\begin{table}[htbp]
\centering\scriptsize
\caption{E3 decision rule after two random measurements, learned by five-fold table rotation and priced by the escalation cost $\lambda$ (measurement-equivalents). Costs are mean $T^{*}$ over the 107 tiers (three targets $\times$ 10 seeds), with $\lambda$ added to every escalation; the constant arms, which share the two phase-A rows, cost 3.31 (continue, warm-start), 3.91 (none) and 3.71 (switch) at every $\lambda$ (P0 of Table~\ref{tab:e1policy} starts without phase-A rows). ``Oracle'' is the best action chosen after the fact for every unit. The rule holds at $\lambda$ when it beats both ``always continue'' and ``never adapt'' with block intervals excluding zero. The rule's difference to ``always continue'' under drifted oracles is given in the text. Each interval comes from its own bootstrap resample, so identical rules can differ in the last digit.}
\label{tab:e3lambda}
\setlength{\tabcolsep}{3pt}
\resizebox{\linewidth}{!}{%
\begin{tabular}{@{}cp{0.34\linewidth}rrrlll@{}}
\toprule
$\lambda$ & fold rules (A1 / D1 / D2 / R2-A1 / R2-A2 held out) & rule & escalate $+\lambda$ & oracle & rule $-$ continue [95\%] & rule $-$ none [95\%] & holds \\
\midrule
0 & always escalate in all five folds & 2.34 & 2.34 & 2.31 & $-0.96$ [$-1.37$, $-0.48$] & $-1.57$ [$-2.48$, $-0.67$] & yes, fragile \\
5 & two folds: high bias ratio $\to$ switch, else continue; three folds: always continue & 3.32 & 7.34 & 2.77 & $+0.01$ [$+0.00$, $+0.03$] & $-0.59$ [$-1.09$, $-0.24$] & no \\
10 & as for $\lambda = 5$ & 3.32 & 12.34 & 2.82 & $+0.01$ [$+0.00$, $+0.03$] & $-0.59$ [$-1.10$, $-0.24$] & no \\
\bottomrule
\end{tabular}}

\end{table}

\subsubsection{E4: does the source-model family matter inside the loop?}
\label{sec:e4}
\begin{figure}[htbp]
\centering
\includegraphics[width=0.8\linewidth]{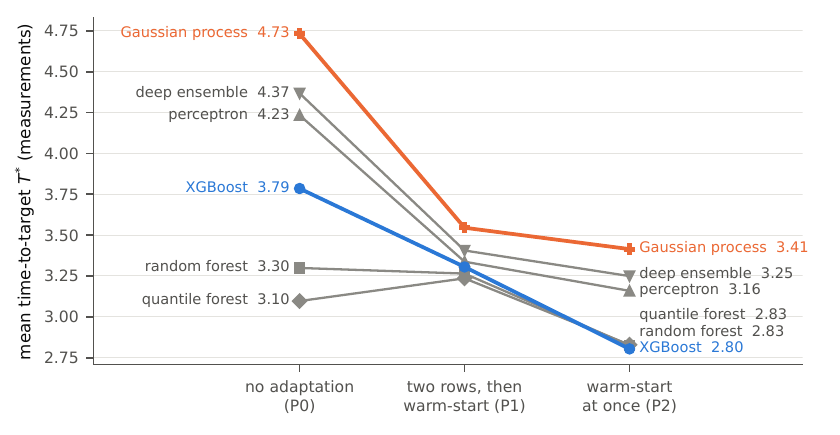}
\caption{Mean time-to-target of the six source-model families without adaptation (P0), after two random rows and warm-start (P1), and with warm-start from the first measurement (P2); 107 tiers, $\tau = 0.10$, $B = 20$. Adaptation removes most of the spread between families that the unadapted models show.}
\label{fig:e4slopes}
\end{figure}

The method matrices showed that model families differ under shift when nothing is measured
(Section~\ref{sec:r8}). E4 asks whether that difference survives inside the loop
(Table~\ref{tab:e4family}, Figure~\ref{fig:e4slopes}). Without adaptation (P0) it does: the quantile forest and the random
forest need 0.69 and 0.49 fewer measurements than XGBoost. Once the loop warm-starts on its
measurements the differences shrink or reverse: under P2 no family beats XGBoost, the forests are
indistinguishable from it, the perceptron and the deep ensemble trail by 0.36--0.45
measurements, and the Gaussian process, whose optimiser raised 739 convergence warnings, is worse by
$+0.61$ [$+0.32$, $+1.01$]. XGBoost has the lowest or tied-lowest $T^{*}$ on 42\% of tiers (the quantile forest on 36\%, the random forest and the perceptron on 31\%); a single family is strictly lowest on only 60 of the 107 tiers, XGBoost on 16\% of all tiers. The XGBoost arm
reproduces the E1 loops of P0, P1 and P2 exactly. The ranking a developer would compute from a random
split carries little of this: per tier, Kendall's $\tau_b$ between the random-split ranking of the six
families and their loop ranking averages $+0.08$ [$-0.04$, $+0.19$] and is positive on 49\%
of tiers, so it does not predict the loop ranking. It does agree, weakly, with the zero-shot ranking on
the shifted tier ($+0.18$ [$+0.03$, $+0.32$]), and even the label-informed zero-shot ranking
predicts the loop ranking only weakly ($+0.23$ [$+0.04$, $+0.40$]). Adaptation absorbs most of
what distinguishes the families; the choice that matters is to adapt.

\begin{table}[htbp]
\centering\footnotesize
\caption{E4: six source-model families inside the loop (107 tiers; three targets $\times$ 10 seeds per tier;
$\tau = 0.10$, $B = 20$). Mean $T^{*}$ under P0 (no adaptation), P1 (two random rows, warm-start) and P2
(warm-start at once); difference to XGBoost under P2 with table-level block interval and
Benjamini--Hochberg verdict; share of tiers where the family has the lowest or tied-lowest $T^{*}$ under P2 (ties are common); median
zero-shot MAPE on the tier rows.}
\label{tab:e4family}
\setlength{\tabcolsep}{3pt}
\resizebox{\linewidth}{!}{%
\begin{tabular}{@{}lrrrlrr@{}}
\toprule
Family & P0 & P1 & P2 & $\Delta$ P2 vs XGBoost [95\%] & lowest or tied (P2) & zero-shot MAPE \\
\midrule
XGBoost (reference) & 3.79 & 3.31 & 2.80 & --- & 42\% & 20.2\% \\
random forest & 3.30 & 3.26 & 2.83 & $+0.02$ [$-0.18$, $+0.17$] n.d., fragile & 31\% & 18.8\% \\
perceptron & 4.23 & 3.34 & 3.16 & $+0.36$ [$+0.01$, $+0.67$] n.d., fragile & 31\% & 23.5\% \\
quantile forest & 3.10 & 3.24 & 2.83 & $+0.02$ [$-0.07$, $+0.11$] n.d. & 36\% & 18.4\% \\
Gaussian process & 4.73 & 3.55 & 3.41 & $+0.61$ [$+0.32$, $+1.01$] worse & 22\% & 22.7\% \\
deep ensemble & 4.37 & 3.41 & 3.25 & $+0.45$ [$-0.02$, $+1.38$] n.d. & 24\% & 22.9\% \\
\bottomrule
\end{tabular}}
\end{table}
\subsubsection{Robustness of the closed-loop findings: drift and difficulty (E1b)}
\label{sec:e1b}
\paragraph{Oracle drift.} With every oracle label multiplied by $1 \pm 0.2$ and the target
expressed in the drifted scale, the ordering of the adaptive policies is unchanged: P2 needs
3.24 / 3.25 measurements, P1 3.47 / 3.40, P0
5.16 / 4.50, while the ceiling, a perfect prior that is never
updated, falls to 7.28 / 5.81, behind every warm-start policy. At
$\pm 0.5$ (descriptive) P1 needs 3.66 / 3.53 and P0
7.51 / 6.21. A loop that keeps measuring and updating is robust to a
constant scale error of the instrument; a frozen model, however good its prior, is not.

\paragraph{A harder task (E1b).} Section~\ref{sec:q2} to this point was obtained at $\tau = 0.10$ and
$B = 20$, where the median loop succeeds in under three measurements and differences are fractions of
a measurement. E1b repeats E1 at a halved tolerance, a halved budget, and both
(Table~\ref{tab:e1bgrades}); the reference loop then needs 4.75, 3.17 and
4.31 measurements, and the shuffled-label control is accepted in every cell. Every
decision above survives: starting the search at once is better in all three cells
($-0.66$, $-0.45$, $-0.50$, none fragile), no acquisition rule
is better than random in any cell, log-affine and scratch adaptation and the portfolio
proposal are worse in all three cells, and the ordering P2 $<$ P1 $<$ P0 holds under drift in all
three, with the ceiling again falling behind. On the dense tiers $k$-means seeding is ``better'' in
every cell ($-0.46$, $-0.27$, $-0.46$, none fragile), but the same policy is ``better'' under
shuffled labels (Section~\ref{sec:robust}), so these gains are not counted; starting at once is not
reliably better on the dense tiers.

\begin{table}[htbp]
\centering\scriptsize
\caption{E1b difficulty sensitivity: the E1 decisions recomputed at a halved tolerance and a halved budget, $(\tau, B)$ cells $(0.05, 20)$, $(0.10, 10)$ and $(0.05, 10)$ beside E1 itself. Entries are $\Delta$ vs P1 with table-level block intervals and the Benjamini--Hochberg verdict; a finding is ``robust to difficulty'' when it holds in all three new cells. The last two rows recompute the two acquisition-and-$k_0$ decisions on the 18 dense tiers alone (the original eligibility gate of the first frozen protocol, Section~\ref{sec:eloop}).}
\label{tab:e1bgrades}
\setlength{\tabcolsep}{3pt}
\begin{tabular}{@{}>{\raggedright\arraybackslash}p{0.175\linewidth}*{4}{>{\raggedright\arraybackslash}p{0.16\linewidth}}>{\raggedright\arraybackslash}p{0.11\linewidth}@{}}
\toprule
Decision & E1 $(0.10, 20)$ & $(0.05, 20)$ & $(0.10, 10)$ & $(0.05, 10)$ & grade \\
\midrule
P1 mean $T^{*}$ & 3.31 & 4.75 & 3.17 & 4.31 & --- \\
CEIL mean $T^{*}$: $\delta = 0$; $-0.2$ / $+0.2$ & 1.23; 7.28 / 5.81 & 1.55; 9.63 / 7.94 & 1.21; 6.17 / 5.07 & 1.45; 7.53 / 6.48 & robust to difficulty \\
F14: P2 ($k_0 = 0$) vs P1 & $-0.50$\newline \mbox{[$-0.87$, $-0.10$]}\newline better & $-0.66$\newline \mbox{[$-1.07$, $-0.40$]}\newline better & $-0.45$\newline \mbox{[$-0.76$, $-0.04$]}\newline better & $-0.50$\newline \mbox{[$-0.87$, $-0.20$]}\newline better & robust to difficulty \\
F15: P5 ($k$-means) vs P1 & $-0.075$\newline \mbox{[$-0.137$, $-0.027$]}\newline better, fragile & $-0.14$\newline \mbox{[$-0.40$, $+0.07$]}\newline n.d. & $-0.04$\newline \mbox{[$-0.12$, $+0.14$]}\newline n.d., fragile & $-0.07$\newline \mbox{[$-0.16$, $+0.08$]}\newline n.d., fragile & robust to difficulty \\
F16: P7 (log-affine) vs P1 & $+0.49$\newline \mbox{[$+0.33$, $+0.86$]}\newline worse & $+0.42$\newline \mbox{[$+0.29$, $+0.83$]}\newline worse & $+0.37$\newline \mbox{[$+0.26$, $+0.59$]}\newline worse & $+0.35$\newline \mbox{[$+0.26$, $+0.62$]}\newline worse & robust to difficulty \\
F16: P8 (scratch) vs P1 & $+0.79$\newline \mbox{[$+0.13$, $+1.62$]}\newline worse & $+1.11$\newline \mbox{[$+0.24$, $+2.37$]}\newline worse & $+0.73$\newline \mbox{[$+0.16$, $+1.38$]}\newline worse & $+0.94$\newline \mbox{[$+0.25$, $+1.76$]}\newline worse & robust to difficulty \\
F16: P9 (portfolio) vs P1 & $+0.39$\newline \mbox{[$+0.11$, $+0.89$]}\newline worse & $+0.77$\newline \mbox{[$+0.25$, $+1.85$]}\newline worse & $+0.26$\newline \mbox{[$+0.07$, $+0.61$]}\newline worse & $+0.45$\newline \mbox{[$+0.16$, $+1.06$]}\newline worse & robust to difficulty \\
dense tiers only: P5 vs P1 & $-0.34$\newline \mbox{[$-0.51$, $-0.19$]}\newline better & $-0.46$\newline \mbox{[$-0.97$, $-0.14$]}\newline better & $-0.27$\newline \mbox{[$-0.41$, $-0.16$]}\newline better & $-0.46$\newline \mbox{[$-0.83$, $-0.17$]}\newline better & artefact (shuffled-label control) \\
dense tiers only: P2 vs P1 & $-0.14$\newline \mbox{[$-0.67$, $+0.45$]}\newline n.d., fragile & $-0.61$\newline \mbox{[$-1.30$, $-0.08$]}\newline better, fragile & $-0.07$\newline \mbox{[$-0.57$, $+0.50$]}\newline n.d., fragile & $-0.48$\newline \mbox{[$-0.88$, $-0.10$]}\newline better, fragile & not reliable on dense tiers \\
\bottomrule
\end{tabular}
\end{table}

\subsection{Do the closed-loop answers survive negative controls and harder targets? (E6, E7)}
\label{sec:robust}
\subsubsection{E6: negative controls for every gain}
\label{sec:e6}
\begin{figure}[htbp]
\centering
\includegraphics[width=0.98\linewidth]{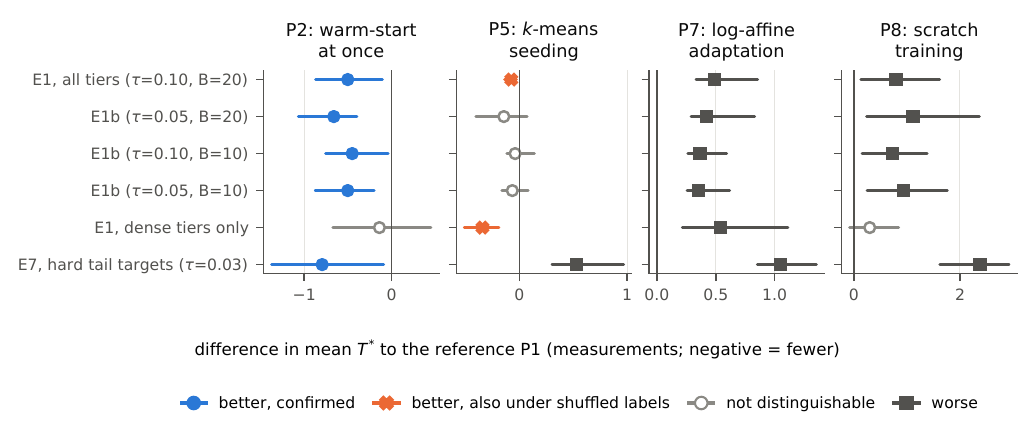}
\caption{Difference in mean $T^{*}$ to the reference P1 for four policies across the closed-loop settings, with table-level block intervals. Marker and colour give the status after both negative controls: better and confirmed (blue circle), better also under shuffled labels (orange cross), not distinguishable (open grey), worse (dark square).}
\label{fig:e6forest}
\end{figure}

On labels permuted within each tier no surrogate can know where the target lies, so no policy should
beat the reference. The reference needs 4.11 measurements there, every source-model policy between
3.89 and 4.36, and the analytic chance level of random proposals is 4.39. Eleven of the twelve
source-model policies are not distinguishable from the reference; $k$-means seeding is ``better'' by
$-0.22$ [$-0.51$, $-0.09$], and the ceiling, refitted on the permuted labels, is better by construction.
Table~\ref{tab:e6ledger} lists the fate of every ``better'' verdict of E1, E1b, E3 and E4, and Figure~\ref{fig:e6forest} shows four policies across all settings (20 in all). Four survive
both controls, and all four are the same finding (start the warm-started search at once), on all tiers
in E1 and in each of the three harder E1b cells. The two dense-tier P2 gains are fragile under drift,
every $k$-means gain fails the pre-registered shuffled-label control (three post-hoc permutations,
Section~\ref{sec:q3}, do not reproduce the shuffled gain), the free-escalation rule of E3 is fragile, and no
family of E4 is better than XGBoost. The loops beat chance where the surrogate carries information
(chemistry: P2 2.53 against a chance level of 5.68) and are at chance on the laboratory tiers (P1
3.16 and P2 3.52 against 3.18), where the source model's prior misleads the first proposals.

\begin{table}[htbp]
\centering\footnotesize
\caption{E6: every ``better'' verdict (Benjamini--Hochberg) of E1, E1b, E3 and E4, classified by the two
negative controls. Confirmed = sign unchanged under a $\pm 20\%$ oracle drift and not ``better'' under
shuffled labels; fragile = sign changes under the drift; artefact = also ``better'' under shuffled
labels; positive control = label-informed by construction.}
\label{tab:e6ledger}
\begin{tabular}{@{}p{0.34\linewidth}p{0.40\linewidth}p{0.18\linewidth}@{}}
\toprule
Gain & where it was found & status \\
\midrule
P2 warm-start at once & E1 and the three E1b cells, all tiers & 4 confirmed \\
P2 warm-start at once & E1b dense tiers, two cells & 2 fragile \\
P5 $k$-means seeding & E1 all and dense tiers; E1b dense tiers & 5 artefact \\
E3 rule ``always escalate'' at $\lambda = 0$ & E3 & 1 fragile \\
CEIL ceiling (refitted on the tier labels) & E1, E1b & 8 positive control \\
any source-model family vs XGBoost & E4 & none ``better'' \\
\bottomrule
\end{tabular}
\end{table}

\subsubsection{E7: the closed-loop questions on hard targets}
\label{sec:e7}
\begin{figure}[htbp]
\centering
\includegraphics[width=0.62\linewidth]{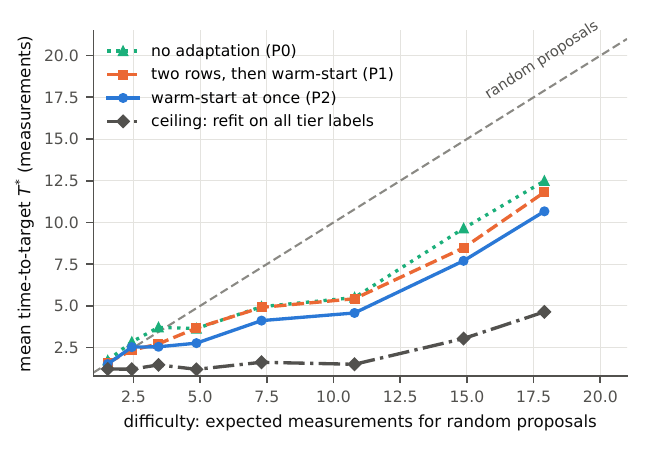}
\caption{Mean time-to-target against task difficulty, measured by the expected number of measurements a uniformly random proposer needs (dashed diagonal). Units of E1 (targets at the 20th, 50th and 80th label percentiles, $\tau = 0.10$) and E7 (targets at the 5th, 10th, 90th and 95th, $\tau = 0.03$) are pooled and binned by that difficulty (bin edges at 2, 3, 4, 6, 9, 13 and 17 measurements; markers at the bin means). On easy units every policy is close to chance; the gap between the warm-started loop and chance grows with difficulty.}
\label{fig:e7difficulty}
\end{figure}

E1 was too easy to resolve the acquisition and decision questions: at a 10\% tolerance around central
targets a median of about a quarter of a tier's rows hit the target, random proposals hit in about four
measurements, and on the easiest third of units no policy beat chance. E7 asks the same questions on
tail targets at a 3\% tolerance; Figure~\ref{fig:e7difficulty} places the units of E1 and E7 on one difficulty axis. Of 428 units, 301 in 92 tiers are hard (median hit fraction
5.4\%); random proposals need 11.35 measurements there, and the shuffled-label control is
accepted. Table~\ref{tab:e7hard} gives every policy on the hard units. The loop that starts at once and
warm-starts needs 5.72, which is 5.63 fewer than chance ($-5.63$ [$-7.64$, $-3.34$]), and is better than
the reference by $-0.79$ [$-1.37$, $-0.09$], neither fragile nor an artefact. Every other answer of E1
holds again. No label-free acquisition rule helps: farthest-point and widest-interval seeding are not
distinguishable (the latter also flagged by its shuffled replay) and $k$-means seeding is worse by
$+0.53$. The interval-aware proposal is slightly worse than the greedy one ($+0.16$
[$+0.10$, $+0.33$] relative to P2). Log-affine and scratch adaptation, five calibration rows and the portfolio
are worse ($+1.05$, $+2.38$, $+1.46$, $+0.78$).

With four measured rows instead of two, the decision rule still does not help. At $\lambda = 5$ and
10, four of five folds learn ``always continue'' and the rotated rule costs $+0.08$
[$+0.00$, $+0.39$] relative to always continuing; only free escalation wins ($\lambda = 0$), and that
verdict is fragile. The headroom is larger than with two easy rows (a perfect per-unit decision would
save 1.37 measurements at $\lambda = 5$, and 36.7\% of units prefer another action), but the bias
ratio, residual magnitude, sign consistency, interval width and support overlap of four rows do not
reveal which units those are. Only three tiers have extrapolative targets; there the unadapted model
fails most often, and three tiers are too few for a verdict. The null answers to Q3 and Q4
are therefore not artefacts of easy tasks.

\begin{table}[htbp]
\centering\scriptsize
\caption{E7 on the 301 hard units (tail targets, $\tau = 0.03$, $B = 20$; at most 10\% of a tier's rows within
tolerance of the target; 92 tiers; chance level 11.35 measurements). Mean $T^{*}$, success within the
budget, difference to P1 with table-level block interval and Benjamini--Hochberg verdict, fragility
under a $\pm 20\%$ oracle drift, and the shuffled-label check.}
\label{tab:e7hard}
\setlength{\tabcolsep}{3pt}
\begin{tabular}{@{}lrrlll@{}}
\toprule
Policy & mean $T^{*}$ & success & $\Delta$ vs P1 [95\%] & fragile & shuffled \\
\midrule
P0 no adaptation & 6.58 & 90\% & $+0.07$ [$-0.60$, $+0.88$] n.d. & no & --- \\
P1 two rows, warm-start (reference) & 6.52 & 94\% & --- & --- & --- \\
P2 warm-start at once & 5.72 & 93\% & $-0.79$ [$-1.37$, $-0.09$] better & no & --- \\
P3 five rows & 7.97 & 94\% & $+1.46$ [$+0.82$, $+1.94$] worse & no & --- \\
P4 farthest-point & 6.64 & 93\% & $+0.12$ [$-0.26$, $+0.41$] n.d. & yes & --- \\
P5 $k$-means & 7.05 & 93\% & $+0.53$ [$+0.30$, $+0.97$] worse & no & --- \\
P6 widest interval & 6.77 & 93\% & $+0.25$ [$-0.07$, $+0.45$] n.d. & yes & artefact \\
P7 log-affine & 7.56 & 89\% & $+1.05$ [$+0.85$, $+1.35$] worse & no & --- \\
P8 scratch & 8.89 & 91\% & $+2.38$ [$+1.62$, $+2.92$] worse & no & --- \\
P9 portfolio & 7.29 & 93\% & $+0.78$ [$+0.43$, $+1.29$] worse & no & --- \\
P13 interval-aware (new) & 5.88 & 94\% & $-0.63$ [$-1.22$, $+0.11$] n.d. & no & --- \\
CEIL ceiling & 2.31 & 98\% & $-4.20$ [$-5.53$, $-3.41$] better & yes & positive control \\
\bottomrule
\end{tabular}
\end{table}

\section{Synthesis: What the Evidence Says for a Self-Driving Loop}
\label{sec:synthesis}
The units of Section~\ref{sec:results} support twenty-three framework-level
findings. Each is stated once in Table~\ref{tab:findings} with the question it
answers, its evidence tier and the units that carry it; the sub-conclusion
registry in the repository (\path{scaleup/synthesis}) records the individual
statements they condense and the places where two units are in tension. Findings F1--F13 come from the sealed, pre-registered and frozen
programme on the six tables, F14--F23 from the offline closed loop. None is a
recommendation for a laboratory that has not run the same protocol on its own
data; each is a measured regularity of these tables under the framework's
definitions.

{\scriptsize
\setlength{\tabcolsep}{3pt}
\renewcommand{\arraystretch}{0.92}
\setlength{\LTcapwidth}{\linewidth}
\begin{longtable}{@{}p{0.04\linewidth}p{0.625\linewidth}p{0.075\linewidth}p{0.055\linewidth}p{0.15\linewidth}@{}}
\caption{Framework-level findings. Tier: S sealed, P pre-registered, F frozen
protocol with two runs, H post hoc under a frozen protocol, R retrospective.
Q: question (Q1--Q4), P provenance, M method. Units as in Table~\ref{tab:resultsmap}.}
\label{tab:findings}\\
\toprule
 & Finding & Q & tier & units \\
\midrule
\endfirsthead
\toprule
 & Finding & Q & tier & units \\
\midrule
\endhead
\bottomrule
\endlastfoot
F1 & The cost of shift is stratified by its type (geometry and literature, then chemistry, then device type and laboratory), and no fixed multiplier of the development reference describes it. & Q1 & S, P & R2, R7, R8, R10 \\
F2 & With no target label, no statistic of the inputs separates reliable from unreliable rows at the candidate or query level; the gains of label-free screening are source-level, and a size proxy matches them under literature and laboratory shift. & Q1 & F, R & R4, R7, R8, R9, R10 \\
F3 & Under shift, interval coverage is recovered only by widening (grouped calibration) or by abstaining (weighted conformal); no interval family keeps nominal coverage where supports do not overlap. & Q1 & F & R3, R7, R8, R10 \\
F4 & Random-split accuracy and model rankings do not predict accuracy or rankings under shift; grouped cross-validation predicts chemistry shift but understates device and laboratory shift. & Q1 & P & R7, R8, R10 \\
F5 & Output bounds remove physically invalid predictions and narrow the worst-group error, but do not bring chemistry-macro error below the published XGBoost baseline; the constraint form must match the flow regime. & Q4 & F & R6, R9, R10 \\
F6 & On the revealed sealed sets a few target labels recover chemistry shift but not the device tier or laboratory B; across the 107 tiers most chemistry and laboratory tiers recover within twenty rows and most device tiers do not; the gain is usually positive and its size depends on the domain. & Q2 & H, F & R10 constructive, E2 \\
F7 & A diagnostic computed from two measured rows (the bias ratio of the log residuals) predicts the size of the few-shot gain; label-free diagnostics do not; a threshold rule on it beats always warm-starting at $k = 2$ only, short of its two-of-three criterion. & Q4 & F & H-J \\
F8 & Where to measure first was the least answered question of the sealed programme: label-free space-filling selection did not beat random, applicability filtering before retrieval did not systematically help, and the retrieval portfolio helped in one unit. & Q3 & H & R10 constructive \\
F9 & Failure has two layers: a support gap visible from inputs, and a conditional shift inside the marginal ranges that no input-side screen sees. & Q1 & S & R5, R9 \\
F10 & The evaluation key is itself a result: the choice of grouping changes conclusions at every stage, from the development reference to the ranking of model families. & M & P, F & R7, R8, R10 \\
F11 & Sealing adds the exclusion of tuning and the isolation of modelling choices, at a documented cost: rules must be fixed before any label is read, and defects surface after the reveal. & M & S & R5, R9 \\
F12 & The surrogates are independent re-implementations whose fidelity to the upstream code is established up to one library default; the deployed upstream model was trained on the literature rows (shown by a post-hoc refit that reproduces it exactly; the pre-registered inference rule was inconclusive). & P & R & R1, Appendix~\ref{app:fidelity} \\
F13 & Model family matters under shift in a way random splits do not reveal: forests degrade least under chemistry shift, unbounded perceptrons and ensembles extrapolate by orders of magnitude, and every family's native uncertainty is beaten by a size control in at least one cell. & Q1, Q4 & F, P & R8, R10 \\
F14 & Measuring before searching does not pay on these tables: the best loop proposes at once and warm-starts on every measurement; over all tiers it is the one closed-loop gain that survives drift and shuffled-label controls, and it holds on hard tail targets, but it is carried by the sparse tiers (not distinguishable on the 18 dense ones) and is not better on the laboratory tiers (3.52 against 3.16). & Q2 & F & E1, E1b, E6, E7 \\
F15 & Label-free acquisition rules do not shorten the search, on all tiers or on hard targets; the apparent gain of $k$-means seeding is also present under the pre-registered shuffled-label control and is classed as an artefact of the replay (three post-hoc permutations do not reproduce it). & Q3 & F & E1, E1b, E6, E7 \\
F16 & In the loop, warm-start residual boosting dominates the log-affine correction, scratch training (not distinguishable on the dense tiers) and the threshold preview; hedged portfolio and interval-aware proposals are not better than greedy ones. & Q4 & F & E1, E1b, E7 \\
F17 & Under a $\pm 20\%$ drift of the oracle scale the ranking of adaptive policies is unchanged, while a perfect prior that is never updated falls behind every warm-start policy. & Q4 & F & E1, E1b, E3 \\
F18 & Within twenty target rows warm-start returns the surrogate to $1.25 \times$ its development reference in most chemistry, laboratory and composition tiers and in few device tiers (the chemistry share falls to 27\% if the unapplied noise-floor rule is read strictly); density of sampling matters as much as shift type. & Q2 & F & E2 \\
F19 & After two measurements no threshold rule on the available diagnostics beats always continuing once escalation has a price, under table rotation or in sample; a perfect after-the-fact decision would save about half a measurement, and no diagnostic tested here captures that headroom. & Q4 & F & E3 \\
F20 & A measurement budget planned from the shift type alone, learned on other tables, transfers to a new table by the pre-registered criterion (its transfer regret interval includes zero) and saves about thirteen rows per domain against planning the full budget; refinements by density or by diagnostics do not transfer. & Q2 & F & E5 \\
F21 & Inside an adaptive loop no source family beats XGBoost and adaptation absorbs most family differences that zero-shot use shows; the random-split ranking of families does not predict their loop ranking. & Q1, Q4 & F & E4 \\
F22 & Of twenty closed-loop gains, four survive both negative controls and all four are the start-at-once finding; every $k$-means gain fails the shuffled-label control and the rest are fragile or positive controls. & M & F & E6 \\
F23 & On hard tail targets at a 3\% tolerance the loop beats chance by more than five measurements, and the answers to Q2--Q4 are unchanged: start at once, no acquisition or interval-aware gain, warm-start dominance, and no four-row diagnostic that beats always continuing. & Q2--Q4 & F & E7 \\
\end{longtable}
}

\paragraph{A practitioner's reading.} Table~\ref{tab:loopguide} restates the
findings as what a self-driving loop built on a learned surrogate would do at
each stage, with the finding that supports each line. The reading is
conditional on the framework's instantiation: tables as oracles, recorded
conditions as proposals, the policies and adaptation methods defined here.

\begin{table}[htbp]
\centering\footnotesize
\caption{What the findings imply at each stage of a loop built on a learned
surrogate, on tables like these.}
\label{tab:loopguide}
\begin{tabular}{@{}p{0.2\linewidth}p{0.62\linewidth}p{0.12\linewidth}@{}}
\toprule
Stage & What the evidence supports & findings \\
\midrule
Before any target measurement & Do not read the random-split error or the random-split ranking of model families as the deployment estimate; expect a shift-type-dependent factor, largest for a new device type or a new laboratory. No label-free screen certifies individual candidates; use it, with a size control, to choose sources and to find support gaps. Do not expect nominal interval coverage to transfer. Run a bounded model beside every unbounded one. & F1--F5, F9, F13 \\
Choice of source model & Once the loop adapts, no tested family beats XGBoost; the forests, the perceptron and the deep ensemble are not distinguishable from it and the Gaussian process is worse. Do not choose the family by its random-split error, and avoid a poorly converged Gaussian process as the source. & F13, F21 \\
First measurements & Start proposing at once and warm-start the source model on every measurement; do not spend a calibration budget first (under laboratory shift the reference loop did as well). Choose rows at random and propose greedily: no label-free acquisition rule or interval-aware proposal helps, even on hard targets. & F14, F15, F16, F22, F23 \\
Budget & Plan the budget from the shift type: a handful of rows under chemistry, composition or laboratory shift, while under device shift most tiers are not usable within twenty rows; a plan learned on other tables was not measurably worse than one learned on the new table. Expect many domains to be usable at the first row and some not within the budget. & F6, F18, F20 \\
After two or four rows & The bias ratio says how far the model will move, not whether to stop adapting, switch or escalate; keep adapting unless a full recalibration is nearly free. & F7, F19, F23 \\
Throughout & Keep measuring and updating: a frozen prior, however good, loses to a constant instrument scale error that an adaptive loop absorbs. & F17 \\
\bottomrule
\end{tabular}
\end{table}

\paragraph{What the framework still lacks.} After E0--E7 the open items are of three kinds.
A perfect after-the-fact decision on when to stop adapting would save more than a measurement on
hard targets, but none of the cheap diagnostics tested here captures that headroom;
a better diagnostic, or a decision taken later, is the first open question. Extrapolation beyond the
source table's label range occurs in three tiers only, so the loop's behaviour there is described, not
tested. Every answer is also bounded by the offline oracle: it returns only conditions a table
recorded, and the sealed evidence on laboratory shift rests on one laboratory (the 29 laboratory
tiers of the loop are literature sources of D1, source papers of R2-A2 and laboratory B). The prospective arm of the
framework, the same protocols on a laboratory's own instrument, is stated in
Section~\ref{sec:outlook}.

\section{Discussion}
\label{sec:discussion}

\paragraph{What the evidence supports.} The findings are collected with their
evidence tiers in Section~\ref{sec:synthesis} (Table~\ref{tab:findings}) and are
not restated here. This section discusses what they do not support, what
sealing adds, the deployment assumptions they test, the mechanisms behind the
two sealed failures and the limits of the programme.

\paragraph{What it does not support.} No material, formulation, device or
operating path is shown to be wet-lab feasible; no chemical compatibility,
manufacturability, stable dripping or prospective discovery is claimed.
Conformal coverage is not guaranteed for any new chemistry or device. The
negative \tox{} result is one sealed test of one published-feature baseline
on one dataset, not a general failure of ML. No applicability rule here is a
final or optimal pruning algorithm, and none is shown to work at the candidate
level within a source. The second sealed result is one frozen configuration
(laboratory-A data, the XGBoost configuration of V1) under one two-laboratory
shift that is confounded with buffer; it is not a general evaluation of the Smart Microfluidics
dataset, nor of its authors' models, and it says nothing about formulation
quality or clinical relevance. The candidate-level pass of the transductive
domain classifier under geometry shift (12 of 30 seeds) is a batch-screening
observation made with the test inputs in hand, not a guarantee for
candidates that arrive one at a time.

\paragraph{What sealing adds.} The sealed \tox{} error (19.84\%) is matched
by the retrospective leave-one-chemistry-out fold on the same group
(20.75\%), and the other five folds of that leave-one-chemistry-out cell lie between 5.22\% and
16.15\%; a
cross-validation would have shown the same magnitude. Sealing did not make
the failure larger or more surprising than that, and we do not present it as
such. Sealing adds three narrower properties: the model, endpoints and decision rule were
fixed before any label was seen, so the negative result cannot be the
residue of tuning on the held-out group; the estimate is prospective rather
than selected; and the chronology is verifiable. The agreement with the
retrospective fold is evidence that the sealed number is representative, not
that the seal was unnecessary. The same holds for D3, where the post-hoc
log-target refit shows how much of the frozen result was a modelling
choice, visible precisely because the choice was fixed before the outcome
was known. D3 also carries a protocol lesson for future sealed validations, stated
retrospectively: the development-set diagnostics that would have revealed the
right-skewed target (a grouped cross-validation reference of 39.41\% on the
raw scale against 16.89\% on the log scale) use development labels only and
leave the seal intact, so they should precede the lock. The D3
comparison thus separates a target-scale mis-specification visible inside the
development set (39.41\% falling to 16.89\% under the log-target refit, and
240.39\% to 43.05\% on the sealed rows) from the cross-laboratory shift that remains (16.89\% to 43.05\%, a factor
of 2.55).

\paragraph{What the sealed programme says together.} Together, the ten results R1--R10
test five assumptions that are commonly made when such surrogates are
deployed. Each is stated with the results that bear on it, and every
statement is confined to these data and these cells.

First, that accuracy on a random row-level split represents deployment
accuracy. Accuracy for microfluidic surrogates is typically reported on
random splits~\cite{lashkaripour2021,lashkaripour2024,chagot2022,samari2025},
and the limits of such estimates when rows share structure are documented in
other fields~\cite{kapoor2023,sheridan2013,wu2018}. Within these data the
shifts order themselves by severity: row-random and exact-input splits are
nearly interchangeable; geometry grouping and cross-literature transfer
multiply the reference error by 1.74--1.85; whole-chemistry holdout by 2.62;
device-type shift by 3.30 and 10.31; and the cross-laboratory liposome test
lies far above its own development reference (Table~\ref{tab:d3}) (R1--R3,
R7, R8). The frozen pipelines agree with one another where they overlap (the
\tox{} fold at 20.75\% versus the sealed 19.84\%; the chemistry macro at
10.24\% versus 10.27\%), so the shift effects are properties of the data
rather than of one implementation. R10 extends the gradient to three
further tables and falsifies a fixed multiplier band in all three: an unseen
fluid family or fluid pair costs 1.32 and 1.86 times the development
reference, an unseen device 5.03 and 4.95 times, and in the literature-compiled
LNP table, whose development reference is itself a cross-paper grouped error,
unseen lipids, papers and preparation methods cost no more than that
reference (0.80, 0.86 and 0.62). The multiplier depends on which
heterogeneity the development set already contains, and the retrospective
leave-one-group-out reference (0.77, 0.99 and 0.75 of dev-ref) understates
the sealed device tiers in both droplet tables.

Second, that a label-free applicability screen can pick out reliable
individual candidates under shift. Applicability-domain screening is commonly
recommended to flag unreliable predictions~\cite{tropsha2003,netzeva2005,sahigara2012},
and selective prediction formalises the trade between coverage and
risk~\cite{elyaniv2010}. Here failures come in two layers that need different
detectors: a support gap, which marginal screens see from inputs alone (D3:
PEG and HSPC in-range fractions 0.316 and 0.684), and a conditional shift,
which they cannot see (\tox{}: every row in range; D3: median error 0.884
even inside the marginal range). Only the first layer is addressable
label-free; the second requires either the unrecorded state variables or
labels from the target domain. Where a screen does help, the gain operates at
the level of literature sources rather than candidates, and the right
negative control depends on the shift type: a size proxy outranks every
applicability score under cross-literature and cross-laboratory shift, yet is
worse than random under single-to-double-emulsion shift, so both controls must be reported
whenever an applicability benefit is claimed. Because relative error divides
the residual by observed size, in our data any score correlated with length
scale lowers retained MAPE mechanically by discarding small-diameter
candidates; the size-only control exists to expose this, and without it the
denominator effect could be mistaken for out-of-distribution detection (R4,
R7, R8, R9). In the constructive retrieval experiments on the revealed R10
tiers a kNN applicability filter before retrieval did not systematically
raise the success rate, and on the R2-A1 device tier it rejected all 109
rows, so the second layer again required target labels.

Third, that physics guidance, hard structural constraints and native
variables resolve extrapolation to an unseen chemistry. Physics-guided
learning is widely expected to extrapolate better than purely data-driven
models~\cite{karniadakis2021,willard2022}, and hard constraints are advocated
over soft penalties for that purpose~\cite{naturecs2021,lu2021,beucler2021}.
In these data the frozen V2 did not reduce chemistry-macro error relative to
the published baseline (R6). What the bounds did is remove physically
impossible outputs: every unbounded model of the V2 matrix produced
physically invalid points under chemistry shift (B0 1.29\%, B1 3.22\%, B2
2.58\%, B3 13.73\%), as did the fixed-architecture perceptron of the method
matrix and the frozen liposome surrogate, while the bounded V2 variants
produced none. The post-hoc log-target refit of the liposome model
shows that positivity alone is conferred by a target transform with no
architecture-level bound; what the $(0,D)$ parameterisation adds is the
regime's upper limit and the anchor, and neither form of bound brings
chemistry-macro error below the published baseline here (R6, R8, R9), although
the bounded V2-H improves on its unbounded twin B3 (10.59\% against 16.64\%). R10 pre-registered the positivity part as \mbox{H-B1}: the
log-target twin of the perceptron holds on both new tables (on R2-A1 after the
post-reveal re-run that corrected the bounded-perceptron contrast; 97.5\% versus
2286\% with 103 of 109 negative diameters on the R2-A1 device tier; 31.8\%
versus 36.7\% on the LNP lipid tier) and fails on A1, where it extrapolates
to $1.66\times10^{6}$\% against 1134\% for the raw-scale model, so a target
transform is neither a safe nor a harmful default. Native versus dimensionless
inputs (\mbox{H-C1}) stayed undetermined ($+0.28$~pp, $p=0.45$, on A1; two pairs on
R2-A1).

Fourth, that model families with native uncertainty, or conformal prediction
reweighted for covariate shift, restore coverage under shift. Deep ensembles
and related families are the standard proposal for predictive uncertainty
under shift~\cite{lakshminarayanan2017,ovadia2019}, and weighted conformal
prediction extends coverage guarantees beyond exchangeability when the
density ratio is known or estimable~\cite{tibshirani2019,barber2023}. In
these cells every interval method loses coverage under shift; what differs
is the price of recovery (grouped calibration at two- to fourfold width) and
the robustness ranking (quantile-regression intervals above symmetric
residual intervals under chemistry shift). The three uncertainty-aware
families confirm that this is a property of the shift rather than of
XGBoost; the only interval that survives a device-type cell does so by
growing to the size of the quantity it predicts. Reweighting calibration
residuals by estimated input density ratios does not change this: where the
shift is large the weights diverge and every interval becomes infinite, and
where it is mild the domain-classifier weighting reaches nominal coverage at
the cost of one interval in five being infinite (R3, R7, R8). R10 adds that interval-contained retrieval answers no query in
all six sealed tiers ($A=W=0$), and
that the quantile forest and the deep ensemble widen under shift while
tracking row error on A1 and R2-A1 but not on the LNP table; the ensemble's
width grows about $1.10\times10^{5}$-fold on the A1 device tier and 169-fold
on the R2-A1 device tier, which is coverage by abstention.

Fifth, that retrospective cross-validation without pre-registration or
sealing suffices to assess generalisation. Cross-validation is the default
evidence of generalisation in applied machine learning, and its exposure to
leakage~\cite{kapoor2023} and to analyst degrees of freedom is why
pre-registration has been proposed~\cite{hofman2023,vepa2026}. Here the sealed \tox{} number
is matched by the retrospective fold, so sealing did not change the
magnitude; what it rules out is tuning on the held-out group, and what the D3
case adds is that a modelling choice fixed before the outcome, the raw-scale
target, could be isolated afterwards precisely because it was frozen: the
post-hoc log-target refit removes the negative predictions but leaves a
laboratory shift of about two and a half times the development reference
(R5, R9; see ``What sealing adds'' above). R10 makes the ranking point
quantitative: six-model rankings under random splits failed to predict the
sealed rankings in all three further tables (Spearman $\rho$ of
$-0.77$ / $-0.14$, $0.37$ / $-0.71$ and $-0.54$ / $0.20$ on S2/S3), the
pre-registered success condition of H-E.

\paragraph{Conditional shift and unobserved state.} The \tox{} failure
occurred with every row inside the marginal training ranges. The public table
omits dynamic interfacial tension, adsorption kinetics, molecular descriptors,
interfacial rheology and wetting behaviour, state that the source authors
themselves flag as relevant~\cite{chagot2022} and that the droplet-microfluidics
literature treats as governing surfactant-laden break-up~\cite{anna2016}. More
model capacity or more
physics on the four recorded predictors cannot reconstruct unobserved physics;
the V2 verdict is consistent with this reading.

\paragraph{The role of hard constraints.} We do not argue that hard
constraints beat soft penalties; that comparison has been examined
elsewhere~\cite{naturecs2021,baty2023,prnet2026}. The evaluation's quantitative
product is narrower: physics guidance plus architecture-level bounds did not
improve accuracy relative to the published baseline under chemistry shift here, but bounds did guarantee physically
admissible outputs where the unconstrained B3 produced 13.73\% invalid points.
Within the V2 matrix the descriptive progression of Section~\ref{sec:r6}
(B3 16.64\% $\to$ V2-H 10.59\% $\to$ V2-P 12.05\% chemistry-macro MAPE;
invalid points 13.73\% $\to$ 0 $\to$ 0) is reported without attributing the
steps to the bound or to the anchor; that attribution requires a
pre-registered test.
Two further observations point the same way without changing the verdict: in
the method matrix an unbounded multilayer perceptron with fixed
hyperparameters extrapolates to 266.34\% and 3592.49\% error under
device-type shift where the tree models stay at 73.67\% or below, and on the
liposome dataset the unbounded frozen surrogate returned negative diameters
for 38.6\% of 57 sealed rows. Neither shows that bounds improve accuracy; both
show what their absence permits. The liposome case also bounds the claim
from the other side: refitting the same configuration to the logarithm of
size removes every negative prediction (Section~\ref{sec:r9}), so the
cheapest insurance against a sign error is a positivity-preserving target
transform, and the case for architecture-level bounds rests on the upper
limit and on regime-specific constraints rather than on positivity. A last
connection is speculative and we offer it as such: a hard
structural bound fixed before fitting also acts as a guard against post-reveal
tuning impulses, because it cannot be relaxed to fit a revealed group without
breaking the frozen protocol. We found no explicit treatment of this link in
the literature we searched.

\paragraph{Methodological role of incident disclosure.} The serialisation
failure at the reveal boundary is an engineering fact, not a scientific
finding, and we present it as a reproducibility case report: what was frozen,
what failed, what the recovery touched and what it did not. Its value is in
making the chronology verifiable, not in being novel.

\paragraph{Limitations.} The external set has 64 rows from four sources with
per-source $n$ as small as 6; $R^2$ is unstable there. Retention is
source-blocked, so candidate-level applicability benefit is untested.
Only six chemistry groups exist for V2, so group-level conformal calibration
is not finite-sample attainable and all V2 intervals are heuristics. The
\tox{} reveal gate is consumed; any new blind claim, for V1 or V2, requires a
different, never-inspected dataset. The DAFD normalised workbook lacks
row-level viscosities, interfacial tensions and densities, so V2 cannot be
transferred to DAFD as a fully native model. The matrix models are untuned,
so the multilayer-perceptron figures describe one fixed configuration rather
than the architecture's limit; the matrix reference cells use the fit
partition and are not comparable with Table~\ref{tab:repro}; the SE$\to$DE
cell has two test groups. The V2 anchor map clips before normalising and
falls back discontinuously to the midpoint for $y_{\mathrm{phys}}\ge 1$
(Section~\ref{sec:v2protocol}); neither path is exercised on the 466-row
table, and both are recorded as limitations of the frozen code. The
uncertainty-aware families are likewise
untuned: the deep ensemble has no likelihood head, so its under-dispersion
does not bound what a proper deep ensemble could do, and the Gaussian
process's length scales reach their fixed bounds on several features. The
liposome development reference is a grouped
cross-validation MAPE on laboratory A rather than a published held-out error,
because none was retrieved before the lock; the sealed laboratory is
confounded with buffer; the 12-row operator set supports no inference; and
both D3 label sets are now open, so no further blind claim can be made on
them. The V1 development reference and the D3 log-target baseline are
retrospective; neither is a validation result, and the latter was chosen
after the reveal. The three R10 tables add their own limits: A1 and R2-A1
confound the device tier with fluids absent from development, so the device
multipliers are compound shifts; after the sealing rule R2-A1 leaves two
fluid pairs for development, so its dimensionless-versus-native test rests on
two pairs; the LNP Atlas is a literature compilation of reported sizes with
heterogeneous measurement protocols rather than a laboratory's raw
measurements, and its development reference already spans 33 source papers,
which is why no sealed tier exceeds it; and the bounded-perceptron contrast on
R2-A1 was recomputed after the reveal because of an implementation defect,
with every other sealed prediction byte-identical to the pre-reveal bundle.
The offline closed loop has limits of its own: the oracle is row-indexed, so
the loop can realise only conditions a table recorded and never pays for a
proposal the table cannot answer; targets are recorded labels, so every target
is reachable; at $\tau = 0.10$ the point task is easy for most tiers and the
policy differences are fractions of a measurement, which is why E1b halves the
tolerance and the budget and E7 moves to tail targets at a 3\% tolerance; the
E4 families are untuned and the Gaussian process often stopped at its
optimiser's limits; the device stratum is dominated by D1 geometry tiers
whose small development reference makes $\varepsilon$ strict; 25 of the 29
laboratory tiers are source papers of one literature compilation (R2-A2); the
E2 shares do not apply the protocol's noise-floor rule, which rests on three
and four replicate groups where it binds; the budget transfer of E5 rests on
five tables and an interval that includes zero, with 14\% of tiers
under-planned; escalation is modelled as the ceiling model, which favours
escalation; the E3 and E7 rule families are single splits on one diagnostic;
and the E4 runs differ between repetitions only in their wall-clock fit times.
No wet experiment was performed.

\section{Outlook: The Prospective Arm of the Framework}
\label{sec:outlook}
The offline instantiation says what a loop built on a learned surrogate does on
recorded tables; the prospective arm runs the same protocols on an instrument.
The framework's code reads any table with the row schema of
Section~\ref{sec:data}: a laboratory's own development table is $S$, its new
device, fluid or chemistry is $T$, the instrument replaces the row-indexed
oracle, and $\tau$, $B$ and $\varepsilon$ are pre-registered before the first
measurement. The E0 checks then use the laboratory's own replicate rows for the
noise floor and a deliberate scale error of the instrument for the drift check, and
the E1 policies become the loop's actual acquisition, adaptation and proposal
rules; the quantity that the offline replay could not produce, the cost of a
proposal that the instrument cannot realise, is recorded as a failed proposal.
The programme below is what we would run next in that arm, stated as
hypotheses and designs rather than results. Nothing in the prospective arm has
been executed.

\paragraph{Record the state that the failures point to.} Both sealed failures
occurred where the recorded inputs omit variables that plausibly govern the
outcome: for \tox{}, dynamic interfacial tension, adsorption kinetics and
wetting; for the liposome dataset, buffer ionic strength and
equipment-specific mixing. A schema for future collections should carry, per
row, the surfactant or lipid identity with molecular descriptors,
time-resolved interfacial tension where measurable, viscosity curves for
non-Newtonian phases, contact angle or a wettability class, temperature,
buffer composition, chip material and fabrication batch, equipment and
operator identifiers, and a flow-regime label (dripping, squeezing, jetting).
The unified long-format row schema in our repository already reserves fields for device class,
experiment cluster and shift-group keys so that sealed splits can be drawn
mechanically; the state variables listed here are the additions it needs.

\paragraph{Fill support gaps by design, not by chance.} The liposome failure
was predictable from inputs alone: two formulation components had support
fractions of 0.316 and 0.684, and the buffer was confounded with the
laboratory. A prospective collection should place design points along the
boundary of the existing data's applicability domain rather than inside
it (space-filling designs over the full component ranges, both buffers in
both laboratories, both device types in one study), so that calibration
partitions contain the regions a deployment will visit. The applicability
scores evaluated here could not certify predictions, but they remain
candidates for choosing where to measure next.

\paragraph{Measure the laboratory effect and learn how many target rows
recalibration needs.} A multi-laboratory round robin with the same
formulations on the same chip model, each laboratory's rows sealed in turn
under the protocol used here, would separate the equipment, operator and
buffer effects that D3 could not disentangle. On top of it we would test
few-shot recalibration: freeze the source model, reveal $n$ target-laboratory
rows chosen at random or by an applicability score, adapt by warm-start
residual boosting, by a bias-and-scale (log-affine) correction or by a
hierarchical model with laboratory and device random effects, and evaluate on
the sealed remainder; the offline results (Sections~\ref{sec:q3} and
\ref{sec:q4}) favour random rows and warm-start, which this test would check
on an instrument. The curve of sealed error as a function of $n$ is
the quantity a practitioner moving to new equipment needs. The operator-B set
hints that operator shift is milder than laboratory shift, but with twelve
rows it cannot say by how much.

\paragraph{Treat scale-up as regime change.} Larger channels and higher flow
rates move a device across capillary-, Reynolds- and Weber-number regimes and
across the confinement ratio $d/D$. The bound $0<\hat d<D$ that held for one
oval junction fails for plugs and jets, and dimensionless features, the
natural transfer prior, fell from $R^2$ 0.80--0.83 to 0.48 on a new channel
depth in a droplet study that tested them~\cite{labchip2025dimless}, and did
not prevent the 3.30 and 10.31 error ratios under device-type shift here. We would therefore (i)~train a
regime classifier from recorded inputs and fit regime-specific anchors and
bounds; (ii)~use dimensionless groups as a prior to be tested by a sealed
holdout on the larger device, not as a guarantee; (iii)~make a hard bound
appropriate to the regime a mandatory baseline for every model, since the
unbounded models here produced negative diameters and errors above 200\%;
and (iv)~check the input schema for identifiability before training,
following~\cite{cieslak2026}, so that the model is not asked for a quantity
its inputs cannot determine.

\paragraph{Uncertainty that survives shift.} Weighted conformal prediction
with estimated density ratios~\cite{tibshirani2019} is the natural candidate
under covariate shift. Tested here with the transductive estimates already in
hand (Section~\ref{sec:r8}), it abstains on every row under device-type and
cross-literature shift and repairs geometry shift only with one interval in
five infinite, so the next step is better-calibrated density-ratio estimators
together with an explicit abstention rule, not more reweighting.
Group-conditional (Mondrian) calibration needs roughly ten independent
calibration groups before a 90\% group-level target becomes attainable, and
collections should be planned with that count in mind. Quantile-regression
intervals are the sensible default when groups are few, given their behaviour
under chemistry shift here.

\paragraph{An evaluation standard.} For each new dataset: a protocol hashed
before download; a sealed whole-group holdout (laboratory, device type or
chemistry) revealed once; a reference error computed on development data
before reveal; both a random and a size-proxy control for every
applicability claim; batch-level and candidate-by-candidate deployment
reported separately; a bounded baseline beside every unbounded model; two
independent runs with byte-identical manifests; and publication of the
result whatever its sign. The method matrix of Section~\ref{sec:r8} is the
template, with the cross-laboratory row to be filled by the round robin above
and a real-versus-simulation row from public CFD data whose licence is still
to be verified.

\section{Reproducibility}
\label{sec:repro}
Data provenance is fixed by SHA-256 digests of the four OSF workbooks, the
two Zenodo MATLAB tables and the Zenodo 17867478 archive
(Appendix~\ref{app:hashes}); the Zenodo MD5 values match their records, and
the six bound raw files of the liposome dataset are hashed individually in
the custodian manifest. The V1 protocol (\hashtrunc{14027524}{e68b1c}),
pre-registration lock (\hashtrunc{1f6901bc}{57edc7}), prediction bundle,
consumed gate (\hashtrunc{e285c4cb}{b13344}), recovery implementation
(\hashtrunc{06b6099e}{263a74}) and incident record
(\hashtrunc{75615dd3}{7db162}) are hashed and cross-referenced. The V2
protocol digest is recorded in each run's configuration before fitting; the
V2 implementation (\hashtrunc{21ab9ea2}{3b08c7}) and its tests are hashed in
the same file. Both the V2 and the expanded DAFD suites were run twice in
separate directories; the two \texttt{artifact\_hashes.json} manifests (24 and
45 entries) are byte-identical, and every listed file in both runs
re-hashes to its manifest entry. Per-row predictions for every model, fold and
seed, split manifests proving zero prohibited overlap, inner-selection
records, applicability calibration metadata and label-free selection files
hashed before scoring are retained. All released JSON passes a strict parser
that rejects \texttt{NaN} and infinities. Superseded fast/smoke outputs and
pre-fix JSON are excluded from the release inventory. The test-evidence
placeholder in each run states explicitly that the analysis script did not
execute the unit tests; test execution is recorded separately by the release
process. All numbers in this manuscript are checked programmatically against
the artifact files by a ledger-driven script, including a prefix-and-suffix
match for every truncated digest and a denominator check for the pruning
percentages.

The method-matrix protocol (\hashtrunc{8f058306}{6cb3f8}) was hashed and
committed before any run; its two canonical runs in separate directories
produced byte-identical 13-entry manifests, and every listed file re-hashes
to its entry. The uncertainty-aware comparison protocol was likewise
committed before its two runs (digest \hashtrunc{23dc7d0a}{951aa8}); the
two 11-entry manifests are byte-identical, and the 24 register entries are Gaussian-process optimiser convergence clarifications (W4-C1) plus the same DE$\to$SE calibration fallback as the matrix (W4-D1). The covariate-shift-weighting protocol
was committed before its two runs (digest \hashtrunc{03af5544}{e60b40}); the
7-entry manifests are byte-identical, and the unweighted branch reproduces the
matrix's split-conformal coverage in every cell. The second sealed validation is bound by
a protocol hashed before download (\hashtrunc{1fa4b404}{c91419}), a parameter binding hashed
after inspecting headers and metadata only (\hashtrunc{c943ce8b}{7d284f}), a
custodian manifest, a numbered deviation register, frozen prediction and
reveal programs with recorded digests, a prediction bundle containing inputs,
predictions, interval endpoints, applicability scores and masks and no labels
(\hashtrunc{79be728c}{296bd2}) and a one-time gate
file; each step is a separate commit, so the order cannot be changed without
rewriting history. The reveal program re-verified every digest before reading
a label; the single post-reveal correction lives in a separate file and
touched no frozen artifact. The five post-hoc analysis blocks of
Section~\ref{sec:posthoc} write to a separate directory with their own hash
manifest and environment record, read only revealed or never-sealed labels,
and are marked post hoc wherever their numbers appear.

\paragraph{Data and code availability.} The three Stage~1 datasets are public under
their own licences (OSF 938rs; Zenodo 7055018, CC~BY~4.0; Zenodo 17867478,
CC~BY~4.0) and are not redistributed. The protocols, binding file, deviation
registers, ledgers, reconciliation script, matrix runner, custodian,
prediction and reveal programs and byte-copies of the summary artifacts are
released in the public repository \url{https://github.com/LibertychaserUS/microfluidic-native-physics} at tag
\texttt{v0.5.5-preprint}, with every digest and its file listed in
\path{hashes/appendix_d_registry.json}; the full per-row artifacts
of the canonical runs (about 26~MB per matrix run) and the custodian's
sealed-label files are held in our workspace. The sealed-label files
will not be released, because they are rows of the public dataset that any
reader can reconstruct from the binding file. With the release a reader can
reproduce the custodians' splits and sealed-label digests, the hashed
pre-reveal bundles and every printed number; the order of events asserted in
this paper (protocol before download, binding before fit, one consumed gate)
rests in addition on the commit history of our working repository, which the
release does not contain, and we state this rather than imply otherwise.
Independent verification requires re-downloading the public datasets, re-running the canonical runs in
the pinned environment, comparing manifests, and executing the ledger script,
which re-derives every printed number from the artifact files; the per-loop
tables behind the closed-loop intervals (\texttt{loops.csv}, \texttt{budget.csv})
stay in the working directory and can be regenerated by the frozen runners,
whose two counted runs produced byte-identical tables.

\paragraph{R10 artifacts.} The frozen protocols (\texttt{H\_PROTOCOL\_v1} and
\texttt{v2}), the per-dataset bindings, the custodian, runner, reveal and
combination programs with their SHA-256 sidecars, the deviation register and
byte copies of every manifest, run summary and reveal result are under
\path{scaleup/validation_plan}, \path{scaleup/w6} and the four result
directories \path{scaleup/h_A1}, \path{h_R2A1}, \path{h_R2A2} and
\path{h_combined}; the constructive experiments are under
\path{scaleup/h_portfolio}, \path{posthoc_round3} and \path{posthoc_round4}.
The A1 ESI (CC~BY~3.0), the Wang et al.\ deposit (CC~BY~4.0) and the LNP
Atlas (CC~BY~4.0) are not redistributed, and no per-row third-party value
appears in the repository. The reproduction-fidelity protocol, scripts and both result sets are under \path{scaleup/fidelity}; the few-shot meta-analysis protocol is under \path{scaleup/validation_plan} and its results under \path{scaleup/h_fewshot_rule}; the pooled pre-training feasibility gate is \path{scaleup/validation_plan/H_I_POOL_FEASIBILITY_GATE.md}.

\paragraph{Closed-loop artifacts.} The seven protocols, their sidecar digests
and the deviation register are under \path{scaleup/validation_plan}; the frozen
runner and analyzer, the E1b, E3, E4, E6 and E7 wrappers and the E5 analyzer under
\path{scaleup/w7}; and byte copies of every run manifest, environment record,
hash manifest and analysis result of the two runs of E0, E1/E2, E1b, E3, E4, E5,
E6 and E7, the dense-tier sensitivity, the post-hoc permutation, tie and
noise-floor checks, the plotted figure values, the report and the ledger
entries under \path{scaleup/e_closed_loop}. The per-loop tables (\texttt{loops.csv},
\texttt{budget.csv}) are held in our workspace; no per-row third-party value
enters the repository. Each pair of runs has byte-identical hash manifests
before and after analysis, except that the two E4 runs differ in their recorded
wall-clock fit times (the fit-time columns of \texttt{families.csv} and
\texttt{table\_cv.csv} and the \texttt{fit\_seconds\_mean} block of
\texttt{results.json}); an identity check stored with the runs confirms that the
loop, family and cross-validation tables agree in every other value, and the two
results files differ only in that block.

\paragraph{Acknowledgement of sources.} This independent work re-implements,
stress-tests or builds on public work by Lashkaripour et al.\ and Chagot et
al.; it does not imply their endorsement.

\bibliographystyle{unsrtnat}
\bibliography{references}

\appendix

\section{Search basis for the two limited claims}
\label{app:search}
Both ``to our knowledge'' statements in Section~\ref{sec:intro} rest on two
structured web searches (16~October~2026) whose full records are kept in the
project repository. We summarise scope, findings and
limits here; this appendix is the documented basis required by our own release
checklist and is not a systematic review.

\paragraph{Scope.} Five axes were searched for the design side: ML/DL for
microfluidic device and inverse design; physics-guided or physics-informed
microfluidic modelling, including hard constraints; the phrase ``native
variables'' and its contrast with dimensionless learning; reliability evaluation, preregistration and sealed prediction in ML for science; and ML--microfluidics
reviews and benchmarks. Five further questions were searched for the
methodology side: negative-result publication venues; preregistration and
blind holdout in ML; hard-versus-soft constraint comparisons; use of the
phrase ``reliability audit''; and disclosure of engineering incidents in
scientific ML.

\paragraph{Findings that forbid absolute claims.} ML inverse design of
microfluidic devices is saturated across droplet generators, DLD, micromixers,
chip architecture and circuits~\cite{lashkaripour2024,vatandoust2022,
samari2025,li2026pathchip,mufg2026,ceyehao2026}. Physics-guided microfluidic
modelling is occupied, predominantly with soft penalties or
PINNs~\cite{bpgnn2026,flexpinn2026}. Publishing negative results, preregistering
predictive modelling, and comparing hard with soft constraints each have
prior work~\cite{karl2024,hofman2023,naturecs2021,baty2023,prnet2026}.
``Reliability audit'' is a descriptive phrase in use (for example
in~\cite{schulam2019}) and is not claimed as a term. Materials
informatics already has pre-registered negative validations~\cite{vepa2026}.

\paragraph{Findings that permit the two qualified claims.} No retrieved
peer-reviewed work implements a sealed-prediction or preregistered reliability validation on a microfluidic ML design surrogate; microfluidic reviews name the gap
without filling it~\cite{labchip2025review,das2025}. No retrieved work unifies
physics guidance, architecture-level hard constraints and inverse-design
evaluation for microfluidics; the nearest are a periodicity-enforcing network
for DLD~\cite{lee2025} and an optimisation-layer repair
operator~\cite{micromixer2026}, and one physics-guided micromixer paper lists
the combination as an open gap~\cite{bpgnn2026}.

\paragraph{Items not retrieved.} (i)~Any peer-reviewed paper disclosing a
self-inflicted reveal or serialisation incident in a sealed holdout as a
methodological section; (ii)~``native variables'' as an established modelling
term; (iii)~a microfluidic ML paper titled or framed as a ``reliability audit'';
(iv)~a benchmark review of reliability for microfluidic ML design surrogates.
Non-retrieval is not evidence of absence. (v)~One item was resolved on
17~October~2026: the full text of the dimensional-analysis droplet
paper~\cite{labchip2025dimless}, unavailable during the scan, was read from
its PubMed Central deposit, and Section~\ref{sec:related} now follows its
reported $R^2$ values.

\paragraph{Limits.} Web search is not exhaustive; several 2026 hits on
``sealed holdout'', ``preregistration'' and ``negative result'' come from
non-peer-reviewed self-archives and are neither cited as prior art nor relied
upon. Because they occupy the vocabulary, we avoid any unqualified ``first''.
Bibliographic metadata (author lists, titles, volumes, article numbers and
DOIs) were checked against Crossref, arXiv and Zenodo records on
17~October~2026, and again for every entry on 24~October~2026.

\section{Implementation details and code organisation}
\label{app:impl}
This appendix records the configuration constants and program structure
behind Sections~\ref{sec:w3protocol} and \ref{sec:d3protocol}, so that the
runs can be repeated without reading the code. Values here are design
constants fixed in the frozen protocols, not measured results.

\paragraph{Software environment.} Python~3.12.13, numpy~2.5.1,
pandas~3.0.3, scikit-learn~1.9.0 and XGBoost~3.3.0 on macOS (arm64), with
single-threaded numerical libraries and \texttt{n\_jobs=1} everywhere; each
run writes its own environment record. The earlier suites (the R1 baseline, expanded
DAFD, V1, V2) record their versions in their own reproducibility manifests.

\paragraph{Method matrix: partitions and models.} Within each cell the
training rows are split into fit and calibration partitions by
\texttt{GroupShuffleSplit} (test size 0.20, random state equal to the seed)
on dependency groups for D1 and by row for D2; zero group overlap is
asserted. Point models are XGBoost in the published D1 configuration (100
trees, depth 6, learning rate 0.3, L2 regularisation 1, minimum child weight
1) and in the depth-3 B1 configuration for D2; a random forest with 400 trees
and minimum leaf size 2; and a multilayer perceptron on standardised inputs
with two hidden layers of 64 units, ReLU activations, Adam at learning rate
$10^{-3}$, early stopping on a 15\% internal validation split with 30
iterations of patience and at most 2000 iterations, with the D2 response
standardised. No model is tuned per cell. TabPFN was excluded because its
weights require a network download that the runs do not perform.

\paragraph{Method matrix: intervals and applicability.} All interval
calibration happens in model-response space on the calibration partition and
is mapped to micrometres through the dataset's monotone positive map. Split
conformal uses the finite-sample quantile of absolute residuals. CQR fits
gradient-boosted quantile regressors at 0.05 and 0.95 (300 trees, depth 3,
learning rate 0.03, minimum leaf 5), the same hyperparameters as the expanded
suite. Mondrian conformal aggregates absolute-residual scores by group
maximum over the calibration partition's test-group key and takes the
finite-sample order statistic, reporting the attainable coverage ceiling when
the number of groups is too small. The ensemble-variance interval refits the
point model on 20 cluster-bootstrap resamples, takes the per-row standard
deviation, and sets the radius to $q\cdot\max(\mathrm{sd},\mathrm{floor})$, the floor being one
tenth of the tenth percentile of the positive calibration standard deviations,
with $q$ the finite-sample 0.90 quantile of $|\mathrm{residual}|/\mathrm{sd}$
on the calibration partition. Applicability scores are fitted on
fit-partition inputs with thresholds taken from calibration-partition scores:
standardised 5-NN distance; OAS-shrinkage Mahalanobis distance (the expanded
suite's implementation); Isolation Forest with 400 trees and automatic
contamination; a class-balanced logistic regression separating fit inputs
from test inputs (transductive, recorded as such); and the 5-NN density ratio
of Section~\ref{sec:w3protocol}. Rule (a) thresholds at the 0.95 calibration
quantile; rule (b) keeps the lower-scoring half of every test group. Point
endpoints are MAPE, MAE in micrometres and $R^2$ (flagged unstable for test
groups with fewer than 15 rows), with an equal-weight macro over test groups;
interval endpoints are empirical coverage, mean width, simultaneous group
coverage and the attainability flag; applicability endpoints are Spearman
correlation with relative error, the tie-aware AURC, within-group AURC over
groups with at least four rows, and precision, recall and retention of both
rules at 10\% relative and 10~\um{} absolute error. The candidate-level test
permutes scores within each test group (2000 permutations, one-sided) and
applies BH across the five scores of a cell.

\paragraph{Method matrix: code.} The runner is a small package whose modules
mirror the protocol: a workspace loader that reuses the evaluated suite's
frozen data loaders and core functions; dataset specifications with units and
size proxies; cell construction with the fit/calibration split and overlap
assertions; point fits that never see rows outside the supplied fit index;
interval and applicability modules in which no function accepts a target
column; an evaluation module that is the only place test labels enter; and an
orchestrator that writes CSV with twelve-significant-digit floats, JSON
through a strict writer that rejects non-finite values, a hash manifest of
every artifact and a \texttt{compare} command for the two-run check. Seven
unit tests pin the invariants: the AURC equals the expanded suite's
risk-curve scalar under ties; the within-group median rule keeps half of
every group; BH adjustment is monotone and bounded; grouped cells have zero
unit overlap and the expected sizes; fit and calibration partitions are
disjoint and grouped; scores and intervals are label-free and deterministic;
D2 models predict in micrometres. A canonical run takes about 25 minutes on
one core; fast runs with two seeds and 200 permutations are labelled
non-inferential and never summarised.

\paragraph{Second sealed validation: programs.} Four programs implement the
D3 protocol. The custodian is the only program that reads laboratory-B
targets before the reveal; it assigns each row an opaque identifier (a UUID
derived from the source-file digest and row index), quarantines rows by index
without reading targets, writes the labelled development table, the
input-only sealed tables and the sealed label files (mode 0600), and prints
nothing but counts and digests. The prediction program reads development
labels and sealed inputs only, never opens a label file, fits the locked
XGBoost (100 trees, depth 3, learning rate 0.3, seed 42) on six numeric
inputs plus one-hot aqueous medium over the development levels, fits the
secondary PDI regressor and formation classifier in the same configuration,
calibrates the 90\% split-conformal radius on a 20\% \texttt{GroupShuffleSplit}
partition of the development clusters, computes applicability scores with
thresholds at the 0.95 calibration quantile, and writes a hashed bundle
containing inputs, predictions, interval endpoints, scores and masks together
with an environment record and the digests of the protocol, binding, register
and program files. The reveal program verifies every one of those digests,
creates the gate file with an exclusive-create flag so that a second reveal
is refused, joins labels to the bundle by opaque identifier, and scores every
pre-declared endpoint (matched-budget random retention with 10\,000
repetitions; within-stratum permutations with 2000 draws over strata of at
least four rows), writing through the strict JSON writer regardless of the
result's sign. The correction program for register entry D3-R1 recomputes
only the applicability block from the already-revealed scored rows into a
separate file. Before the real run, the prediction and reveal programs were
exercised on a pseudo-sealed split of laboratory-A rows, and a second reveal
attempt was refused by the gate.

\paragraph{Uncertainty-aware comparison models.} The W4 runner reuses the
matrix's cells, partitions, loaders, applicability scores and endpoint
functions and adds one module with the three model families. The quantile
forest is a scikit-learn random forest (400 trees, minimum leaf size 5,
seeded) whose leaf assignments on the fit rows give each query row a weight
vector (one over the leaf size for every tree, averaged over trees, checked
to sum to one), from which weighted empirical quantiles at 0.05, 0.95 are
read; the implementation is tested against a brute-force inverse CDF. The
Gaussian process uses a constant-times-anisotropic-RBF kernel plus a white
noise term with bounds $[10^{-3},10^{3}]$, $[10^{-2},10^{2}]$ and
$[10^{-6},10]$, target normalisation, two optimiser restarts and the cell seed
as random state; optimiser convergence warnings are counted and logged as register clarifications (W4-C1), never suppressed. The deep ensemble fits five copies of the
matrix's 64--64 perceptron with random states $1000\,s+m$ for seed $s$ and
member $m$ on the same fit rows and reports the member mean and standard
deviation. The native uncertainty, the two conformal corrections and all
scores are computed in response space and mapped to micrometres as in the
matrix. Six unit tests pin the weighted-quantile arithmetic, row-stochastic
weights, interval ordering and determinism of all three models.

\paragraph{Covariate-shift-weighted intervals.} The W5 runner adds one pure
module: a weighted radius that sorts calibration residuals, accumulates the
normalised weights $w_i/(\sum_j w_j + w(x))$ and returns the first residual at
which the cumulative mass reaches 0.90, or $+\infty$ if it never does; the
domain-classifier odds with probabilities clipped to
$[10^{-6},1-10^{-6}]$; the $k$-NN density ratio
$(r_{\mathrm{fit}}/r_{\mathrm{test}})^{d}\,n_{\mathrm{fit}}/n_{\mathrm{test}}$;
and the effective sample size $(\sum w)^2/\sum w^2$. Seven unit tests check
that constant weights reproduce the matrix's clamped finite-sample quantile
whenever that quantile is attainable and return an infinite radius otherwise,
that the radius is monotone in the test weight and in up-weighting of large
residuals, and the scaling of the two weight estimators. Infinite radii are
written as \texttt{inf} in CSV and never enter a mean width.

\paragraph{Post-hoc analyses.} A single script implements the four analyses
of Section~\ref{sec:posthoc}. It refuses a non-empty output directory, reads
the V1 development table, the D3 development table, the D3 sealed inputs and
revealed scored rows, and the matrix's canonical per-row predictions; it
re-derives the frozen D3 development reference and the frozen raw-scale
predictions before fitting the log-target model, and aborts if either
differs; bootstrap resampling uses a fixed seed; outputs pass the strict JSON
writer and are listed in a hash manifest with an environment record. A second
script implements the mechanism checks (v): it imports the frozen V2 module
and evaluates its anchor-to-logit function on the anchor values of the
locked sensitivity table, asserting that the recomputed validity mask equals
the frozen one; reads the matrix's per-row predictions and the D1 table to
express predictions in model-response space; reads the expanded suite's
scored rows and permutation controls; and reads the V2 and uncertainty-aware
per-row prediction files for interval widths. Its outputs are hashed with an
environment record that lists the digests of every input.

\paragraph{Number reconciliation.} A standard-library script reads a ledger
in which every printed number names the artifact file and the JSON path or
CSV selector it is recomputed from. It fails if a printed value disagrees
with the artifact at the stated precision, if the printed string is absent
from the manuscript, if a percentage that needs its denominator is not
accompanied by the denominator wording nearby, if any truncated digest does
not match one registered SHA-256 on both prefix and suffix, or if any of the
five canonical run pairs it re-hashes (V2, expanded DAFD, W3, W4, W5) differ in
their hash manifests; the closed-loop pairs are compared by their own
byte-identical manifests (Section~\ref{sec:repro}).

\clearpage
\section{Artifact sources of tables and figures}
\label{app:sources}
Figures~\ref{fig:design} and~\ref{fig:loop} and Tables~\ref{tab:resultsmap}, \ref{tab:findings} and \ref{tab:loopguide} are manuscript-level
schematics; every empirical table and figure
(Tables~\ref{tab:repro}--\ref{tab:e7hard} and~\ref{tab:fidelity},
Figures~\ref{fig:cqr}--\ref{fig:e7difficulty}) is derived from the artifact
files below (paths relative to the workspace root); the ledger script of
Appendix~\ref{app:impl} recomputes every printed number from them. Figures
re-rendered for this version from the same artifact data are marked.
{\scriptsize
\begin{longtable}{lp{0.80\linewidth}}
\toprule
Object & Artifact file(s) \\
\midrule
\endfirsthead
\toprule
Object & Artifact file(s) \\
\midrule
\endhead
Table~\ref{tab:repro} & \texttt{artifacts/\allowbreak reproduction\_\allowbreak metrics.json}. \\
Figure~\ref{fig:cqr} & \texttt{artifacts/\allowbreak reliability\_\allowbreak metrics.json} (pilot pair; the artifact data behind \texttt{artifacts/\allowbreak figure\_\allowbreak cqr\_\allowbreak coverage\_\allowbreak shift.png}) and \texttt{expanded\_\allowbreak artifacts\_\allowbreak canonical\_\allowbreak run2/\allowbreak interval\_\allowbreak summary.csv} (expanded pair); re-rendered for v0.3 and extended for v0.3.3 by \texttt{tools/\allowbreak render\_\allowbreak figures.py}. \\
Table~\ref{tab:sources} & MAPE, MAE and $R^2$: \texttt{artifacts/\allowbreak reproduction\_\allowbreak metrics.json}; $n$: \texttt{artifacts/\allowbreak reliability\_\allowbreak metrics.json}. \\
Table~\ref{tab:pruning} & \texttt{artifacts/\allowbreak reliability\_\allowbreak metrics.json}; \texttt{expanded\_\allowbreak artifacts\_\allowbreak canonical\_\allowbreak run2/\allowbreak applicability\_\allowbreak binary\_\allowbreak summary.csv} and \texttt{applicability\_\allowbreak source\_\allowbreak binary\_\allowbreak summary.csv}. \\
Figure~\ref{fig:distance} & \texttt{artifacts/\allowbreak figure\_\allowbreak distance\_\allowbreak vs\_\allowbreak error.png} (re-rendered for v0.3 by \texttt{tools/\allowbreak render\_\allowbreak figures.py} from the artifact data behind that file; legend wording updated for v0.3.3). \\
Figure~\ref{fig:pruning} & \texttt{artifacts/\allowbreak figure\_\allowbreak pruning\_\allowbreak precision.png} (re-rendered for v0.3.3 by \texttt{tools/\allowbreak render\_\allowbreak figures.py} from the tolerance table in \texttt{artifacts/\allowbreak reliability\_\allowbreak metrics.json}; legend wording updated). \\
Table~\ref{tab:v1} & \texttt{blind\_\allowbreak artifacts/\allowbreak blind\_\allowbreak reveal\_\allowbreak results\_\allowbreak recovered.json}; bias and BH-adjusted values from \texttt{blind\_\allowbreak artifacts/\allowbreak posthoc\_\allowbreak analysis\_\allowbreak corrected/\allowbreak blind\_\allowbreak posthoc\_\allowbreak summary.json}. \\
Figure~\ref{fig:v1diag} & \texttt{blind\_\allowbreak artifacts/\allowbreak posthoc\_\allowbreak analysis\_\allowbreak corrected/\allowbreak figure\_\allowbreak 1\_\allowbreak blind\_\allowbreak prediction\_\allowbreak diagnostics.png} (re-rendered for v0.3.2 by \texttt{tools/\allowbreak render\_\allowbreak figures.py} from \texttt{blind\_\allowbreak tx100\_\allowbreak merged\_\allowbreak diagnostics.csv} in the same directory, with wider panel spacing). \\
Figure~\ref{fig:v1risk} & \texttt{blind\_\allowbreak artifacts/\allowbreak posthoc\_\allowbreak analysis\_\allowbreak corrected/\allowbreak figure\_\allowbreak 2\_\allowbreak blind\_\allowbreak risk\_\allowbreak coverage.png}. \\
Table~\ref{tab:v2} & \texttt{v2\_\allowbreak artifacts\_\allowbreak canonical\_\allowbreak run2/\allowbreak model\_\allowbreak summary\_\allowbreak metrics.csv}; run1 is byte-identical by hash manifest. \\
Figure~\ref{fig:v2paired} & \texttt{v2\_\allowbreak artifacts\_\allowbreak canonical\_\allowbreak run2/\allowbreak paired\_\allowbreak group\_\allowbreak differences\_\allowbreak v2p\_\allowbreak vs\_\allowbreak b1.csv}. \\
Figure~\ref{fig:splits} & \texttt{expanded\_\allowbreak artifacts\_\allowbreak canonical\_\allowbreak run2/\allowbreak figure\_\allowbreak 1\_\allowbreak split\_\allowbreak generalization.png} (re-rendered for v0.3.3 by \texttt{tools/\allowbreak render\_\allowbreak figures.py} from \texttt{split\_\allowbreak metrics.csv} in the same directory, with larger fonts). \\
Figure~\ref{fig:riskcov} & \texttt{expanded\_\allowbreak artifacts\_\allowbreak canonical\_\allowbreak run2/\allowbreak figure\_\allowbreak 3\_\allowbreak relative\_\allowbreak risk\_\allowbreak coverage.png} (re-rendered for v0.3 by \texttt{tools/\allowbreak render\_\allowbreak figures.py} from the artifact data behind that file; fonts and control line styles updated for v0.3.3). \\
Figure~\ref{fig:retention} & \texttt{expanded\_\allowbreak artifacts\_\allowbreak canonical\_\allowbreak run2/\allowbreak figure\_\allowbreak 4\_\allowbreak source\_\allowbreak retention.png} (re-rendered for v0.3 by \texttt{tools/\allowbreak render\_\allowbreak figures.py} from the artifact data behind that file; fonts enlarged for v0.3.3). \\
Table~\ref{tab:w3matrix} & \texttt{w3\_\allowbreak artifacts\_\allowbreak canonical\_\allowbreak run2/\allowbreak matrix\_\allowbreak summary.json} and \texttt{summary\_\allowbreak interval.csv}; run1 is byte-identical by hash manifest. \\
Table~\ref{tab:w3aurc} & \texttt{w3\_\allowbreak artifacts\_\allowbreak canonical\_\allowbreak run2/\allowbreak summary\_\allowbreak ad.csv}. \\
Table~\ref{tab:w3robust} & \texttt{posthoc\_\allowbreak review\_\allowbreak round2/\allowbreak w3\_\allowbreak metric\_\allowbreak robustness.json}. \\
Table~\ref{tab:w4} & \texttt{w4\_\allowbreak artifacts\_\allowbreak canonical\_\allowbreak run2/\allowbreak w4\_\allowbreak summary.json}. \\
Table~\ref{tab:w5} & \texttt{w5\_\allowbreak artifacts\_\allowbreak canonical\_\allowbreak run2/\allowbreak w5\_\allowbreak summary.json}. \\
Table~\ref{tab:d3} & \texttt{d3\_\allowbreak blind\_\allowbreak artifacts/\allowbreak reveal/\allowbreak blind\_\allowbreak reveal\_\allowbreak results.json}; thresholds and pre-reveal acceptance from \texttt{d3\_\allowbreak blind\_\allowbreak artifacts/\allowbreak prereveal/\allowbreak blind\_\allowbreak prediction\_\allowbreak bundle.json}; negative-prediction counts from \texttt{d3\_\allowbreak blind\_\allowbreak artifacts/\allowbreak reveal/\allowbreak posthoc\_\allowbreak applicability\_\allowbreak corrected.json}. \\
Table~\ref{tab:r10tiers} & \texttt{scaleup/\allowbreak h\_A1/\allowbreak reveal/\allowbreak h\_\allowbreak reveal\_\allowbreak results.json}; \texttt{scaleup/\allowbreak h\_R2A1/\allowbreak h\_\allowbreak reveal\_\allowbreak results\_\allowbreak b1merged.json}; \texttt{scaleup/\allowbreak h\_R2A2/\allowbreak h\_\allowbreak reveal\_\allowbreak results.json} (byte copies in the repository of the three reveal results; sealed-tier results after the single reveal of each table; development references from the same files). \\
Table~\ref{tab:r10decisions} & \texttt{scaleup/\allowbreak h\_combined/\allowbreak H\_\allowbreak COMBINED\_\allowbreak DECISIONS.json} (protocol-level decisions combined across the three tables by \texttt{scaleup/\allowbreak w6/\allowbreak h\_\allowbreak combine.py}); per-table decisions from the three reveal files above. \\
Table~\ref{tab:hj} & \texttt{h\_\allowbreak fewshot\_\allowbreak rule\_\allowbreak run1/\allowbreak results.json}; fractions and medians from \texttt{h\_\allowbreak fewshot\_\allowbreak rule\_\allowbreak posthoc/\allowbreak descriptive.json}; run2 is byte-identical by hash manifest. \\
Table~\ref{tab:fidelity} & \texttt{fidelity\_\allowbreak run1/\allowbreak fidelity\_\allowbreak results.json}; row-wise table comparison from \texttt{fidelity\_\allowbreak posthoc/\allowbreak g0\_\allowbreak rowwise\_\allowbreak diagnostic.json}; run2 is byte-identical by hash manifest. \\
Tables~\ref{tab:e1policy}, \ref{tab:e2budget}, \ref{tab:e1bgrades} & \texttt{e1e2\_\allowbreak run1/\allowbreak results.json} (E1 levels, differences, questions, control), \texttt{e1\_\allowbreak reanalysis/\allowbreak results.json} (drift levels), \texttt{e1\_\allowbreak dense\_\allowbreak sensitivity/\allowbreak results\_\allowbreak dense.json}, \texttt{e1b\_\allowbreak run1/\allowbreak cell\_*/\allowbreak results.json}, \texttt{e1b\_\allowbreak run1/\allowbreak cell\_*/\allowbreak results\_\allowbreak dense.json} and \texttt{e1b\_\allowbreak run1/\allowbreak e1b\_\allowbreak summary.json} (grades); tier counts and noise floors from \texttt{e0\_\allowbreak run1/\allowbreak tiers\_\allowbreak e0.csv} and \texttt{exclusions.json} \\
Table~\ref{tab:e3lambda} & \texttt{e3\_\allowbreak run1/\allowbreak results.json}; constants from \texttt{e3\_\allowbreak run1/\allowbreak run\_\allowbreak manifest.json} \\
Table~\ref{tab:e4family} & \texttt{e4\_\allowbreak mac\_\allowbreak run1/\allowbreak results.json} \\
Table~\ref{tab:e5transfer} & \texttt{e5\_\allowbreak run1/\allowbreak results.json} \\
Table~\ref{tab:e6ledger} & \texttt{e6\_\allowbreak run1/\allowbreak fragility\_\allowbreak ledger.csv}, \texttt{e6\_\allowbreak run1/\allowbreak results.json}; post-hoc permutations \texttt{e6\_\allowbreak posthoc/\allowbreak summary.json} \\
Table~\ref{tab:e7hard} & \texttt{e7\_\allowbreak run1/\allowbreak results.json} \\
Figure~\ref{fig:e2budget} & \texttt{e1e2\_\allowbreak run1/\allowbreak budget\_\allowbreak table.csv}; drawn by \texttt{scaleup/\allowbreak w7/\allowbreak figures\_\allowbreak closed\_\allowbreak loop.py}, plotted values in \texttt{figures\_\allowbreak closed\_\allowbreak loop/\allowbreak values.json} \\
Figure~\ref{fig:e4slopes} & \texttt{e4\_\allowbreak mac\_\allowbreak run1/\allowbreak results.json}; same script and values file \\
Figure~\ref{fig:e6forest} & \texttt{e1\_\allowbreak reanalysis}, \texttt{e1\_\allowbreak dense\_\allowbreak sensitivity}, \texttt{e1b\_\allowbreak run1} cells, \texttt{e7\_\allowbreak run1} and \texttt{e6\_\allowbreak run1} results; same script and values file \\
Figure~\ref{fig:e7difficulty} & \texttt{e1e2\_\allowbreak run1/\allowbreak loops.csv} with \texttt{e6\_\allowbreak run1/\allowbreak chance.csv}; \texttt{e7\_\allowbreak run1/\allowbreak loops.csv} with \texttt{units.csv}; same script and values file \\
\bottomrule
\end{longtable}
}
\addtocounter{table}{-1}

\section{Reproduction fidelity of the XGBoost baselines}
\label{app:fidelity}
Our XGBoost constructors were written from the method parameters of the
source papers, not copied from upstream code (Section~\ref{sec:e0}). A
frozen protocol (\path{scaleup/fidelity/FIDELITY_PROTOCOL_v1.md}, digest
\hashtrunc{a273102c}{eb701c}) compared them with the upstream implementation
after the fact; both runs are byte-identical by hash manifest and the
deviation register has seven entries.

\emph{Pinned upstream.} The repository named in the source paper's
code-availability statement (\texttt{CIDARLAB/dafd-website}, commit
\texttt{4902f8ce}; the citable snapshot Zenodo 10156784 is tag v3.0.0 =
commit \texttt{d1e6296}) trains \texttt{XGBRegressor(tree\_method="hist")}
with every other argument at its library default under xgboost~1.3.3, after
a \texttt{StandardScaler} on the inputs; the published parameters (100 trees,
depth 6, learning rate 0.3, L2 penalty 1, minimum child weight 1) are those
defaults. The deployed model file carries digest \hashtrunc{91140b28}{d93c62}.
Neither upstream repository carries a licence file, so no upstream file is
redistributed.

\emph{Variants on identical splits.} M0 is our constructor (xgboost~3.3.0,
explicit parameters); M1 the 3.3.0 defaults; M2 adds the upstream input
scaler; M3 adds \texttt{base\_score}~$=0.5$, the 1.3.3 default; M4 is the
upstream line executed verbatim in xgboost~1.3.3 with the scaler; M5 the
same without the scaler. All are scored on the 90 canonical grouped and
random splits of Section~\ref{sec:expanded} and on the external unit (one
fit on all 868 rows scored on the 64 literature rows).

\emph{Gates.} G1 (self-consistency): M0 matched the 90 canonical split rows,
the external row and the R1 baseline mean within $10^{-9}$ before any comparison. G0
(data equivalence): by the pre-registered rule, which matches rows on the
eight rounded inputs and pairs duplicates with their first occurrence, the
upstream tables were declared not equivalent because duplicate input tuples
were paired with different replicate measurements; a post-hoc comparison
that also matches on the measured diameter and the dispersed flow shows that
the 868-row upstream training table equals the OSF comprehensive table row
for row, while 26 of the 64 literature rows in the upstream 932-row table
differ from the OSF generalisability file in hydraulic diameter and
normalised size (6 beyond $10^{-3}$, maximum relative difference 0.153).

\emph{Decisions.} Table~\ref{tab:fidelity} gives the unit means. The
random-split difference between M4 and M0 is 0.021 percentage points
(tolerance 0.5), the sign of every shift unit and the rank order of the five
units are identical, so the pre-registered verdict is ``faithful''. The
ablation chain isolates the only difference: M0, M1 and M2 give identical
predictions (explicit parameters equal the defaults; the scaler changes no
tree), the step to M3 shifts the unit means by between $-0.444$ and $+0.111$
percentage points ($-0.021$ on random splits; largest in magnitude on the 64 literature rows) with a maximum
prediction difference of 0.271 in normalised diameter, and M3, M4 and M5
agree to machine precision (the library version matters only through
\texttt{base\_score}).

\emph{Deployed model.} Loaded in both environments the model file gives the
same predictions. Under the pre-registered inference rule (rows with
absolute percentage error below 1\% count as likely training rows; at least
95\% means trained on all rows) the result is ``inconclusive'': 65.0\% of the
932 rows qualify under standardised inputs (in-sample MAPE 1.15\%, against
40.2\% with unscaled inputs). Separately from that rule, the upstream
constructor fitted with the scaler on all 932 rows, the 868 comprehensive
rows together with the 64 literature rows, gives predictions identical to
the deployed model's to machine precision; for the deployed model the
literature rows are therefore in-sample (MAPE 0.313\%), whereas our external
evaluation holds them out.

\emph{D2.} Our B1 constructor fitted on the source paper's own 392-row
training table reaches 3.17\% MAPE ($R^2=0.927$) on its 76-row test table
against the published 3.9\%, a difference of 0.73 percentage points that
exceeds the pre-registered tolerance of half a point in the favourable
direction; the published configuration was transcribed from the paper's
Methods, the paper's deposit contains data only, and the discrepancy is left
open until the Methods can be re-read.

\begin{table}[htbp]
\centering\small
\caption{Reproduction fidelity on identical splits: mean MAPE (\%) of our D1
constructor (M0, xgboost~3.3.0, explicit parameters) and of the upstream
training line executed verbatim in xgboost~1.3.3 with its input scaler (M4).
Units as in Section~\ref{sec:expanded}; the external unit is one fit on all
868 rows scored on the 64 literature rows.}
\label{tab:fidelity}
\begin{tabular}{lrrr}
\toprule
Unit & splits & M0 (ours) & M4 (upstream, 1.3.3) \\
\midrule
Random 80/20 & 15 & 5.268 & 5.247 \\
Exact-input group & 15 & 5.742 & 5.699 \\
Geometry group & 30 & 11.689 & 11.661 \\
Geometry + fluid group & 30 & 10.176 & 10.287 \\
External (64 rows) & 1 & 11.258 & 10.814 \\
\bottomrule
\end{tabular}
\end{table}

\section{Hash registry}
\label{app:hashes}
Full SHA-256 digests for the truncated values cited in the text; the release lists every digest with its file in \path{hashes/appendix_d_registry.json}.
\begin{center}\scriptsize
\resizebox{\linewidth}{!}{%
\begin{tabular}{ll}
\toprule
Object & SHA-256 \\
\midrule
Comprehensive\_normalized.xlsx & \texttt{aebb0136ff812f7233c8fd57ebc0f27f1b935736e8a9c1d4fdc20585f06b9c06} \\
Generalizability\_data\_normalized.xlsx & \texttt{e90996da3e866e64922dc3b8aaef4b092ee6845fe95aac356c73e90ed108b303} \\
Final\_Comprehensive\_plus\_Generalizability\_data\_normalized.xlsx & \texttt{14ebfaae93a7bbc7e1a73be91a629c57b1720dffac0dd45cef6d7df4e7eb6ac3} \\
Raw\_SE\_dataset.xlsx & \texttt{996862065c33a3f17a36ad2df825179937ddbd7f3c8da41765946a9c1db5e579} \\
matTrainingDataSet.mat & \texttt{7fab0693a7065198468a5d654226f5e2164985e334ed105f91355bdd49fb1a17} \\
matTestDataSet.mat & \texttt{55f202bd19433331c86ddc65507e834036d294f7f9b4db2275f21ad91f6729da} \\
microfluidics\_dataset.zip (Zenodo 17867478) & \texttt{54272086fb390c788c25780856439ba32bfceeae3d40f19bc72af76529af13ee} \\
V1 protocol (BLIND\_VALIDATION\_PROTOCOL.md) & \texttt{1402752425f111a97222ae5c4cb7d13bd68ca1fdfabab978594ac79370e68b1c} \\
V1 pre-registration lock & \texttt{1f6901bc79de2cd189c7813d63e16ba561b7c502ce42721f4ed9e15a2357edc7} \\
V1 prediction bundle & \texttt{2b6eb42d709b367c563e62ca7c82ee764c06ac2b67dffc9f1127b3b732d6932d} \\
V1 sealed labels & \texttt{0e4c9a2577449446875d69ac51713aba510f7f492d7dffc4b6e3595746cc4fb9} \\
V1 frozen reveal implementation & \texttt{46a289ad4369b10b56f8d1c233ba3a401769dec8fbcf392b1235a0baebd27a48} \\
V1 consumed gate & \texttt{e285c4cb4ad7db74b9a3aca4b6240759081095c821865ae76852486f6fb13344} \\
V1 recovery implementation & \texttt{06b6099e0d7c754681bcee7de8f16b8eec8c6c902d4ca996d3e9d0086e263a74} \\
V1 recovered result & \texttt{008d026163af37b7ecddf220d2ffc5aea0818784561a389419e088cf0eaa1624} \\
V1 incident record & \texttt{75615dd3fc6f1d3ea26224c5243b56dd25003dcba48d935e502a6fa72e7db162} \\
V2 protocol (PHYSICS\_CONSTRAINED\_V2\_PROTOCOL.md) & \texttt{5e4a1e4c4a8bb344ed0d1393477a7807b69cc57517eb359a03aec3cbf6471d55} \\
V2 implementation (native\_physics\_v2.py) & \texttt{21ab9ea25eb1bfb8421c49d6a21b216242283286d33999eea88550c5f63b08c7} \\
Method-matrix protocol (W3\_PROTOCOL.md) & \texttt{8f058306de4d51de1c3708443ed46d7c1556fef95ba3479a1cc7389ee46cb3f8} \\
Uncertainty-aware comparison protocol (W4\_PROTOCOL.md) & \texttt{23dc7d0ae5431aeb89e3fe9a9174552e0d0258824658f97a674ccb216e951aa8} \\
Covariate-shift weighting protocol (W5\_PROTOCOL.md) & \texttt{03af554431259a7971ca695db3a80e8cddc270814365c2f4c0c1f63960e60b40} \\
D3 sealed protocol (D3\_BLIND\_PROTOCOL.md) & \texttt{1fa4b404c964c1cfd2d78d2c1fbf5a9965e11560e4d07d2f4e005e15adc91419} \\
D3 parameter binding (D3\_BINDING.json) & \texttt{c943ce8bfab2e2db225e7d3fa3947b8f15ca54ac9063b866af2c1ca9157d284f} \\
D3 prediction bundle & \texttt{79be728c7673803e329ce85574def2c34a869f52d3e2526a1a3161343d296bd2} \\
Fidelity protocol (FIDELITY\_PROTOCOL\_v1.md) & \texttt{a273102cbde3e69137a0a24c2192c34b50637e2c21694668698b940b20eb701c} \\
Deployed upstream model (20231001\_DAFD3\_XGB.json) & \texttt{91140b28dcf309892cbef105d34132ab5e0285473faaa9a75094b67ef8d93c62} \\
Closed-loop protocol E0--E2 (E0\_E1\_E2\_CLOSED\_LOOP\_PROTOCOL\_v1.md, revision 1.1) & \texttt{022aa19d3acc3bf9afcfac99ff53bf244ed54254adcb759cf7010dd120d6fb5f} \\
Closed-loop runner (closed\_loop.py) & \texttt{645b73b2a0afa82154ee54b4018e33973427ea1640358878395afebc5ee13776} \\
Closed-loop analyzer (closed\_loop\_analyze.py) & \texttt{4715eb029191c3266517ca3ef78c03cbd29444234b12504a61e7c30fe3b28c7a} \\
E1b protocol (E1B\_DIFFICULTY\_PROTOCOL\_v1.md) & \texttt{7fa2a69c8c59901fc00669ac544d63d09e014afe94e49cffc65b335eb8826c8e} \\
E1b runner wrapper & \texttt{6f65d59941fac6e39312dd281914bf8f6174fc9e08b37c95ad23ba7606f36661} \\
E1b analyzer & \texttt{c8e7dba8213d24e3379ec3da47296bce7a153c62a555a260aaebcb26b97626c6} \\
E3 protocol (E3\_DECISION\_POLICY\_PROTOCOL\_v1.md, revision 1.1) & \texttt{71166fca292884e45e7585c03e2323fd8b2936455b53934a89b8236d06c1e310} \\
E3 runner wrapper & \texttt{ddf25617321bff59fbd2fec4f8307a842a9e08f93f39cbe69635b677bd2c40b0} \\
E3 analyzer & \texttt{5a091d1641873917df2ea094b6f39cd9c206f7e60e0807389a162e8d26fff76c} \\
E4 protocol (E4\_MODEL\_FAMILY\_PROTOCOL\_v1.md) & \texttt{0b0b68d4e1119bc785a631b20b3d2b1c3af9709c3a4d402138eb5839597c59ce} \\
E4 runner wrapper & \texttt{6bb3772a1290d4c7f8d3e1e3cad2a05df1fffe8b5f4367ccdddb3633e4b23ab4} \\
E4 analyzer & \texttt{27cdb8c7bee460bd4f39d4680b478beb01f9284d539da36d8a1356c97e9eaa34} \\
E5 protocol (E5\_BUDGET\_TRANSFER\_PROTOCOL\_v1.md) & \texttt{52fd522ed0d9dbac962db91dfe732df75ea5c630f22e52d4d88f13ce9a34e1bc} \\
E5 analyzer & \texttt{a724a9b6d872e14f4d3f4fca3fc99f11a0a7a1eb2ad2f69f52b976929af2e494} \\
E6 protocol (E6\_NEGATIVE\_CONTROLS\_PROTOCOL\_v1.md, revision 1.1) & \texttt{d28896aea17d28d427857c8db862209cffb50616a0446d4a9e9ae910092b8a06} \\
E6 runner wrapper & \texttt{8d76252092eab9b903c836dac05149f8d450871d365eca9ae4f8470b7932c6fc} \\
E6 analyzer & \texttt{edc6f1c50c2bff397538eb214e32b62e8ab09221155397358c76fb2e912aa056} \\
E7 protocol (E7\_HARD\_TASK\_PROTOCOL\_v1.md, revision 1.1) & \texttt{130a150965ef742753e63692da5474d6fe9fa925987a16d9914fa13b44f71425} \\
E7 runner wrapper & \texttt{09ae934d9878a8230c3ce7ebfdbe00f09a2986215d4fe758ecf5758e7c654eb7} \\
E7 analyzer & \texttt{f64edbeeae25fec8e29c82f65cddf26e0beb44f543f38d672b5302e43f39a48a} \\
\bottomrule
\end{tabular}}
\end{center}

\end{document}